\RequirePackage{fix-cm} 
\documentclass[ twoside,
                titlepage,numbers=noenddot,
                footinclude=false,cleardoublepage=empty,
                paper=a4,fontsize=11pt
                ]{scrreprt}

\PassOptionsToPackage{utf8}{inputenc}
	\usepackage{inputenc}

\PassOptionsToPackage{eulerchapternumbers,
                     listings,
					 pdfspacing,
					 linedheaders, 
					 beramono,
					 }{classicthesis}                                        

\newcommand{\myTitle}{Adversarial Robustness of Representation Learning for Knowledge Graphs\xspace} 
\newcommand{\myDegree}{Doctor of Philosophy\xspace} 
\newcommand{\myName}{Peru Bhardwaj\xspace} 
\newcommand{\myEmail}{\texttt{bhardwap@tcd.ie}}
\newcommand{\myTime}{September 2022\xspace}
\newcommand{\myLocation}{Dublin, Ireland\xspace}

\newcommand{\typeofthesis}{Thesis\xspace} 
\newcommand{\keywords}{Machine Learning, Knowledge Graphs, Adversarial Robustness} 
\newcommand{\mySchool}{\href{http://www.scss.tcd.ie}{\color{black}{School of Computer Science and Statistics}}\xspace} 

\newcounter{dummy} 
\providecommand{\mLyX}{L\kern-.1667em\lower.25em\hbox{Y}\kern-.125emX\@}

\usepackage[english]{babel}                  

\usepackage{csquotes}

\usepackage[]{natbib}
\PassOptionsToPackage{fleqn}{amsmath}       
    \usepackage{amsmath}

\PassOptionsToPackage{T1}{fontenc} 
    \usepackage{fontenc}     
\usepackage{textcomp} 
\usepackage{scrhack} 
\usepackage{xspace} 
\usepackage{mparhack} 
\usepackage{fixltx2e} 
\PassOptionsToPackage{printonlyused,smaller}{acronym} 
    \usepackage{acronym} 
    
\usepackage{tabularx} 
\usepackage{caption}
\usepackage{listings} 
\PassOptionsToPackage{pdftex,hyperfootnotes=false,pdfpagelabels}{hyperref}
    \usepackage[backref=page]{hyperref}  
\PassOptionsToPackage{pdftex}{graphicx}
    \usepackage{graphicx} 

\addto\extrasenglish{%
}

\renewcommand*{\backref}[1]{}
\renewcommand*{\backrefalt}[4]{({\small%
    \ifcase #1 Not cited.%
          \or Cited on page~#2%
          \else Cited on pages #2%
    \fi%
    })}
\hypersetup{%
    colorlinks=true, linktocpage=true, pdfstartpage=3, pdfstartview=FitV,%
    breaklinks=true, pdfpagemode=UseNone, pageanchor=true, pdfpagemode=UseOutlines,%
    plainpages=false, bookmarksnumbered, bookmarksopen=true, bookmarksopenlevel=1,%
    hypertexnames=true, pdfhighlight=/O,
    urlcolor=TCDBlue, linkcolor=TCDBlue, citecolor=TCDBlue, 
    pdftitle={\myTitle},%
    pdfauthor={\myName},%
    pdfsubject={\myDegree},%
    pdfkeywords={\keywords},%
    pdfcreator={pdfLaTeX},%
}   

\usepackage{classicthesis} 

\usepackage[%
  paper=a4paper,
  left=35mm,
  right=35mm,
  top=38mm,
  bottom=38mm,
]{geometry}
\definecolor{tcdblue}{RGB}{5, 105, 185}
\definecolor{TCDBlue}{cmyk}{100, 50, 0, 0}

\usepackage{layout}  
\usepackage[onehalfspacing]{setspace}  
\usepackage{amsmath}
\usepackage{amsthm}
\usepackage{amssymb}
\usepackage{multirow}
\usepackage{lipsum}
\usepackage{comment}
\usepackage{booktabs}
\usepackage{subcaption} 
\usepackage{graphicx}
\usepackage{bm}
\usepackage{nicefrac}
\usepackage{mathtools} 
\usepackage{listings} 

\theoremstyle{definition}
\newtheorem{definition}{Definition}[section]
\newcommand{\norm}[1]{\left\lVert#1\right\rVert}
\newcommand{\abs}[1]{\mid #1 \mid}
\renewcommand*{\vec}[1]{\ensuremath{\bm{#1}}}  
\newcommand{\mat}[1]{\bm{#1}} 
\newcommand{\tensor}[1]{\underline{\bm{#1}}} 
\newcommand{\set}[1]{\ensuremath{\mathcal{#1}}}  

\newcommand{\loss}{\mathcal{L}}  
\newcommand{\atk}{\mathcal{A}}  
\newcommand{\atkloss}{\ensuremath{{\mathcal{L}_{\text{atk}}}}} 
\newcommand{\trainloss}{\ensuremath{{\mathcal{L}_{\text{train}}}}} 
\newcommand*{\param}{\ensuremath{ \bm{ \theta }}}
\DeclareMathOperator*{\argmin}{arg\,min}
\DeclareMathOperator*{\argmax}{arg\,max}
\newcommand{\tdot}[3]{\ensuremath{\langle #1, #2, #3 \rangle}}
\newcommand{\negative}[1]{#1^{\ast}}  
\newcommand{\adversarial}[1]{#1'} 

\begin{document}
\frenchspacing
\raggedbottom  
\pagenumbering{roman}
\deftriplepagestyle{pgnumbottomcenter}{}{}{}{}{\pagemark{}}{}
\pagestyle{pgnumbottomcenter}
\renewcommand{\chapterpagestyle}{pgnumbottomcenter}
\thispagestyle{empty}
\begin{center}
    \spacedlowsmallcaps{\myName} \\ \medskip                        

    \begingroup
        \color{tcdblue}\spacedallcaps{Adversarial Robustness of \\Representation Learning for Knowledge Graphs}
    \endgroup
\end{center}        


\begin{titlepage}
\begin{center}
    
    \includegraphics[width=1\textwidth]{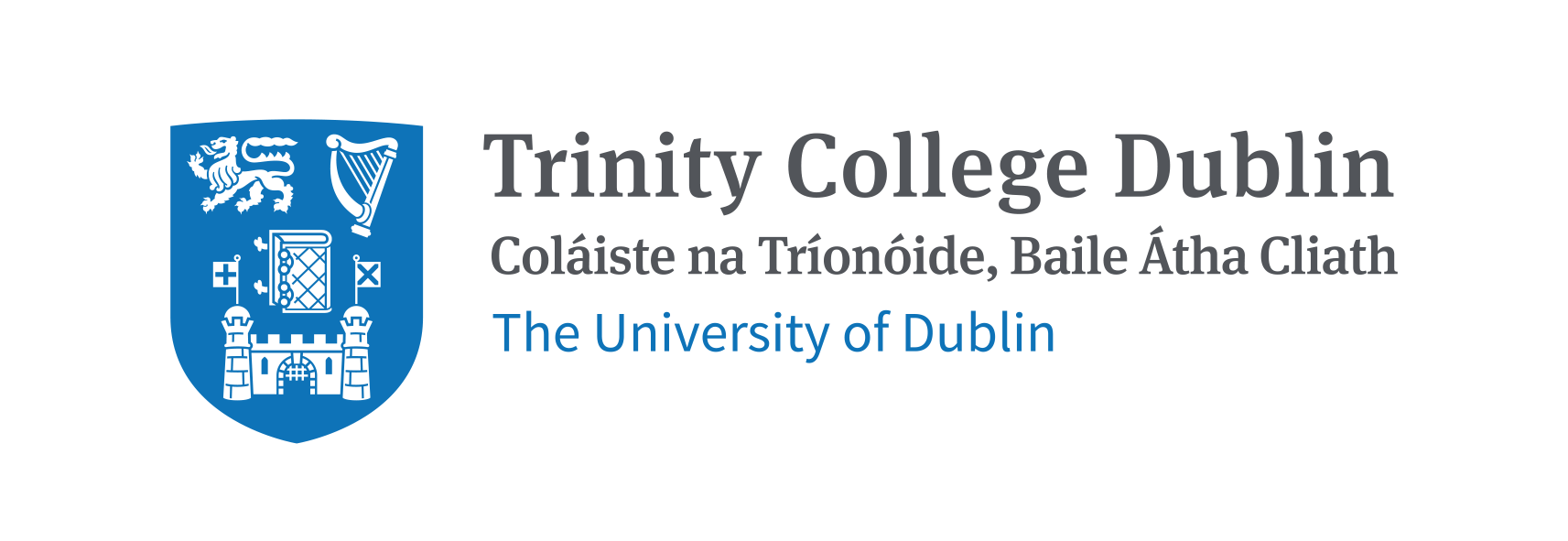} 
    \Large \textsf{\mySchool}
    
    \vfill
    \vfill

        \begingroup
           \huge \color{tcdblue} {
           Adversarial Robustness of \\Representation Learning for \\Knowledge Graphs
           } \\ \bigskip
        \endgroup

        \Large \spacedlowsmallcaps{\myName}
        
        \large{\myEmail}

        \vfill
        
        \large \myTime\
        
    \vfill
     A \typeofthesis\ submitted in partial fulfilment\\of the requirements for the degree of\\
    \myDegree
    
    \vfill

\end{center}

\end{titlepage}
\thispagestyle{empty}

\hfill

\vfill

\begingroup

\raggedright
\noindent\myName: 
\linebreak
\textit{\myTitle,}
\linebreak
\textcopyright\ \myTime

\endgroup

\cleardoublepage
\refstepcounter{dummy}
\pdfbookmark[0]{Declaration}{declaration}
\chapter*{Declaration}
\thispagestyle{empty}
\vspace{1cm}
I declare that this thesis has not been submitted as an exercise for a degree at this or any other university and it is entirely my own work.

\vspace{1cm}
\noindent
I agree to deposit this thesis in the University’s open access institutional repository or allow the Library to do so on my behalf, subject to Irish Copyright Legislation and Trinity College Library conditions of use and acknowledgement.
\vspace{1cm}

\noindent
I consent to the examiner retaining a copy of the thesis beyond the examining period, should they so wish (EU GDPR May 2018).
\vspace{3cm}

\bigskip
 
\noindent\textit{\myLocation, \myTime}

\smallskip

\begin{flushright}
    \begin{tabular}{m{5cm}}
        \\ \hline
        \centering\myName \\
    \end{tabular}
\end{flushright}

\cleardoublepage
\pdfbookmark[0]{Abstract}{Abstract}

\chapter*{Abstract}
\thispagestyle{empty}
Knowledge graphs represent factual knowledge about the world as relationships between concepts and are critical for intelligent decision making in enterprise applications. New knowledge is inferred from the existing facts in the knowledge graphs by encoding the concepts and relations into low-dimensional feature vector representations. The most effective representations for this task, called Knowledge Graph Embeddings (KGE), are \emph{learned} through neural network architectures. Due to their impressive predictive performance, they are increasingly used in high-impact domains like healthcare, finance and education. However, are the black-box KGE models adversarially robust for use in domains with high stakes?

This thesis argues that state-of-the-art KGE models are vulnerable to data poisoning attacks, that is, their predictive performance can be degraded by systematically crafted perturbations to the training knowledge graph. To support this argument, two novel data poisoning attacks are proposed that craft input deletions or additions at training time to subvert the learned model's performance at inference time. These attacks target the task of predicting the missing facts in knowledge graphs using Knowledge Graph Embeddings.

To degrade the model performance through adversarial deletions, the use of model agnostic instance attribution methods is proposed. These methods are used to identify the training instances that are most influential to the KGE model’s predictions on target instances. The influential triples are used as adversarial deletions. %
To poison the KGE models through adversarial additions, their inductive abilities are exploited. The inductive abilities of KGE models are captured through the relationship patterns like symmetry, inversion and composition in the knowledge graph. Specifically, to degrade the model’s prediction confidence on target facts, this thesis proposes to improve the model’s prediction confidence on a set of decoy facts. Thus, adversarial additions that can improve the model’s prediction confidence on decoy facts through different relation inference patterns are crafted. 

Evaluation of the proposed adversarial attacks shows that they outperform state-of-the-art baselines against four KGE models for two publicly available datasets. Among the proposed methods, simpler attacks are competitive with or outperform the computationally expensive ones. %
The thesis contributions not only highlight and provide an opportunity to fix the security vulnerabilities of KGE models, but also help to understand the black-box predictive behaviour of these models.


\cleardoublepage
\pdfbookmark[0]{Acknowledgement}{Acknowledgement}

\bigskip

\begingroup
\let\clearpage\relax
\let\cleardoublepage\relax
\let\cleardoublepage\relax
\chapter*{Acknowledgement} 
\thispagestyle{empty}
This research was conducted with the financial support of Accenture Labs and Science Foundation Ireland (SFI) at the ADAPT SFI Research Centre at Trinity College Dublin (TCD). The ADAPT SFI Centre for Digital Content Technology is funded by Science Foundation Ireland through the SFI Research Centres Programme and is co-funded under the European Regional Development Fund (ERDF) through Grant No. 13/RC/2106 P2.

\noindent
A fees-free extension for the submission of thesis was supported by the Office of the Dean of Graduate Studies at TCD, and research costs by the Irish Higher Education Authority (HEA) through Trinity HEA Covid Extension Fund 2020.

\vspace{1cm}
\noindent
The research project was supervised by Prof. Declan O'Sullivan (ADAPT Centre, TCD); and co-supervised by Dr. Luca Costabello (Accenture Labs) and Prof. John Kelleher (ADAPT Centre, TU Dublin). The project was also co-supervised, at different stages by Dr. Fabrizio Orlandi (ADAPT Centre, TCD), and Dr. Freddy Lecue (Accenture Labs). The order is based on the length of involvement in the project.

\vspace{1cm}
\noindent
Thanks to the project sponsors and supervisors for facilitating the research. 

\noindent
Thank you to colleagues, housemates, friends and family for support and companionship over the years.

\bigskip

\endgroup

\pagestyle{scrheadings}  
\cleardoublepage
\refstepcounter{dummy}
\pdfbookmark[0]{\contentsname}{tableofcontents}
\setcounter{tocdepth}{2} 
\setcounter{secnumdepth}{3} 
\manualmark
\markboth{\spacedlowsmallcaps{\contentsname}}{\spacedlowsmallcaps{\contentsname}}
\tableofcontents 
\automark[section]{chapter}
\renewcommand{\chaptermark}[1]{\markboth{\spacedlowsmallcaps{#1}}{\spacedlowsmallcaps{#1}}}
\renewcommand{\sectionmark}[1]{\markright{\thesection\enspace\spacedlowsmallcaps{#1}}}
\clearpage


\cleardoublepage\pagenumbering{arabic}
\cleardoublepage

\chapter{Introduction}
\label{ch:Introduction}
\section{Motivation}
Consider a financial regulator that wants to detect and anticipate money laundering activities, where malicious parties funnel their income from illicit activities into legitimate bank accounts through a series of financial transactions. 
Given the complex and interconnected nature of such activities, analysing the financial transactions from isolated bank accounts is often insufficient to catch fraudsters in practice.
Rather, the relationships between different financial entities are leveraged to perform an integrated analysis across diverse sources of financial data.
To enable this integrated analysis, knowledge graphs have emerged as the de-facto standard for modeling and integrating the factual knowledge about multiple financial entities from diverse sources \citep{hogan2021knowledgegraphs, noy2019knowledgegraphs}.
Entities like bank accounts, their owners and their assets are represented as nodes of the knowledge graph, and the interactions and transactions between these entities as the labelled edges of the graph.
Due to the growing volume of digital financial transactions, data-driven Machine Learning (ML) methods are used to mine the patterns in this interconnected graph of transactions.
Learning and inference on the graph data drives intelligent decisions like predicting the suspicious transactions between different accounts or hidden affiliations between the account owners \citep{khalili2020deloitte, chang2020ficonsulting, singson2021graphaisummit}.
However, malicious parties with illicit income to launder would be highly motivated to escape detection from these systems.
Thus, such parties might attempt to sabotage the predictions of data-driven intelligent systems by manipulating their personal details and transactions in the input knowledge graph.
\emph{Are the predictions from Machine Learning models on knowledge graphs reliable for use in such adversarial settings?}

Like the anti-money laundering application in finance, knowledge graphs are a ubiquitous representation of the factual knowledge about interconnected entities and the relationships between them \citep{hogan2021knowledgegraphs}. In recent years, several large-scale knowledge graphs have been developed to support intelligent decision making for applications ranging from search engines, e-commerce and social networks to biomedicine and finance. 
Commercial enterprises like Google and Microsoft have built web-scale knowledge graphs from textual sources on the Web to support Google Search and Bing. Similarly, Facebook and LinkedIn rely on the graph representation of the knowledge about their users to understand user preferences and recommend prospective connections or job opportunities. Enterprise knowledge graphs are also used by online vendors like Amazon and eBay to encode shopping behaviour of their users and information of their products for improved product recommendations. Other companies like Accenture, Deloitte and Bloomberg have deployed knowledge graphs for financial services. These financial graphs power applications like enterprise search, financial data analytics, risk assessment and fraud detection \citep{hogan2021knowledgegraphs, noy2019knowledgegraphs}. 

More generally, in the domain of Natural Language Processing (NLP), background knowledge represented as factual knowledge graphs is injected to support knowledge-aware applications, like knowledge based question answering or explainable fact checking over knowledge bases \citep{ji2022knowledgegraphssurveyNLP, kotonya2020explainableautomatedfactcheckingsurvey}. An emerging research direction in this domain investigates methods to combine the structured knowledge representation with the language representation learned from unstructured text. These methods seek to improve the contextual reasoning and understanding capabilities of NLP models using factual as well as commonsense knowledge \citep{malaviya2020commonsense, he2020_graph_languagemodels, zhang2022greaselm_graphreasoning_languagemodels}. 
Similarly, in the domain of Computer Vision (CV), machine learning for tasks like image classification, Visual Question Answering and skeleton based action recognition, is enhanced by representing the relationships among the objects in a scene or an image as knowledge graphs \citetext{\citealp[see][Chapter~11]{ma2021deeplearningongraphs}}. 
On the other hand, there are emerging applications of knowledge graphs in healthcare and biomedical research. Here, biological networks as used to model the associations and interactions between different protein structures, drugs and diseases. Additionally, combining the biomedical knowledge with patients' electronic health records enables an integrated analysis of disease comorbidity for personalised precision medicine \citep{rotmensch2017healthkgelectronicmedicalrecords, mohamed2020knowledgegraphdrugdiscovery, li2021graphrepresentationlearningbiomedicine, bonner2021knowledgegraphdrugdiscovery}.
Thus, knowledge graphs are the backbone for learning and inference in modern intelligent systems.

To incorporate the graph data into the standard ML pipeline, the symbolic graph structure needs to be represented as differentiable feature vectors. Traditional algorithms used heuristics and domain engineering to hand-craft the features based on the statistics about graph or node topology, or the kernel methods \citetext{\citealp[see][Chapter~2]{hamilton2020graphrepresentationlearning}}.
The lack of flexibility of feature engineering has led way to the \emph{graph representation learning} algorithms that \emph{learn} to represent the graph structure as low-dimensional continuous feature vectors, also called embeddings. 
To learn the embedding for an entity, these algorithms aim to preserve the structural information about the entity's neighbourhood in the graph domain as a similarity measure in the embedding domain. 
This way, algebraic operations on the embeddings of entities reflect the graph structural interactions between those entities, which in turn, allows the topological information from the graph to be used for different ML tasks.
Deep learning based neural network architectures are used to optimize the representations of entities and relations in the knowledge graph, such that the learned representations best support the downstream ML task to be performed on the graph \citep[Part~1]{hamilton2017graphrepresentationlearning, hamilton2020graphrepresentationlearning}.
Due to their effectiveness for different downstream tasks, graph representation learning algorithms have become state-of-the-art methods for learning and reasoning with knowledge graphs \citep{chen2020reasoningonknowledgegraphs,nickel2015review}.

The success of deep learning based methods is ascribed to their ability to extract rich statistical patterns from large volumes of input data. However, being data-driven, the learned models are non-interpretable and the reasons for their predictions are unknown. Due to this black-box predictive behaviour, the failure modes of the models are also unknown. It has been shown that the predictions from deep learning models can be manipulated by manipulating their input data \citep{biggio2018wild, joseph_nelson_rubinstein_tygar_2019}. This is especially concerning for high-stakes domains like healthcare, finance, education or law enforcement where the representation learning algorithms for knowledge graphs are increasingly used \citep{mohamed2020knowledgegraphdrugdiscovery,bonner2021knowledgegraphdrugdiscovery}. In these domains, decision outcomes impact human lives and the stakes for model failure are very high. On the other hand, due to the high-stakes, there are likely going to be motivated adversarial actors that want to manipulate the model predictions. Additionally, knowledge graphs are often automatically extracted from textual sources on the Web or curated from user generated content \citep{nickel2015review, ji2022knowledgegraphssurveyNLP}. This makes it easy for the adversaries to inject carefully crafted false data to the graph.

Thus, to deploy graph representation learning models in high-stakes user facing domains, impressive predictive performance of the models is not enough. It is vital to ensure that the models are also safe and robust for use. Yet, building adversarially robust models requires methods to measure the adversarial robustness of a model. In other words, a necessary pre-requisite for adversarially robust graph representation learning models are methods to identify the failure modes or the security vulnerabilities of existing models.

\emph{The research in this thesis is motivated by the need to identify the adversarial vulnerabilities of black-box graph representation learning algorithms as a crucial step towards their responsible integration in high-stakes user facing applications.}

\section{Research Problem}
The research in Machine Learning (ML) aims to build intelligent computer applications that learn to perform real-world tasks from past experience instead of being explicitly programmed by human operators \citep{goodfellow2016deeplearning}. 
ML systems automatically recognize and extract patterns from the past evidence provided to the learner in the form of sample data. The patterns and assumptions observed in the input data are captured implicitly in a latent model that enables inferences on the unseen data. This predictive model design has been used successfully to make intelligent decisions from input data modalities like images, text and graphs. Because of the ability of the learner to adapt to input evidence, Machine Learning has emerged as the most promising approach to tackle real-world problems. With the availability of large volumes of training data, these systems now have applications in several daily tasks \citep{barreno2010security, biggio2018wild}.

However, the adaptability of ML systems to input data can also become a vulnerability in real-world settings.
In these settings, there might be motivated malicious actors that attempt to exploit the adaptive nature of the learning system to manipulate its predictions. The adversary can carefully design training data samples that steer the model to learn patterns that result in manipulated predictions. Such intentional perturbations of an ML model's input data, that aim to manipulate the model predictions, constitute the security threat called \emph{adversarial attacks}. The security analysis of ML systems is the focus of the research field of Adversarial Machine Learning (AML), that lies at the intersection of the wider fields of Machine Learning and Computer Security. 
The latter advocates for a proactive approach to system security, also called security by design.
Instead of designing countermeasures in response to the adversarial attacks on deployed systems, the proactive approach anticipates these attacks during system design and evaluates the system's resilience (that is, its adversarial robustness) to the simulated attacks \citep{biggio2018wild, joseph_nelson_rubinstein_tygar_2019, barreno2010security}.

This design approach has been widely used to investigate the adversarial robustness of ML systems for domains like images and natural language. 
However, adversarial attacks designed against ML systems for images or text are not directly applicable to graph data. This is because of the discrete perturbation space for the graph domain and the interconnected nature of the graph structure \citep{gunnemann2022graphadversarialrobustness}.
There have been recent efforts towards investigating the adversarial robustness of machine learning for graph data, as surveyed in \citet{gunnemann2022graphadversarialrobustness}, \citet{jin2021adversarialattacksongraphsSIGKDD} \citet{xu2020advgraphsurvey} and \citet{sun2018advgraphsurvey}. However, these studies focus on the adversarial vulnerabilities of undirected graphs and neural network architectures that do not scale to large-scale knowledge graphs. Research on the adversarial vulnerabilities of representation learning for knowledge graphs has received very little attention. This is precisely the research gap addressed by this thesis.

\emph{This thesis examines the adversarial robustness of representation learning algorithms for knowledge graphs through the design, implementation and evaluation of data poisoning attacks against them.}

Data poisoning attacks aim to degrade the predictive performance of the learned model at inference time by making edits to the input knowledge graph at training time. The focus of this thesis are Knowledge Graph Embedding (KGE) models that are the state-of-the-art algorithms for representation learning on knowledge graphs \citep{ali2021bringinglightintodark, rossi2021kgesurvey}. 
Since the knowledge graphs are often curated automatically from textual sources on the Web or from user generated content, they are incomplete in practice. Thus, an important and state-of-the-art downstream application of KGE models is to predict the missing facts in knowledge graphs \citep{nickel2015review}.
To investigate the adversarial robustness of KGE models, the thesis focuses on this downstream task of missing link prediction in knowledge graphs. 
The aim of the research is to design methods that craft facts which can be added to or removed from the original knowledge graph, such that a KGE model trained on the perturbed graph exhibits worse predictive performance than a KGE model trained on the original graph.

\subsection*{Research Question}

The research question to be addressed in the thesis is - 
\begin{center}
    \emph{To what extent can the predictive performance of state-of-the-art KGE models be degraded for missing link prediction at inference time by adding or removing systematically crafted triples to the input knowledge graph at training time? }
\end{center}

\noindent
To address this research question, the thesis addresses three research objectives - 
\begin{enumerate}
    \item[RO1] Designing an adversarial attack requires a hypothetical attack scenario, which is defined by a threat model of the adversary. The attack threat model specifies the goals, knowledge and capability of the attacker. The first objective of the research is to define this threat model and formulate the problem statement for the corresponding adversarial attacks. This objective also involves the identification of major challenges in designing these adversarial attacks, as well as the existing adversarial attacks against KGE models from the literature.
    \item[RO2] The second research objective for this thesis is to design and implement novel adversarial attacks to systematically craft deletions or additions to the training dataset that aim to degrade the predictive performance of the learned KGE models. This objective further includes two sub-objectives - 
    \begin{enumerate}
        \item[RO2.1] Propose adversarial attacks on KGE models that degrade their predictive performance through \emph{adversarial deletions}.
        \item[RO2.2] Propose adversarial attacks on KGE models that degrade their predictive performance through \emph{adversarial additions}.
    \end{enumerate}
    \item[RO3] The third research objective is to evaluate the effectiveness of the proposed adversarial attacks in degrading the predictive performance of KGE models.
\end{enumerate}

\section{Research Contributions}

Data poisoning attacks aim to degrade the predictions of the KGE model at inference time by removing or adding adversarial triples to the input knowledge graph at training time.
Designing data poisoning attacks against KGE models poses two main challenges. First, to select an adversarial deletion or addition, the impact of a candidate perturbation on the model predictions needs to be measured. However, the naïve approach of re-training a new KGE model for each candidate perturbation is computationally prohibitive. Second, while the search space for adversarial deletions is limited to existing triples in the knowledge graph, it is computationally intractable to enumerate through all the possible candidate additions.

\subsection*{Poisoning via Instance Attribution Methods}
This thesis proposes to use the model-agnostic \emph{instance attribution methods} from Interpretable Machine Learning \citep{molnar2019interpretablemlbook} to craft adversarial deletions against KGE models. Instance attribution methods identify the training instances that are \emph{influential} to model predictions, that is, deleting these instances from the training data would considerably change the model predictions. These methods are widely used to generate post-hoc explanations for deep neural networks on images \citep{koh2017understanding, hanawa2021evaluationsimilaritymetrics} and text \citep{han2020influencenlp, pezeshkpour2021instanceattribution}. Since the KGE models have relatively shallow neural architectures and the instance attribution metrics are independent of the black-box models and the input domain, they are a promising approach to estimate the influence of training triples on the KGE model predictions. Yet, despite their promise, they have not been used for KGE models so far. The instance attribution methods are used in this research to address the challenge of measuring the impact of a candidate adversarial deletion on the KGE model predictions.

The research uses three types of instance attribution methods - Instance Similarity that compares the feature representations of target and training triples \citep{hanawa2021evaluationsimilaritymetrics, charpiat2019inputsimilarity}; Gradient Similarity that compares the gradients of model's loss function due to target and training triples \citep{hanawa2021evaluationsimilaritymetrics,charpiat2019inputsimilarity}; and Influence Function \citep{koh2017understanding} which is a principled approach from the robust statistics to estimate the effect of removing a training triple on the KGE model predictions. The most influential training triple identified by the instance attribution metrics is selected as adversarial deletion.

However, data poisoning attacks based on instance attribution methods require a combinatorial search through the candidate perturbations, which does not scale to adversarial additions. Thus, a heuristic method is proposed to use the instance attribution methods for data poisoning by adversarial additions. Given the influential training triple, the triple for adversarial addition is obtained by replacing one of the two entities of the influential triple with the most dissimilar entity in the embedding space. The intuition behind this step is to add a triple that would reduce the influence of the influential triple. This solution overcomes the scalability challenge for adversarial additions by comparing only the entity embeddings to select the replacement. %
However, this is a heuristic and model-agnostic solution for adversarial additions and might not generate the most effective edits. Further, the method still requires combinatorial search over the neighbourhood triples. Thus, in the next part of the thesis, model-aware adversarial additions, that do not rely on a combinatorial search in the triple space, are proposed.

\subsubsection*{Poisoning via Relation Inference Patterns}
To degrade the model performance, the inductive abilities of KGE models are exploited to craft the facts for adversarial addition.
The inductive abilities of KGE models are expressed through different connectivity patterns like symmetry, inversion and composition between the relations in the knowledge graph \citep[Chapter~4]{trouillon2019inductive, hamilton2020graphrepresentationlearning}.
These connectivity patterns are referred to as \emph{relation inference patterns}.
Specifically, to degrade the KGE model’s performance for predicting a target missing fact, this thesis proposes to improve the model’s predictive performance on a set of decoy facts.

A collection of heuristic approaches are proposed to select the decoy facts and craft adversarial additions that can improve the model’s predictive performance on decoy facts through different relation inference patterns. 
The proposed solution addresses the challenge of a large candidate space by breaking down the search into smaller steps over entities and relations. These steps are - (i) determining the adversarial relations; (ii) determining the decoy entities that most likely violate an inference pattern; and (iii) determining the remaining adversarial entities in the inference pattern that are most likely to improve the predictions for decoy triples. 

Furthermore, the extent of effectiveness of the attack relying on an inference pattern indicates the KGE model's sensitivity to that pattern. 
This means that the proposed data poisoning attacks for adversarial additions can also help in understanding the predictive behaviour of KGE models.

In summary, the main contributions of this thesis are - 
\begin{enumerate}
    \item[RC1] Instance attribution based adversarial attacks for adversarial deletions
    \item[RC2] Relation Inference pattern based adversarial attacks for adversarial additions
\end{enumerate}

\section{Publications}
The research contributions from this thesis have been published as the following articles at international conferences.
\begin{enumerate}
    \item \textbf{Bhardwaj, P.}, Kelleher, J.$^{\ast}$, Costabello, L.$^{\ast}$, \& O’Sullivan, D.$^{\ast}$ (2021, November). Adversarial Attacks on Knowledge Graph Embeddings via Instance Attribution Methods. In Proceedings of the 2021 Conference on Empirical Methods in Natural Language Processing (pp. 8225-8239). \citep{bhardwaj2021instanceattribution}
    
    \item \textbf{Bhardwaj, P.}, Kelleher, J.$^{\ast}$, Costabello, L.$^{\ast}$, \& O’Sullivan, D.$^{\ast}$ (2021, August). Poisoning Knowledge Graph Embeddings via Relation Inference Patterns. In Proceedings of the 59th Annual Meeting of the Association for Computational Linguistics and the 11th International Joint Conference on Natural Language Processing (Volume 1: Long Papers) (pp. 1875-1888). \citep{bhardwaj2021relationinferencepatterns}
    
    \item \textbf{Bhardwaj, P.} (2020, April). Towards Adversarially Robust Knowledge Graph Embeddings. In Proceedings of the AAAI Conference on Artificial Intelligence (Vol. 34, No. 10, pp. 13712-13713). \citep{bhardwaj2020towardsrobustKGE}
\end{enumerate}

For publications with multiple authors, the thesis author proposed and designed the research contributions, conducted the evaluation and wrote the manuscripts. Remaining authors contributed equally (indicated by $\ast$ next to their names) through feedback on the research process and the manuscript drafts.

\citet{bhardwaj2021instanceattribution} is based on the first research contribution RC 1 of this thesis, which is discussed in Chapter \ref{ch:instance_attribution} of the thesis. The contribution was published at EMNLP 2021 and it proposes the data poisoning attacks for adversarial deletions using the instance attribution methods. The second research contribution RC 2 of this thesis has been published as \citet{bhardwaj2021relationinferencepatterns} at ACL-IJCNLP 2021. This contribution is discussed in Chapter \ref{ch:relation_inference} of the thesis and proposes data poisoning attacks for adversarial additions by exploiting the inductive abilities of the KGE models expressed as relation inference patterns. Lastly, the author's research proposal for this thesis was published at the Doctoral Consortium at AAAI 2020 as \citet{bhardwaj2020towardsrobustKGE}.

\section{Thesis Outline}

The next chapter provides the background details for representation learning on knowledge graphs. The pipeline for state-of-the-art KGE models is introduced and relevant components of this pipeline are discussed. The downstream task of missing link prediction and the standard evaluation protocol for this task are also discussed.
Chapter \ref{ch:problem_formulation} formulates the problem statement for designing data poisoning attacks against the KGE models. This chapter defines the attack threat model used for the thesis and identifies the key challenges to be addressed in the design of data poisoning attacks. State-of-the-art poisoning attacks against KGE models and a methodology for evaluating the proposed attacks are also discussed. This chapter addresses the Research Objective RO1.

Chapter \ref{ch:instance_attribution} proposes the adversarial attacks based on instance attribution methods and evaluates the proposed attacks against state-of-the-art adversarial attacks. 
In Chapter \ref{ch:relation_inference}, the adversarial attacks based on relation inference patterns are discussed and evaluated against the state-of-art adversarial attacks. Both Chapters \ref{ch:instance_attribution} and \ref{ch:relation_inference} collectively address Research Objectives RO2 and RO3. 

Chapter \ref{ch:related_work} highlights studies from the wider literature that are related to the research in this thesis.
Families of representation learning on graphs with other modalities are discussed, and the literature on understanding the predictive behaviour of KGE models through post-hoc explanations and theoretical or empirical analysis is analyzed.
Finally, Chapter \ref{ch:conclusion} concludes the thesis by summarizing the broader impact of the research contributions and discussing the open research problems for building adversarially robust KGE models.

\chapter{Representation Learning for Knowledge Graphs}
\label{ch:background}

Knowledge graphs represent factual knowledge about the world as relationships between concepts and are the de-facto standard for modeling and integrating complex, structured data from diverse sources \citep{hogan2021knowledgegraphs, noy2019knowledgegraphs}. Learning and inference on knowledge graphs is critical for intelligent decision making in enterprise applications that aim to utilize the factual world knowledge. Such applications range from question answering systems for search engines and personal assistants, to recommendation systems for social networks and e-commerce, to drug target discovery from biological networks \citep{noy2019knowledgegraphs, nickel2015review, ji2022knowledgegraphssurveyNLP}.

While a knowledge graph is represented as a collection of facts, the standard Machine Learning (ML) pipeline uses feature vectors as inputs. Thus, integrating knowledge graphs into these pipelines requires a mapping of the concepts and relations in the knowledge graph to feature vectors. Traditionally, these feature vectors were hand-engineered to encode specific structural properties of the graph data \citep[Chapter~2]{hamilton2020graphrepresentationlearning}. But the inflexibility and design expense of hand-engineering has led way to the paradigm of \emph{learning} these representations. \emph{Representation learning} approaches use neural network architectures to learn the mapping from the concepts and relations in the graph to a low-dimensional vector space. This mapping preserves the topological information from the graph as the algebraic operations in the latent feature space. The goal of representation learning algorithms is to optimize the mapping such that the graph structural information to be preserved can be reconstructed from the learned representations or embeddings \citep[Chapter~4]{ma2021deeplearningongraphs}. 

Based on the structural information they preserve, there are different families of graph representation learning algorithms. Among these, methods based on edge reconstruction are state-of-the-art for representation learning on knowledge graphs. The learned representations of concepts and relations, called \emph{Knowledge Graph Embeddings} (KGE) are used for different downstream tasks, like knowledge base completion (or missing link prediction), triple classification, entity disambiguation, relation extraction, question answering, etc. \citep{wang2017kgesurvey, ali2021bringinglightintodark}. \emph{The research in this thesis focuses on the task of missing link prediction in knowledge graphs using KGE models.} This is because the KGE techniques were originally designed for knowledge base completion \citep[Chapter~4]{hamilton2020graphrepresentationlearning}, and because this downstream task is state-of-the-art evaluation for KGE models \citep{ali2021bringinglightintodark, wang2017kgesurvey}. The relevant background on Knowledge Graph Embeddings for the thesis is covered in this chapter. The KGE models, training strategies, and evaluation metrics included in this research are introduced. Additionally, an overview of the design space for KGE models is discussed in Section \ref{sec:kge_mulvsadd} and Section \ref{sec:kge_inductive}; these design intuitions are relevant for the thesis contributions in Chapter \ref{ch:relation_inference}.

While the chapter provides the necessary background for the thesis, a comprehensive survey of all the KGE models published and implemented in the literature is beyond the scope of this thesis. Such surveys are available in \citet{ali2021bringinglightintodark}, \citet{rossi2021kgesurvey} and \citet{wang2017kgesurvey}. Of these, \citet{ali2021bringinglightintodark} is the most recent study that benchmarks the predictive performance of a variety of KGE models for different training strategies and loss functions. \citet{rossi2021kgesurvey} provides a taxonomy for KGE models and investigates the effect of structural graph properties on the predictive performance. \citet{wang2017kgesurvey} presents representative techniques for knowledge graph embeddings, methods to incorporate additional graph information, as well as applications for different downstream tasks. Additionally, Chapter \ref{ch:related_work} of the thesis provides a brief overview of the other families of graph representation learning, which do not scale to multi-relational knowledge graphs. 
On the other hand, it is also noteworthy that graph representation learning is not the only method for missing link prediction in knowledge graphs. Methods based on observable graph features, rule mining and relation path ranking are discussed in \citet{nickel2015review} and \citet{chen2020reasoningonknowledgegraphs}, but are not the focus for this research.

\section{Knowledge Graphs}
\begin{figure}[]
    \centering
    \includegraphics[width=1\textwidth]{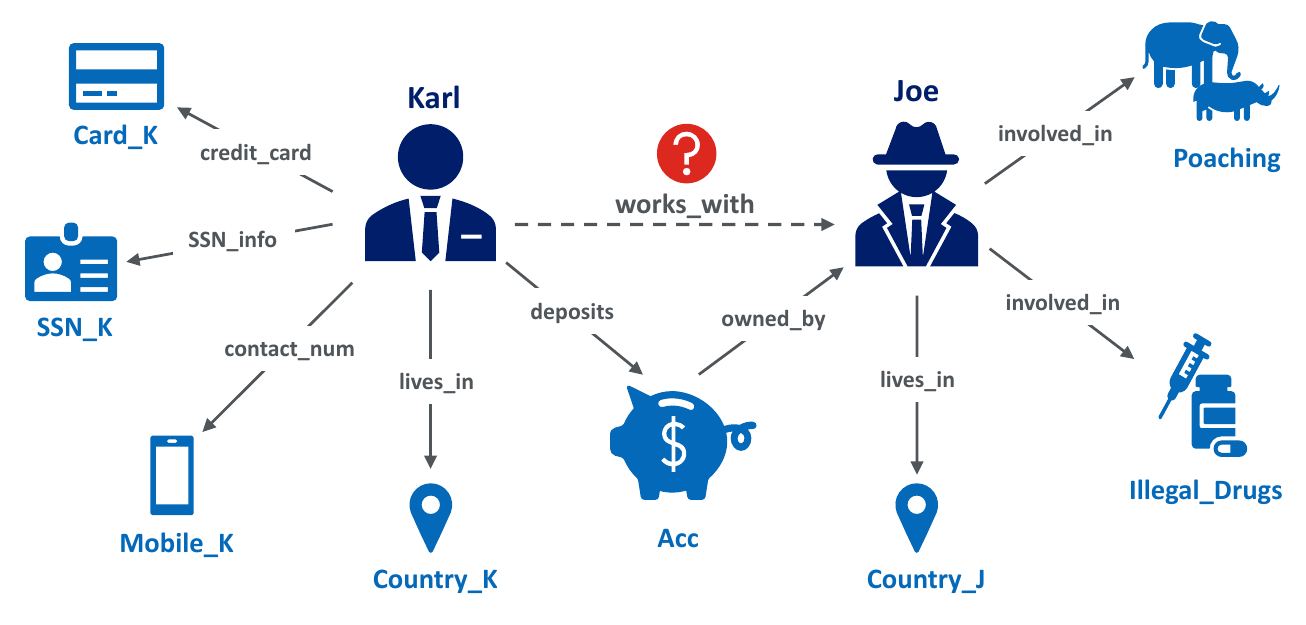}
    \caption{An example Knowledge Graph (KG) of the financial details of a bank's customer $Karl$. Nodes of the graph represent entities and the edge labels represent types of relationships. The existence of an edge between two entities indicates the relationship between them. Knowledge graph embeddings are used to predict the missing link between $Karl$ and $Joe$, given the existing facts in the KG.}
    \label{fig:knowledge_graph}
\end{figure}

A Knowledge Graph (KG) represents the factual knowledge about the world as relationships between entities.
It is a directed, labelled graph where the nodes represent the entities of interest, the edges represent the relationships between them, and the edge labels represent the different types of possible relationships. Based on the domain of interest, the entities might represent various real-world concepts like people, objects, organizations, etc. For example, Figure \ref{fig:knowledge_graph} shows a knowledge graph about the financial details of a bank's customer. The relationships between the entities can then be expressed as factual statements that convey the real-world knowledge about them. 
These factual statements of the form $(subject, \mathtt{relation}, object)$ are called triples. In the example knowledge graph, the triple $(Karl, \mathtt{lives\_in}, Country\_K)$ represents the fact that $Karl$ lives in the country named $Country\_K$. A knowledge graph then is essentially a collection of facts.

Formally, for a set of entities \( \mathcal{E} \) and a set of relations \( \mathcal{R} \), a knowledge graph is a collection of triples represented as \(
 \mathcal{T} \coloneqq \{t \coloneqq (s,\mathtt{r},o) \, |\,  s,o \in \mathcal{E} \, and \, \mathtt{r} \in \mathcal{R} \}
  \), where $s, \mathtt{r}, o$ represent the subject, relation and object in a triple. 
  
All possible triples from the Cartesian product \(\mathcal{E} \times \mathcal{R} \times \mathcal{E}\) can be grouped as a third order tensor. The existence of a triple in the knowledge graph can then be represented through its corresponding entry in the tensor. This alternate tensor representation of a knowledge graph is called an adjacency tensor \citep{nickel2015review}. Formally, ${\tensor{Y} \in {\{0,1\}}^{N_{\mathcal{E}} \times N_{\mathcal{R}} \times N_{\mathcal{E}}}}$ is the adjacency tensor with entries 
\[
y_{ijk} = \begin{cases}
  1,& \text{if the triple } (i,j,k) \text{ exists} \\
  0, & \text{otherwise.}
\end{cases}
\]
  
In this adjacency tensor, $y_{ijk} = 1$ indicates the presence of a true triple in the knowledge graph. However, there are different assumptions for the interpretation of $y_{ijk} = 0$. Under the Closed World Assumption (CWA), all non-existent triples in the knowledge graph are considered false. In this case, the triple $(Karl, \mathtt{works\_with}, Joe)$, which is not a part of the example knowledge graph, indicates that $Karl$ does not work with $Joe$. On the other hand, under the Open World Assumption (OWA), the non-existing triples are considered as unknown and they can either be true or false. This modelling assumption means that the missing links in the knowledge graph are not necessarily false, and can be inferred by using the existing facts.

\section{Knowledge Graph Embeddings}
Knowledge Graph Embeddings (KGE) encode entities and relations as a low-dimensional continuous vector space \(
 \param \coloneqq \{ \mat{E}, \mat{R}\}
  \) where \( \mat{E} \in \mathbb{R}^{k}\) is the embedding matrix for entities, \( \mat{R} \in \mathbb{R}^{k}\) is the embedding matrix for relations and $k$ is the embedding dimension.
  The optimization objective to learn the embeddings aims to reconstruct the relationships between entities as interactions between their latent features. There are many possible ways to model these interactions, which have resulted in different KGE models, characterized by their scoring functions.
  The KGE scoring function \(f : \mathcal{T} \rightarrow \mathbb{R} \) uses the entity and relation embeddings to assign a score to each triple \(f_{t} \coloneqq f(\vec{e}_s, \vec{e}_\mathtt{r}, \vec{e}_o) = f(\vec{s}, \vec{\mathtt{r}}, \vec{o}) \) 
  where \(\vec{e}_s, \vec{e}_o \in \mat{E}\) and \(\vec{e}_\mathtt{r} \in \mat{R}\). 

\begin{figure}[]
    \centering
    \includegraphics[width=1\textwidth]{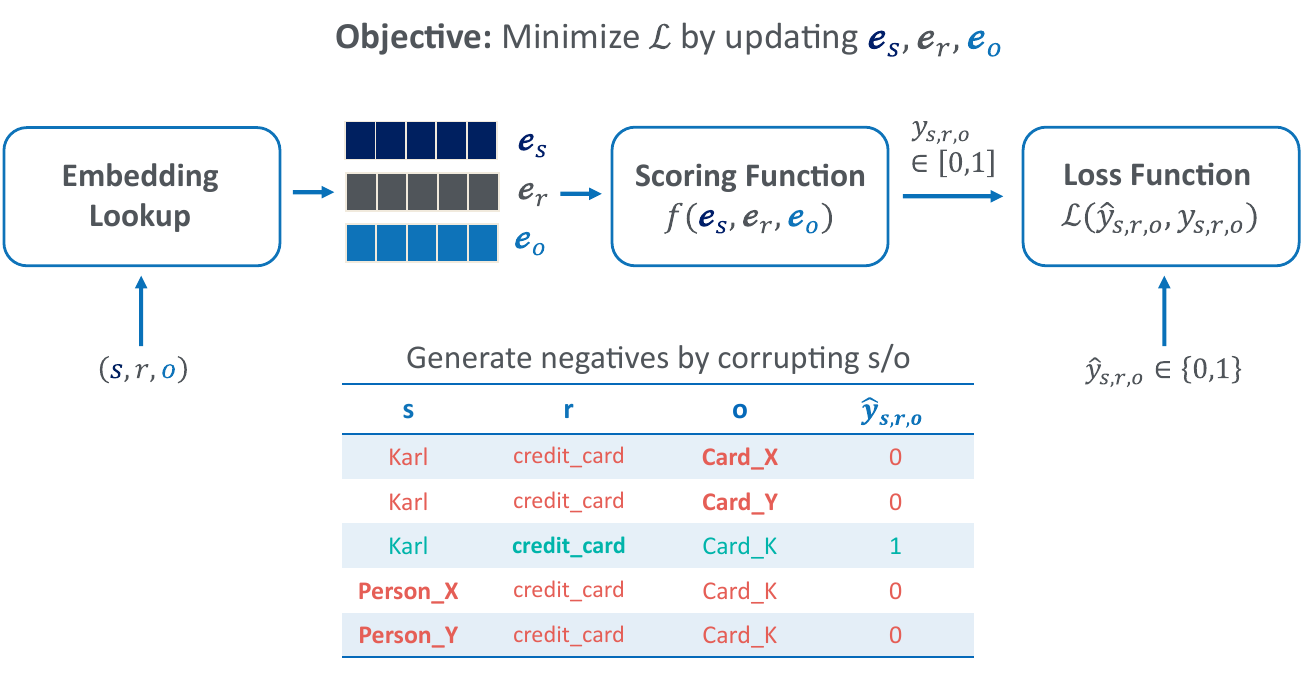}
    \caption{Overall framework for learning Knowledge Graph Embeddings (KGE)}
    \label{fig:kge_flowchart}
\end{figure}

During model training, the embedding matrices are initialized with random values and iteratively updated such that the scores assigned to true (existing) triples in the knowledge graph are higher than the scores for false (non-existing) triples in the knowledge graph. Training this model requires both positive and negative examples, but the knowledge graph only contains positive examples. Thus, negative examples are generated for training by making a Local Closed World Assumption (LCWA), that is, assuming a closed world in the local neighbourhood of an entity. The overall framework for the KGE model training is summarized in Figure \ref{fig:kge_flowchart}, and different components of the pipeline are discussed below. 
  
\section{Scoring Functions}
\label{sec:kge_scoring_functions}
Table \ref{tab:scoring_functions} summarizes the scoring functions of state-of-the-art KGE models studied in this research. Representative models have been selected from the three families of KGE models, namely Tensor Factorization models, Geometric models and Deep-learning based models \citep{rossi2021kgesurvey, wang2017kgesurvey}. These representative KGE models have varying inductive abilities (Section \ref{sec:kge_inductive}) and exhibit multiplicative or additive interactions in the scoring functions (Section \ref{sec:kge_mulvsadd}).

\subsection*{DistMult} 
DistMult \citep{yang2015distmult} models the interactions between entities as a bilinear product of their embeddings. In general, the bilinear product represents the entity embeddings as vectors, the relation embeddings as matrices and combines them through a matrix multiplication. The DistMult model simplifies this parametrization by constraining the relation matrix to be a diagonal matrix. Thus, the scoring function of the DistMult model is a simple Hadamard product of the subject, relation and object embeddings. The Hadamard product is also called a tri-linear dot product and is denoted as $\langle \cdot \rangle$ in the expression below. 
\begin{align*} \label{eq:DistMult_score_fct}
    f(s,\mathtt{r},o) = \vec{s}^{T} \mat{W}_\mathtt{r} \vec{o} = \sum_{i=1}^{d}\vec{s}_i \cdot diag(\mat{W}_\mathtt{r})_i \cdot \vec{o}_i = \langle \vec{s}, \vec{\mathtt{r}}, \vec{o} \rangle \quad 
    where \quad \vec{s}, \vec{\mathtt{r}}, \vec{o} \in \mathbb{R}^k \enspace
\end{align*}

Reducing the relation parameters improves the efficiency of DistMult over earlier models like RESCAL \citep{nickel2011rescal} which used the bilinear product as a scoring function. However, it reduces the expressivity of the model. For example, for the relation $\mathtt {owns}$, $f(Karl,\mathtt {owns}, Account)$ should not be equal to $f(Account,\mathtt {owns}, Karl)$, but DistMult cannot model this relationship because of the commutative nature of the dot product. Yet, a recent study of the reproducibility of KGE models in \citet{kadlec2017kgebaselines} has shown that despite its simplicity, DistMult achieves state-of-the-art predictive performance after proper tuning of the hyperparameters.

\begin{table}
    \centering
    \small
    \begin{tabular}{ l c }
    \toprule
    \multicolumn{1}{c}{\textbf{Model}} &  \textbf{Scoring Function}\\
    \midrule    
         DistMult  & $ \langle \vec{e}_s, \vec{e}_\mathtt{r}, \vec{e}_o \rangle$ \\
         
         ComplEx  & $ \Re(\langle \vec{e}_s, \vec{e}_\mathtt{r}, \overline{\vec{e}_o} \rangle)$  \\
         
         ConvE  & $ \langle \sigma(\mathrm{vec}(\sigma([ \overline{\vec{e}_\mathtt{r}}, \overline{\vec{e}_s}] \ast \boldsymbol{\Omega})) \mat{W}), \vec{e}_o \rangle$ \\
         
         TransE  &  $-\norm{\vec{e}_s + \vec{e}_\mathtt{r} - \vec{e}_o}$    \\
        
    \bottomrule     
    \end{tabular}
    \caption{Scoring functions \(f_{sro}\) of the KGE models used in this research. For ComplEx, $\vec{e}_s, \vec{e}_\mathtt{r}, \vec{e}_o \in \mathbb{C}^k$; for the remaining models $\vec{e}_s, \vec{e}_\mathtt{r}, \vec{e}_o \in \mathbb{R}^k$. Here, $\langle \cdot \rangle$ denotes the tri-linear dot product; $\sigma$ denotes sigmoid activation function, $\ast$ denotes 2D convolution with filters $\boldsymbol{\Omega}$, and $\mat{W}$ denotes additional parameters for ConvE model; $\overline{\ \cdot\ }$ denotes conjugate for complex vectors, and 2D reshaping for real vectors in ConvE model; $\norm{\cdot}$ denotes l-p norm.}
    \label{tab:scoring_functions}
\end{table}

\subsection*{ComplEx}
Similar to DistMult, ComplEx \citep{trouillon2016complex} models the latent features as a tensor factorization of the graph adjacency tensor and constraints the relation embedding matrix to be diagonal. However, the latent features are modelled in a complex vector space instead of the real vector space. Here, the embeddings are composed of a real vector component and an imaginary vector component.
Unlike real-valued vectors, the bilinear product in complex space is not commutative. This is because the bilinear product for complex numbers is defined as a Hermitian product, which uses the complex conjugate of the object embedding instead of the real-valued object embedding. This acts as a form of topological regularization in the latent space and allows ComplEx to model asymmetric relations like $\mathtt {owns}$. 
Thus, the scoring function for a triple is the real valued component of the Hadamard product between the subject embedding, relation embedding and the complex conjugate of object embedding. Using $\langle \cdot \rangle$ to denote the Hadamard product and $\overline{\ \cdot\ }$ for the complex conjugate of a vector, the ComplEx scoring function is expressed as follows.
\begin{equation*}
        f(s,\mathtt{r},o) = 
        \Re(\langle \vec{s}, \vec{\mathtt{r}}, \overline{\vec{o}} \rangle)
        \quad where \quad \vec{s}, \vec{\mathtt{r}}, \vec{o} \in \mathbb{C}^k \enspace %
\end{equation*}
Some more recent models further improve the expressivity of tensor factorization models. For example, QuatE \citep{zhang2019quate} represents the latent features in hypercomplex space; and SimplE \citep{kazemi2018simple} and TuckER \citep{balazevic2019tucker} use further variations of the tensor factorization method like CP decomposition and Tucker decomposition.
However, ComplEx remains the state-of-art model for missing link prediction in terms of the predictive performance \citep{lacroix2018canonical, ruffinelli2020olddognewtricks}.

\subsection*{TransE}
Unlike DistMult and ComplEx, TransE \citep{bordes2013transe} belongs to the family of Geometric models or Translational Distance models. 
These models represent the relationships between entities as geometric transformations in the latent feature space. Specifically, TransE models the object embedding of a triple as a translation from the subject embedding via the relation embedding. In other words, when two entities are related, the embedding of the object is modelled as the embedding of the subject plus the relation embedding. Using $\norm{\cdot}$ to denote the l-p norm of a vector, the TransE scoring function for a triple is expressed as - 
\begin{equation*}
        f(s,\mathtt{r},o) = 
        -\norm{\vec{s} + \vec{\mathtt{r}} - \vec{o}} 
        \quad where \quad \vec{s}, \vec{\mathtt{r}}, \vec{o} \in \mathbb{R}^k \enspace%
\end{equation*}
The model is inspired by word embeddings Word2vec \citep{mikolov2013word2vec} which are learned from free text and coincidentally represent the 1-to-1 relationships between entities as translations in the embedding space. However, the translation operation, explicitly used by TransE, can only model the 1-to-1 relations between entities, which means that TransE cannot handle 1-N, N-1 or N-M relations in the latent space. Despite this limitation, it remains popular because of its computational efficiency and is the representative model for Geometric family of models.

\subsection*{ConvE}
DistMult, ComplEx and TransE score the triples in knowledge graphs through simple pairwise interactions of the latent features. These shallow neural architectures allow them to scale to real-world knowledge graphs with several million triples. However, the learned features have limited expressivity and increasing the embedding size to improve the expressivity does not scale with the number of entities and relations in the knowledge graph. To overcome these challenges, ConvE \citep{dettmers2018conve} adds a convolutional layer to the neural network architecture. To compute the score of a triple, embeddings of subject and relation are reshaped and concatenated; and the resulting matrix is input to the convolutional layer. In this layer, convolutional filters are applied to capture the complex interactions between the subject and relation features. The output feature maps from the convolutional layer are vectorized and projected back into the k-dimensional embedding space. Finally, these embeddings are combined with the object embeddings using the dot product. Using $\langle \cdot \rangle$ to denote the dot product, $\sigma$ to denote the sigmoid activation function, $\ast$ for the 2D convolution operation with filters $\boldsymbol{\Omega}$, and $\overline{\ \cdot\ }$ for 2D reshaping of real vectors, the ConvE score for a triple is given by - 
\begin{equation*}
        f(s,\mathtt{r},o) = 
        \langle \sigma(\mathrm{vec}(\sigma([ \overline{\vec{\mathtt{r}}}, \overline{\vec{s}}] \ast \boldsymbol{\Omega})) \mat{W}), \vec{o} \rangle 
        \quad where \quad \vec{s}, \vec{\mathtt{r}}, \vec{o} \in \mathbb{R}^k \enspace%
\end{equation*}
The convolution operation in ConvE is computationally expensive. Thus, \citet{dettmers2018conve} proposed the 1-K training strategy, which scores each $(s,\mathtt{r})$ pair against all entities simultaneously, instead of scoring the $(s,\mathtt{r},o)$ triples individually. This enables a single pass through the convolution layer for each $(s,\mathtt{r})$ pair, thus scaling the model to real-world knowledge graphs.

\section{Characteristics of Scoring Functions}

\subsection{Multiplicative vs Additive Interactions}
\label{sec:kge_mulvsadd}

The scoring functions of KGE models exhibit multiplicative or additive interactions \citep{chandrahas2018towards}. The multiplicative models score triples through multiplicative interactions of subject, relation and object embeddings. The scoring function for these models can be expressed as $f_{sro} = \vec{e}_{\mathtt{r}}^\top \mathcal{F}(\vec{e}_s, \vec{e}_o)$ where the function $\mathcal{F}$ measures the compatibility between the subject and object embeddings and varies across different models within this family. DistMult, ComplEx and ConvE have such interactions. On the other hand, additive models score triples through additive interactions of subject, relation and object embeddings. The scoring function for such models can be expressed as $f_{sro} = -\norm{\mat{M}_{\mathtt{r}}^1(\vec{e}_s) + \vec{e}_\mathtt{r} - \mat{M}_{\mathtt{r}}^2(\vec{e}_o)}$ where \( \vec{e}_s, \vec{e}_o \in \mathbb{R}^{k_\mathcal{E}} \), \(\vec{e}_\mathtt{r} \in \mathbb{R}^{k_\mathcal{R}} \) and $\mat{M}_\mathtt{r} \in \mathbb{R}^{k_{\mathcal{E}} \times k_{\mathcal{R}}}$ is the projection matrix from entity space $\mathbb{R}^{k_{\mathcal{E}}}$ to relation space $\mathbb{R}^{k_{\mathcal{R}}}$. TransE has additive interactions. 

\subsection{Inductive Abilities of Knowledge Graph Embeddings}
\label{sec:kge_inductive}
Complementary to multiplicative and additive interactions, another way to characterize the scoring functions of KGE models is based on their ability to represent different logical patterns on the relations \citep[Chapter~4]{trouillon2019inductive, hamilton2020graphrepresentationlearning}. 
This is because the general intuition behind the design of the scoring functions is to capture logical properties between relations from the observed facts in the knowledge graph. These logical properties can then be used to generalize to unseen triples and make downstream inferences about the relationships between entities. For example, the relation $ \mathtt{is\_owned\_by}$ is inverse of the relation $ \mathtt {owns}$, and when the fact $(Account,\mathtt{is\_owned\_by}, Karl)$ is true, then the fact $(Karl,\mathtt {owns}, Account)$ is also true and vice versa. A KGE model that can capture the inversion pattern can thus predict missing facts about $\mathtt {owns}$ based on the observed facts about $\mathtt{is\_owned\_by}$.

In this research, the representational abilities of KGE models are referred to as \emph{relation inference patterns}. The most studied relation inference patterns in the current literature are symmetry, inversion and composition since they occur very frequently in real-world knowledge graphs \citep{hamilton2020graphrepresentationlearning, ali2021bringinglightintodark}. For a directed graph, the symmetry pattern indicates that if a relation exists in one direction, it also exists in the opposite direction. Similarly, the inversion pattern indicates that the existence of one relation implies the existence of another with opposite direction. The composition pattern indicates that a relation can be expressed as a composition of two or more other relations. These relation inference patterns are discussed in further detail in Chapter \ref{ch:relation_inference} where they are used to investigate the adversarial vulnerability of KGE models. 

It is noteworthy that the KGE models are latent factor models and the logical patterns are not explicitly encoded in the learning algorithm. Rather, the relation inference patterns are a means to understand the inductive abilities of the models, and reflect the inductive biases that the model might have captured at training time to generalize at inference time.

\section{Training Knowledge Graph Embeddings}

Since the KGs only contain positive triples; to train the KGE model, synthetic negative samples \(\negative{t} \in \negative{\mathcal{T}}\) are generated by replacing the subject or object in the positive triples with other entities in $\mathcal{E}$. That is, for each positive triple $t \coloneqq (s,\mathtt{r},o)$, the set of negative samples is $ \negative{t} \coloneqq \{(\negative{s},\mathtt{r},o) \cup (s,\mathtt{r}, \negative{o})\}$. 

This approach for negative sample generation relies on the Local Closed World Assumption (LCWA) about the knowledge graph, where the graph is considered only locally complete \citep{nickel2015review}. This means that if the entity-relation pair $(s,\mathtt{r})$ has an existing triple $(s,\mathtt{r},o)$ in the knowledge graph, then any non-existing triple $(s,\mathtt{r},\negative{o})$ is treated as false.

The training objective is to learn the embeddings that score positive triples existing in the KG higher than the negative triples generated synthetically. To achieve this, a triple-wise loss function is minimized. This loss function is expressed as
\begin{equation*}
    \loss(t,\param) \coloneqq \quad \ell(t,\param) + \sum_{\negative{t} \in \negative{\mathcal{T}}} \ell(\negative{t}, \param) \quad where \quad  \param \coloneqq \{ \mat{E}, \mat{R}\} \enspace
\end{equation*}
Thus, the optimal parameters \(\widehat \param\) learned by the model are defined by 
\(
  \widehat{\param} \coloneqq \argmin_{\param} \sum_{t \in \mathcal{T}} \loss(t, \param)
  \).

\subsection{Training Strategies}
There are different strategies for training the KGE models based on the differences in the way negative triples are generated \citep{ruffinelli2020olddognewtricks}. 
Using all possible entity corruptions of the training triples can be computationally expensive. Thus, a common approach is to sample some negative triples from the set of all corruptions. Theoretically, the negative corruptions that are false negatives (that is, if the corruption already exists in the graph) should be excluded from the training process. However, in practice, the knowledge graphs are sparse and the likelihood of sampling an actual triple as negative is low. Thus, the additional filter step is omitted for efficiency \citep{ali2021bringinglightintodark}. This training strategy is referred to as negative sampling or \emph{NegSamp}, and was proposed by \citet{bordes2013transe}.
An alternative approach for training KGE models was proposed in \citet{lacroix2018canonical}. Instead of sampling from the negative corruptions, the entire set of corruptions is used for training. This approach is called \emph{1vsAll} or \emph{1-N} training strategy. Though it is generally expensive, it is feasible for knowledge graphs where the number of entities is not excessively large \citep{ruffinelli2020olddognewtricks}. Lastly, the approach proposed by \citet{dettmers2018conve} is called \emph{KvsAll} or \emph{1-K} training strategy. Instead of generating negative corruptions from individual triples, the approach generates corruptions from entity-relation pairs. In this case, the possible triples from the pair $(s,\mathtt{r})$ are treated as positive if they exist in the training knowledge graph and negative otherwise.

\subsection{Loss Functions}
Several loss functions have also been introduced for training KGE models. Given the model parameters $\param$ (that is, embeddings) at a training step, the score for a triple is computed using the scoring functions discussed earlier. The loss functions then compute the error in the scores assigned by the model. This error is computed either from the independent labels for each triple; or by comparing the scores assigned to positive and negative triples; or by comparing the positive triple's score with a set of negative triples. Based on this, the loss functions can be categorized as pointwise, pairwise and setwise loss functions \citep{ali2021bringinglightintodark, mohamed2019kgelossfunctions}.

\subsubsection*{Pointwise Loss Functions}
Pointwise loss functions compute an independent loss term for each triple-label pair.
The representative loss function for pointwise losses is Binary Cross Entropy (BCE) loss. It uses the sigmoid activation function on a triple's score to compute the probability of that triple. The error term is then the cross entropy between the resulting probability and the triple's label. Let $t$ denote a triple (positive or negative example) and $\widehat{y_t} \in \{0,1\}$ denote the label of the triple. Then, the BCE loss term is expressed as - 
\begin{align*}
\loss(t,\param) \coloneqq  -(\widehat{y_t} \cdot \log(\sigma(f_t)) + (1 - \widehat{y_t}) \cdot \log(1 - \sigma(f_t)))
\end{align*}

BCE loss thus, frames the learning problem as a binary classification of the triples, where the model outputs are treated as logits. It is suitable for multi-class and multi-label classification and is usually used with the KvsAll training strategy.

\subsubsection*{Pairwise Loss Functions}
The pairwise loss functions compare the scores for positive triples $t$ with the scores of negative triples $\negative{t}$. This is based on the Open World Assumption (OWA) where the negative triples are considered "less positive" instead of negative. The representative loss function for pairwise losses is the margin based ranking loss introduced in \citet{bordes2013transe}. The loss is applicable to the NegSamp and the 1vsAll training strategies.

Let $\Delta \coloneqq f_t - f_{\negative{t}}$ denote the difference between the scores of positive and negative triples. Then, the expression for pairwise hinge loss or the pairwise max-margin loss is given as \(\loss(t,\param) \coloneqq \max(0, \lambda + \Delta) \enspace \). Here, $\lambda$ denotes the margin parameter for the difference between the scores.

\subsubsection*{Setwise Loss Functions}
Instead of relying on the scores of individual triples or the pairs of triples, the setwise loss functions compare the positive triple's score with a set of negative triples. 
The representative function for computing setwise loss is the Cross Entropy (CE) loss. This loss computes the cross entropy between the model's probability distribution and the data distribution of labels. The model distribution is computed as a normalized softmax distribution over the scores assigned by the model. For a positive triple $t \in \{(s,\mathtt{r},o)\}$ and all possible object-sided negative corruptions $\negative{t} \in \{(s,\mathtt{r},\negative{o})\}$, the softmax normalization for predicted scores is given by - 
\begin{equation*}
     p(o \mid s, \mathtt{r}) = \frac{\exp(f(s, \mathtt{r}, o))}{\sum\limits_{\negative{o} \in \mathcal{E}}\exp(f(s, \mathtt{r}, \negative{o}))} \enspace
\end{equation*}

The data distribution is obtained by normalizing the label values of triples to sum to 1. Finally, the cross entropy between the model and label distribution is -  
\begin{equation*}
    \loss(t,\param) \coloneqq - \sum\limits_{\negative{o} \in \mathcal{E}} \mathbb{I}[(s,\mathtt{r},\negative{o}) \in \mathcal{T}] \cdot \log(p(o\mid s,\mathtt{r})) \enspace
\end{equation*}

Here, $\mathbb{I}$ denotes the indicator function which indicates whether the negative triple already exists in the knowledge graph or not.
Since the loss applies a softmax normalization, it is suitable for multi-class and single-label classification problems. It is thus, usually used with NegSamp and 1vsAll training strategies.

\section{Missing Link Prediction}
Large-scale knowledge graphs which serve as background knowledge for knowledge-aware applications are extracted automatically from unstructured textual sources on the Web. Despite advances in information extraction techniques, the extracted graphs are often incomplete and noisy. Thus, predicting the missing facts in knowledge graphs has emerged as an important ML task on knowledge graphs \citep{nickel2015review, rossi2021kgesurvey}. Representation learning on knowledge graphs is the state-of-the-art solution for this task. 
The learned representations of concepts and relations, called \emph{Knowledge Graph Embeddings} (KGE) are used to score candidate missing facts and select the most plausible one.

The task of knowledge base completion (or missing link prediction) using KGE models is the focus of this research. This task is formulated as an entity ranking problem. Given the entity-relation pairs of the form $(s,\mathtt{r})$ or $(\mathtt{r}, o)$, the task is to predict the corresponding object or subject entity. This amounts to answering questions of the form $(s,\mathtt{r}, ?)$ or $(?, \mathtt{r}, o)$. For example, from the knowledge graph in Figure \ref{fig:knowledge_graph}, to predict whether $Karl$ works with $Joe$, KGE models are used to answer the query $(Karl,\mathtt{works\_with}, ?)$. The evaluation protocol for missing link prediction then measures the capability of the link predictor to answer these queries correctly \citep{ruffinelli2020olddognewtricks, ali2021bringinglightintodark}.

\subsection{Evaluation Protocol}
\label{sec:kge_eval_protocol}
The available triples in the knowledge graph are split into training, validation and test subsets. Evaluation of KGE models is performed by training the embeddings on the training subset and using the learned embeddings to correctly predict the triples in the test subset. Similar to the training process, the test subset does not contain negative examples either. Thus, for each test triple $t=(s,\mathtt{r},o)$, subject-side corruptions $\negative{t_s}={(\negative{s},\mathtt{r},o)}$ and object-side corruptions $\negative{t_o}={(s,\mathtt{r},\negative{o})}$ are generated by replacing the subject/object entity with each entity from \( \mathcal{E} \). The KGE model's scoring function is used to predict scores of the original as well as negative triples. The scores are then sorted in descending order and the rank of the correct test entity is determined. These steps provide two ranks for the test triple - one for subject corruptions, and one for object corruptions. A better KGE model assigns better ranks to the actual entity relative to the corruptions. Thus, the entity ranking task evaluates the model's capability to differentiate between the actual test triples and their corruptions.

While computing the ranks, it is possible that some of the corruptions are valid answers to the query. This happens if the corruptions of the test triple already exist as triples in the training knowledge graph. To avoid the distortion of results due to these false negatives, the filtered evaluation setting was proposed in \citet{bordes2013transe}. In this setting, negative triples that already exist in the training, validation or test set are filtered out. That is, their scores are ignored while computing the ranks. 
Depending on the domain of use, either the subject or object or both ranks of the test triple are used to determine the KGE model's confidence in predicting the missing link.

\subsection{Evaluation Metrics}

Given the ranks computed for the test triples, global summary metrics are used to benchmark the predictive performance of different KGE models. Let the set of triples in the test set be denoted by $\mathcal{T}_{test}$ and the rank of each triple be $rank(t)$. Then, following are the state-of-the-art evaluation metrics for missing link prediction with KGE models \citep{ali2021bringinglightintodark}.

\emph{Mean Rank (MR)} is the mean of the ranks predicted by the KGE model. The value of this metric ranges between $1$ and $|\mathcal{E}|$, and the smaller values indicate better predictive performance. Due to its sensitivity to outliers, the metric is not used widely. It is defined as - 
\begin{equation*}
    \text{MR} = \frac{1}{|\mathcal{T}_{test}|} \sum \limits_{t \in \mathcal{T}_{test}} rank(t)
\end{equation*}

\emph{Mean Reciprocal Rank (MRR)} is the mean of the reciprocal values of the predicted ranks. The value ranges between $0$ and $1$ and the higher values indicate better predictive performance. MRR is defined as - 
\begin{equation*}
\text{MRR} = \frac{1}{|\mathcal{T}_{test}|} \sum\limits_{t \in \mathcal{T}_{test}} \frac{1}{rank(t)}
\end{equation*}

\emph{Hits@N} counts the proportion of correct entities that have been ranked less than or equal to N. That is, it measures the fraction of correct entities ranked in the top-N by the KGE model. Based on the measurement granularity of interest, the value of N can be $1,3,5,10$ etc. Higher values of Hits@N indicate better predictive performance. The metric is defined as - 
\begin{equation*}
    \text{Hits@N} = \frac{|\{t \in \mathcal{T}_{test} \mid rank(t) \leq N\}|}{|\mathcal{T}_{test}|}
\end{equation*}

\subsection{Benchmark Datasets}
\label{sec:kge_datasets}
To evaluate the performance of KGE models for missing link prediction, several benchmark datasets have been created over the years \citep{rossi2021kgesurvey, ali2021bringinglightintodark}. These benchmarks have been sampled from real-world knowledge graphs, instead of being built from scratch. The research in this thesis uses the benchmark datasets WN18RR \citep{dettmers2018conve} and FB15k-237 \citep{toutanova2015observed} that have been built from the Wordnet and Freebase knowledge graphs.

\subsubsection*{WordNet}
Wordnet\footnote{\url{https://wordnet.princeton.edu/}} is a lexical database of English, designed as a structured representation of the language thesaurus for Computational Linguistics and NLP tasks \citep{miller1995wordnet}. Semantically similar words that can be used interchangeably are grouped together as concepts called synsets. The word forms (strings of letters representing words) and synsets are represented as nodes of the graph. The labelled edges represent the multiple semantic and lexical relationships between the word forms and their meanings.

A subset of the Wordnet knowledge base was extracted by \citet{bordes2014semanticmatching} to evaluate KGE models for missing link prediction. Triples containing a pre-selected set of relations were selected. Of these, triples containing entities appearing in less than 15 triples were filtered out. This subset is called WN18 and contains 18 relations between 40,943 entities. \citet{dettmers2018conve} observed that some facts in the test set of WN18 can be predicted trivially because of an existing inverse relation between the entities in the training set. To avoid this test set leakage, WN18RR was created by removing the facts about relations that have an inverse relation in the knowledge graph. This dataset contains 40,943 entities and 11 relations, and is now the benchmark Wordnet subset to evaluate KGE for missing link prediction.

\subsubsection*{Freebase}
Unlike Wordnet, Freebase \citep{bollacker2008freebase} is an encyclopedic knowledge graph, initially built in the Semantic Web community as a structured representation of general human knowledge. Thus, any real-world concept like person, place, profession or sport can be assigned an identifier and represented as a node in the graph. Factual statements about these entities are then modelled as the $(subject, \mathtt{relation}, object)$ triples. To support relations between more than two entities, intermediate nodes called Compound Vale Type (CVT) nodes are used for reification \citep{tanon2016freebasetowikidata, rossi2021kgesurvey}. The dataset was built collaboratively with contributions from volunteers on the Web. Though the Freebase API\footnote{\url{https://developers.google.com/freebase}} has now been migrated to Wikidata \citep{tanon2016freebasetowikidata}, the benchmark subsets extracted from the original dataset continue to be used.

\citet{bordes2013transe} extracted a subset of Freebase for KGE model evaluation. Entities with at least 100 mentions and also appearing in Wikilinks database were selected. Facts about these entities were extracted, and the reified relations from the original graph were converted into cliques with binary edges \citep[Section~4.3.4]{rossi2021kgesurvey}. The dataset, called FB15k has 14,951 entities and 1,345 relationships. However, \citet{toutanova2015observed} observed test set leakage for this dataset, that is, entities in some test set triples are connected directly in the training set. A simple model based on observable features could thus achieve state-of-the-art performance for missing link prediction. The study created a subset of FB15k by selecting facts with 401 most frequent relations and filtering out the facts with equivalent or inverse relations. This dataset called FB15k-237 contains 14,541 entities and 237 relations.

\begin{table}
 \centering
 \small
 \setlength{\tabcolsep}{7pt}
\begin{tabular}{c  l ll}
    \toprule           
    \multicolumn{2}{l}{} & \textbf{WN18RR} &  \textbf{FB15k-237} \\ 
    \midrule
    \multicolumn{2}{l}{Entities}                  &  40,559   & 14,505 \\ 
    \multicolumn{2}{l}{Relations}                 &  11       & 237 \\ 
    \multirow{3}{*}{\shortstack[l]{Triples} }
        & Training      & 86,835    &  272,115    \\
        & Validation    &  2,824    &  17,526    \\
        & Test          &   2,924   &  20,438    \\
    \bottomrule
\end{tabular}
\caption{Statistics for benchmark datasets WN18RR and FB15k-237 used in the thesis.}
\label{tab:kge_data}
\end{table}

It is noteworthy that some of the earliest methods for Knowledge Graph Embeddings were evaluated on the subsets of Wordnet and Freebase \citep{bordes2013transe, bordes2014semanticmatching}. Though several new benchmarks have been introduced, these remain state-of-the-art datasets for evaluating KGE models. Additionally, the two knowledge bases represent knowledge from two different domains and were constructed using different methodologies. While Wordnet is a lexical ontology that was curated by NLP experts, Freebase is a collection of cross-domain general-purpose facts that were curated by crowdsourcing on the Web. Thus, these two knowledge graphs are suitable for evaluating the research contributions of this thesis.
As discussed later in Section \ref{sec:problem_evalprotocol}, the thesis uses WN18RR and FB15k-237 for KGE training and evaluation, as well as for the evaluation of the proposed adversarial attacks. Table \ref{tab:kge_data} summarizes the statistics about these datasets.

\section{Summary}
This chapter introduced the background concepts for representation learning on knowledge graphs. The overall framework of KGE models, as well as its relevant components were discussed. The task of missing link prediction on knowledge graphs was also introduced. Due to the effectiveness of KGE models for this task, they are now increasingly used in high-impact domains like healthcare and finance, where model outcomes affect people's lives \citep{rotmensch2017healthkgelectronicmedicalrecords, mohamed2020knowledgegraphdrugdiscovery, bonner2021knowledgegraphdrugdiscovery}. %
In these high-stakes domains there are likely going to be actors that want to cause KGE model failure. However, the security vulnerabilities of KGE models to adversarial manipulation are largely unknown. The next chapter introduces and formulates the problem statement for examining the adversarial vulnerabilities of KGE models.

\chapter{Problem Formulation}
\label{ch:problem_formulation}

Machine Learning (ML) is often contended as critical for building intelligent systems that can operate in complex realistic scenarios \citep{goodfellow2016deeplearning}. 
With the availability of huge volumes of data, ML systems have indeed enabled the automation of many real-world tasks. 
This is also true for reasoning with knowledge graphs. As discussed in \citet{chen2020reasoningonknowledgegraphs} and \citet{nickel2015review}, latent distributed representations (that is, KGE models) have replaced earlier methods based on logic rules and observable graph features for knowledge graph reasoning. 
Due to their impressive predictive abilities, the learned representations are increasingly used for user-facing applications in complex safety-critical domains like healthcare and finance \citep{mohamed2020knowledgegraphdrugdiscovery, bonner2021knowledgegraphdrugdiscovery}. 
This success is due to the ability of representation learning algorithms to encode patterns from the raw input facts in the knowledge graph as latent feature representations for entities and relations.

However, the adaptability of ML models to input data is also a security vulnerability \citep{barreno2010security, joseph_nelson_rubinstein_tygar_2019}. Malicious actors that want to manipulate the predictions of KGE models can add or remove carefully and intentionally crafted facts to the input knowledge graph. Since the KGE models are non-interpretable, the impact of malicious edits on their predictions, that is, their failure modes are unknown. This is especially concerning for user-facing applications in high-stakes domains where the cost for model failure is very high. Thus, identifying the security vulnerabilities of KGE models is critical to deploy these models in real-world settings that they are designed for. The security vulnerabilities of ML systems have been investigated in the research field of Adversarial Machine Learning \citep{biggio2018wild, joseph_nelson_rubinstein_tygar_2019}. This chapter introduces the relevant concepts from this field of research and formulates the problem statement for examining the security vulnerabilities of KGE models. 

\section{Adversarial Attacks}
Machine Learning systems are vulnerable to attacks in adversarial settings, i.e. an adversary can manipulate the results of a learning system by carefully perturbing the inputs \citep{barreno2010security}. Thus, successful deployment of ML systems requires that their security is evaluated during system design. Several studies, \citet{joseph_nelson_rubinstein_tygar_2019, biggio2018wild, liu2018survey, carlini2019evaluating}, have been unanimous in their approach to evaluating the security of learning systems. The inspiration is from a well established practice in the field of Computer Security, which is to determine the attacks that can happen on a learning system, then evaluate the resilience of the system by simulating these attacks, and then propose defenses to strengthen the system against these attacks \citep{joseph_nelson_rubinstein_tygar_2019}. This approach for ML security is based on the underlying principle of proactive security design. According to this design approach, the system designer should anticipate adversarial attacks on the learning system and incorporate defenses against these attacks during system design. The approach is in contrast to a reactive security design approach, in which the defense is proposed in response to an actual attack by an adversary. Thus, designing adversarial attacks against ML models is essential for a proactive and proper security evaluation of these models \citep{biggio2018wild}.  

Adversarial attacks are \emph{empirical} methods to investigate the security of Machine Learning systems\footnote{In contrast to adversarial attacks, \emph{certificates} of adversarial robustness provide theoretical guarantees for the predictions of an ML algorithm for bounds on input perturbations.}. These methods craft systematic and intentional perturbations to the ML model's input data with the aim to perturb the model's predictions. The key design attribute of adversarial attacks is to simulate the data distribution of real-world deployment scenarios where the training and test samples might not belong to the same empirical distribution \citep{joseph_nelson_rubinstein_tygar_2019}. The stability of the predictive performance of ML models against adversarial attacks indicates their adversarial robustness. Therefore, designing adversarial attacks helps to anticipate the potential security vulnerabilities of an ML model before it is deployed in a real-world (and potentially adversarial) setting.

The aim of adversarial attacks is to perturb the predictions of an ML model by making perturbations to the input data of the model. A taxonomy for adversarial attacks was first proposed in \citet{barreno2010security} and later extended in \citet{huang2011adversarialmachinelearning}. The security threats can be categorized based on three different perspectives - the extent of their influence on the learning system, the security violation they cause, and the specificity of the security threats. 
\begin{enumerate}
    \item \emph{Influence -- }Adversarial attacks can be causative or exploratory. Causative attacks influence the model's training phase by manipulating the training data. On the other hand, exploratory attacks can only influence the inference phase by manipulating the test data. These two types of attacks are more commonly known as data poisoning and evasion attacks.
    \item \emph{Security Violation -- }Adversarial attacks can violate the ML system's integrity by generating false negatives and hence, evading detection without compromising normal system operation; or they can violate availability by generating false positives and hence, compromising the system functionality available to legitimate users; or they can violate privacy by stealing private information from the system.
    \item \emph{Specificity -- }Adversarial attacks can be targeted to misclassify specific examples, or they can be indiscriminate and misclassify any sample.
\end{enumerate}

In general, an adversarial attack can be conducted at training time or inference time; can have restricted or full access to the learned model's parameters; and can make noticeable or unnoticeable perturbations to the input data. These different settings can result in a spectrum of attacks for an ML model. However, the adversarial attacks designed for one setting may not be effective for another setting. 
Hence, for every adversarial attack, an attack threat model provides a specification of the setting for which that attack is designed. This includes the attacker's goals, their knowledge of the ML system and their capabilities to make perturbations to the system's input data. 

\section{Threat Model for Adversarial Attacks}
\label{sec:problem_threatmodel}

A formal threat model is used to quantify the degree of security needed for a learning system by realistically modeling an adversary \citep{joseph_nelson_rubinstein_tygar_2019, biggio2018wild, carlini2019evaluating}. It specifies the conditions and assumptions for which the adversarial attack is designed against the learning system.
Modeling the threat for an ML system involves defining the adversarial attacker's goals, their knowledge of the system and their capabilities. 

The attack threat model is an essential component of proactive security evaluation, as it enables the system designer to envision different attack scenarios. Corresponding attack strategies can then be proposed and implemented for specific attack scenarios. By specifying the assumptions for adversarial attacks, the attack threat model also determines the computational tractability of designing the attack strategies.
It is noteworthy that there is usually a trade-off between the feasibility of threat models for realistic scenarios and the tractability of adversarial attacks \citep{joseph_nelson_rubinstein_tygar_2019, biggio2018wild}. Specifically, the more restricted an attacker's knowledge and capabilities are, the more difficult is the design of adversarial attacks for that attack threat model.

Following is the specification of the threat model for designing the adversarial attacks in this research.
While this attack threat model is applicable to the proposed attacks in both Chapter \ref{ch:instance_attribution} and \ref{ch:relation_inference}, it is further specialized for the attacks proposed in each chapter. Additionally, it closely follows the attack threat model from state-of-the-art adversarial attacks against KGE models \citep{zhang2019kgeattack, pezeshkpour2019criage, lawrence2021gradientrollback}, which are discussed in Section \ref{sec:sota_kgeattack}.

\subsection{Attacker's Goal}
The adversarial attacker's goal is defined using the second and third dimensions of the attack taxonomy. It specifies the security violation, attack specificity and error specificity that an attacker desires to achieve. The attacker may only want to evade detection (integrity attack) or compromise system functionality available to all the users (availability attack) or steal private information from the system (privacy attack). In terms of the attack specificity, the attacker may want to misclassify specific samples (targeted attack) or any sample (indiscriminate attack). Similarly, the attacker may want a sample to be misclassified as a specific class (specific) or as any class other than the true class (generic) \citep{biggio2018wild}.

This thesis studies the targeted integrity attacks against KGE models for the task of knowledge base completion. The attacker aims to misclassify a specific set of missing triples instead of compromising the overall model performance. Additionally, the attacker aims to misclassify the facts that are predicted True by the victim model, instead of predicting additional missing facts as True. This attack setting is realistic for the scenarios where the KGE models are deployed. For example, if a KGE model is used to predict insurance fraud in a financial network, insurance fraudsters might want to misclassify specific facts that are predicted True by the victim model.

Let the notation $z \coloneqq (z_s,z_\mathtt{r},z_o)$ denote the missing \emph{target triple} which is predicted highly plausible by the victim KGE model, that is, assigned a high rank. In this case, $z_s,z_o$ are the \emph{target entities} and $z_\mathtt{r}$ is the \emph{target relation}. The goal of an adversarial attacker then is to \emph{degrade} the predicted ranks of this triple\footnote{This is in contrast to the setting where the attacker would aim to improve the rank of low-ranked triples.}.

\subsection{Attacker's Knowledge}
An attack threat model should define the extent of knowledge that an attacker has about the target ML system. This includes the knowledge about system's training data, feature set, learning algorithm and the trained model parameters \citep{biggio2018wild}. Based on the attacker's knowledge, one can model different attack scenarios --  white-box attacks where the attacker has full knowledge, or black-box attacks where the attacker has no knowledge, or gray-box attacks where the attacker has limited knowledge of the target system \citep{carlini2019evaluating, barreno2010security}.

While black-box settings are more realistic, attacks in white-box settings provide a worst case security evaluation of the ML system. According to Kerckhoffs' principle for Computer Security, the system security should not rely on expectations of secrecy \citep{joseph_nelson_rubinstein_tygar_2019}. %
Thus, to ensure reliable vulnerability analysis, this thesis uses a white-box attack setting. It is assumed that the attacker has full knowledge of the victim KGE model architecture as well as access to the learned embeddings.

\subsection{Attacker's Capability}
A threat model should define the influence that an attacker has on the input data and any reasonable constraints that need to be imposed on the attacker \citep{biggio2018wild}. The two most common attacks in the attack taxonomy are based on the attack influence - causative attacks that can manipulate both the training and test data, and exploratory attacks that can manipulate test data only. These are more commonly referred to as the training time attacks (or data poisoning attacks), and the test time attacks (or evasion attacks) respectively.

To predict scores for the triples at inference time, state-of-the-art KGE models use only the latent representations learned from the training graph. This means that the perturbations to the graph structure at inference time would not affect the scores predicted for the missing triples. Thus, the evasion attacks at inference time are not realistic for KGE models.
This thesis focuses on the data poisoning attacks against KGE models. Data poisoning attacks aim to perturb the predictions of the KGE model at inference time by perturbing the input data at training time.
The attacker cannot manipulate the model architecture or the learned embeddings directly; but only through perturbations to the training knowledge graph.

\begin{figure}
    \centering
    \includegraphics[width=1\textwidth]{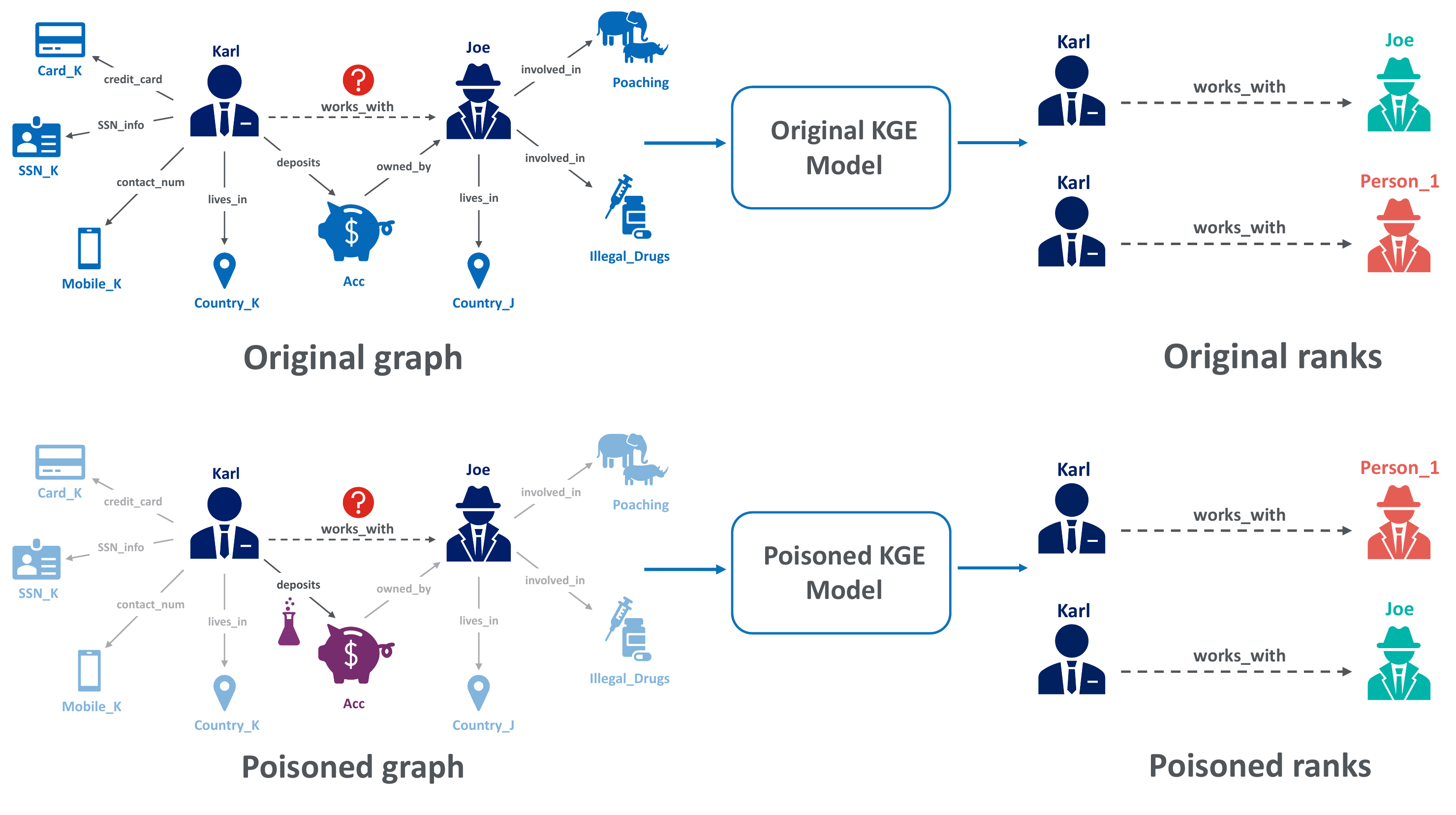}
    \caption{Threat model for the adversarial attacks in this research. The original KGE model assigns a high rank to the target fact $(Karl, works\_with, Joe)$. However, deleting the triple $(Karl, deposits, Acc)$ from the original knowledge graph, and training a KGE model on the perturbed graph degrades the ranks of the target fact $(Karl, works\_with, Joe)$. The change in ranks predicted by the original KGE model and the poisoned KGE model indicates the effectiveness of the adversarial attack.} 
    \label{fig:problem_threatmodel}
\end{figure}

\bigskip
\noindent 
To summarize, this thesis assumes an adversarial attacker that targets the missing facts which have been assigned high ranks by a KGE model. The attacker aims to design perturbations to the knowledge graph that degrade the target triples' ranks. This means that the target triples' ranks predicted by a KGE model trained on the perturbed graph should be worse than the original model's predictions. Figure \ref{fig:problem_threatmodel} illustrates this attack threat model.

\section{Design Space for Adversarial Attacks}
\label{sec:problem_bilevel}

Given the adversarial attacker's goal, knowledge and capability, how can we design an adversarial attack for this threat model?
In general, if the victim model's parameters are $\param$ and the attacker is capable of perturbing the original dataset $\set G$ to $\adversarial{\set G} \in \Phi$, then the adversarial goal can be characterized as an objective function $\atk(\adversarial{\set{G}}, \param)$. Here, $\Phi$ is the set of valid perturbations on $\set G$, and the objective function $\atk$ measures the effectiveness of the attack.
The optimal strategy to design an adversarial attack then, is to identify the perturbations that would maximize this adversarial objective $\atk$. That is, the perturbed dataset is given by the solution of the expression 
\(
    \argmax_{\adversarial{\set{G}} \in \Phi} \quad
    \atk(\adversarial{\set{G}}, \param)
\) \citep{biggio2018wild}.

For data poisoning attacks, the perturbations are staged at the training phase. Thus, the victim model parameters $\param$ are not static and are constrained to be the optimal solution of model training on the perturbed data.
In this setting, the overall attack strategy is formally expressed as a bilevel optimization problem -
\begin{equation*}
    \widehat{ \set{G}} \coloneqq \quad \argmax_{\adversarial{\set{G}} \in \Phi} \quad
    \atk(\adversarial{\set{G}}, \widehat \param) \quad s.t. \quad
    \widehat{\param} \coloneqq \argmin_{\param} \quad \trainloss(\param, \adversarial{\set G})
\end{equation*}
Here, the inner optimization corresponds to learning a model $\widehat{\param}$ on the perturbed dataset $\adversarial{\set G}$ and the outer optimization problem corresponds to evaluating the impact of these perturbations on the target data through the adversarial objective $\atk$. When the attacker's goal is to degrade the predictive performance on specific targets $\set Z$, the adversarial objective can be defined through a loss function over these targets. That is, 
\(
    \atk(\adversarial{\set{G}}, \widehat \param) \coloneqq \atkloss(\set{Z}, \widehat \param)
\)
is used as a measure of the effectiveness of the perturbations on the targets. This loss depends on the perturbed dataset $\adversarial{\set G}$ indirectly through the parameters $\widehat \param$ \citep{biggio2018wild, munoz2017hypergradientpoisoning}.

In the graph domain, the perturbations can manifest as discrete addition and removal of the entities or triples, or continuous-valued perturbations of the entity features \citep{gunnemann2022graphadversarialrobustness, zugner2018nettack}. Since the KGE models are trained in a transductive learning setting, the missing links can only be predicted between entities seen during the training. Thus, deletion or addition of entities is not a realistic setting. Additionally, the knowledge graphs in practice do not have pre-existing node features and the embeddings in a KGE model architecture are initialized with random feature values for training \citep{hamilton2020graphrepresentationlearning}. Thus, perturbations to the entity feature space are also not applicable.

\begin{figure}[]
    \centering
    \begin{subfigure}[htb]{1\textwidth}
        \includegraphics[width=1\textwidth]{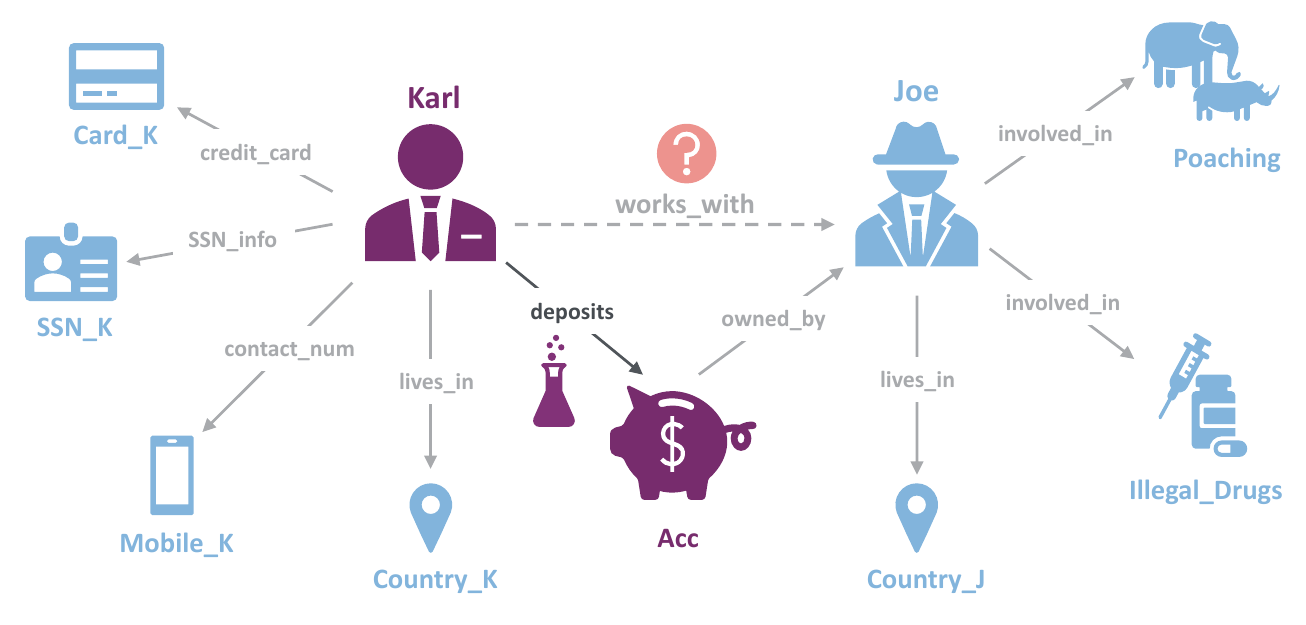}
        \caption{Adversarial Deletions}
    \end{subfigure}
    \newline
    \newline
    \newline
    \newline
    \begin{subfigure}[htb]{1\textwidth}
        \includegraphics[width=1\textwidth]{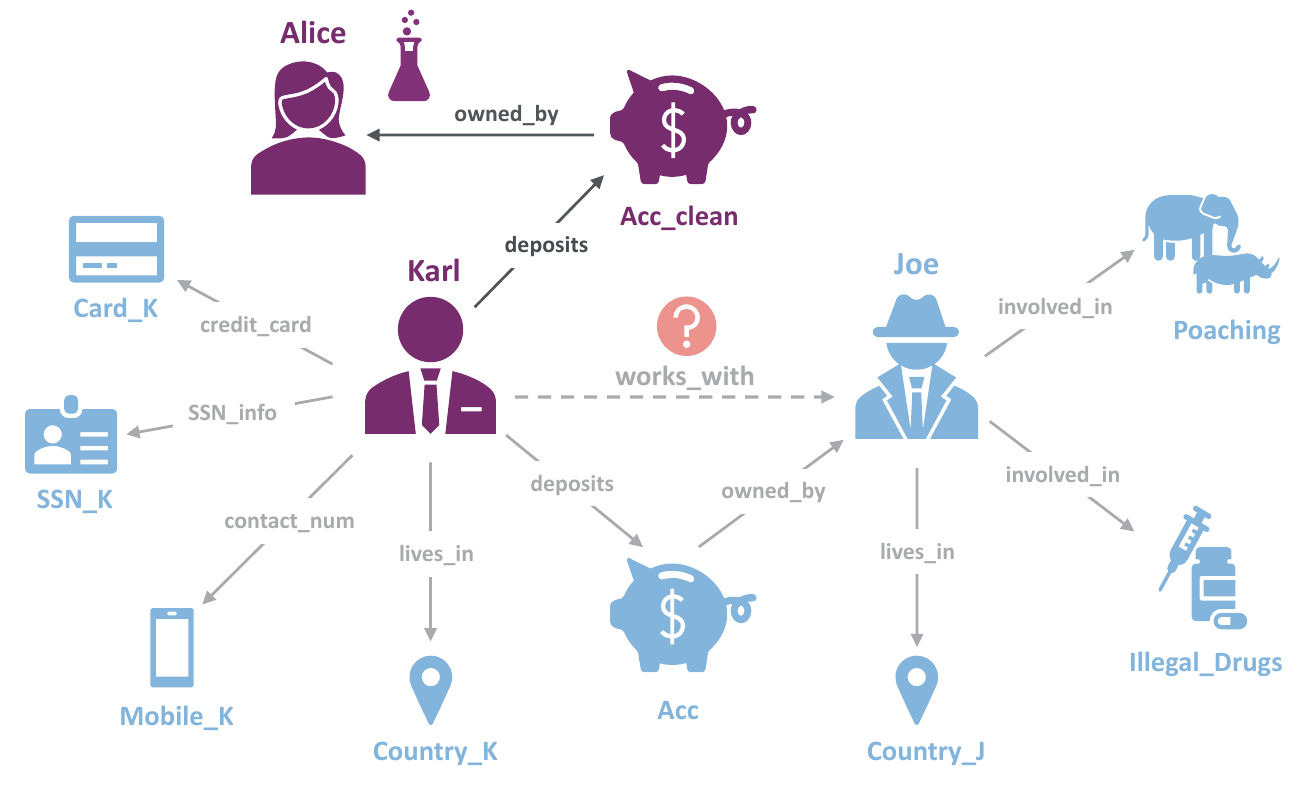}
        \caption{Adversarial Additions}
    \end{subfigure}
    \caption{An illustration of the perturbation space for adversarial attacks on the example financial knowledge graph in Figure \ref{fig:knowledge_graph}. To degrade the KGE model's predictive performance for target triple $(Karl, works\_with, Joe)$, the adversarial attacker can delete the existing triple $(Karl, deposits, Acc)$, or they can add the new triples $(Karl, deposits, Acc\_clean)$ and $(Acc\_clean, owned\_by, Alice)$. }
    \label{fig:problem_perturbations}
\end{figure}

The perturbation space for adversarial attacks in this thesis is the \emph{deletion} or \emph{addition} of triples to the training knowledge graph. 
This means that the perturbed graph $\adversarial{\set G}$ can be obtained from the original knowledge graph $\set G$ by removing or adding triples to $\set G$.
Additionally, the perturbations are restricted to the neighbourhood of the target triples. The neighbourhood of the target triple $z \coloneqq (z_s,z_\mathtt{r},z_o)$ is the set of triples that have the same subject or the same object as the target triple, i.e. $\mathcal{X}:= \{x \coloneqq (x_s, x_\mathtt{r}, x_o)\, |\, x_s \in \{z_s, z_o\} \vee x_o \in \{z_s, z_o\}\}$. This restriction is added to make the attack design tractable. However, it is still a realistic restriction as during model training, the embeddings of entities and relation in a triple are updated only due to its neighbourhood triples.
Figure \ref{fig:problem_perturbations} illustrates the deletion and addition perturbations for the example knowledge graph from Figure \ref{fig:knowledge_graph}. 

\subsection{Challenges}
Solving the bilevel optimization for data poisoning attacks is challenging and often intractable, even for the non-graph data modalities \citep{biggio2018wild, munoz2017hypergradientpoisoning, zugner2018gnnmetagradients}. The main challenge in this optimization is estimating the impact of a candidate poisoning point on the \emph{solution} of the learning algorithm. A naive approach to measure this impact would be to generate candidate poisoned graphs, train separate models on these candidates and evaluate the adversarial objective for each poisoned model. The candidate perturbation that is most effective for the adversarial objective can then be selected as the adversarial perturbation. However, this approach of re-training a new model for each candidate perturbation is computationally intractable. 

Some early studies on data poisoning (for non-graph data modalities) replaced the inner optimization problem with a closed-form solution \citep{biggio2012poisoningsvm, munoz2017hypergradientpoisoning}. This allowed the outer optimization to be solved directly through gradient descent. However, the solution is applicable to very specific families of ML models only, where the equilibrium solution of the inner optimization can be obtained. It is thus, not a practical solution for poisoning neural networks, where these stationarity conditions are difficult to derive. 

Another solution explored in \citet{munoz2017hypergradientpoisoning, zugner2018gnnmetagradients, wallace2021poisoningnlp, franceschi2019structurelearninggnn} relies on solving the bilevel optimization through hyperparameter optimization. In this approach, the perturbed dataset is treated as a hyperparameter to be optimized. The hyperparameter is learned by backpropagating the hyper-gradients, that is the gradients of attacker's loss $\atkloss(\set{Z}, \widehat \param)$ with respect to the hyperparameter $\adversarial{\set{G}}$ in the inner optimization. The hypergradients can be computed by replacing the inner optimization with a set of training iterations performed by the learning algorithm. This solution explicitly saves the parameter updates for the inner optimization, which allows the gradients of the loss in the outer objective to be backpropagated through these updates. However, completely unrolling the entire set of parameter updates is computationally expensive in terms of both time and memory. To resolve this issue, \citet{wallace2021poisoningnlp} and \citet{franceschi2019structurelearninggnn} unroll the training process for a fixed number of steps only. Similarly, \citet{zugner2018gnnmetagradients} estimates the inner optimization for a fixed number of steps based on a first order approximation of hyper-gradients from the meta-learning literature. On the other hand, \citet{munoz2017hypergradientpoisoning} does not store the updates for inner training, but computes the hypergradients directly during backpropagation by reversing the optimization steps.  

However, these solutions are not applicable for selecting the optimal perturbations for KGE models. This is because, unlike other neural architectures, the KGE models do not have a differentiable mapping from the graph structure to the embedding domain. For example, the optimization objective for Graph Neural Network (GNN) architectures \citep{hamilton2020graphrepresentationlearning} performs convolution operation on the adjacency matrix representation of the graph. This means that a loss function based on the model parameters can be differentiated w.r.t. the adjacency matrix. On the other hand, KGE models use a lookup table to map the graph structure to the embeddings (Chapter \ref{ch:background}). Computing the gradients for this lookup operation is not straightforward.

Additionally, the above solutions for data poisoning attacks do not scale to multi-relational knowledge graphs. For images and smaller, undirected graphs, the bilevel formulation optimizes the tensor representation of the image or the adjacency matrix for the graph. The corresponding perturbation space for knowledge graphs would be the adjacency tensor representation of the graph. However, back-propagating the gradients w.r.t. this entire tensor is computationally expensive and in practice, intractable. Thus, for poisoning KGE models in this thesis, the impact of candidate perturbation triples on the attacker's objective is measured for each candidate triple individually. %
This in turn introduces a combinatorial search space over the candidate triples to select the adversarial triple.

For adversarial deletions, the combinatorial search space is limited to the existing triples in the knowledge graph. However, it is computationally intractable to enumerate through all the candidate adversarial additions. 
For example, for every target triple, the possible number of adversarial additions in the neighbourhood of each entity are $\mathcal{E} \times \mathcal{R}$. 
For the benchmark knowledge graphs with thousands of nodes, this search space is of the order of millions for a single target triple. %

\subsection{Design Requirements}
\label{sec:problem_attack_design}
Based on the above discussion, identifying the optimal perturbed knowledge graph by solving the bilevel attack formulation is not tractable. However, the formal problem specification can be used as an intuitive guide for the design of data poisoning attacks against KGE models. Specifically, for the research in this thesis, the challenges highlighted above are addressed through the following design requirements -  
\begin{enumerate}
    \item \emph{Impact Metric:} The inner and outer optimization in the formal attack specification essentially aims to quantify the impact of a perturbation on the adversarial objective $\atk(\adversarial{\set{G}}, \param)$. Even without optimizing the bilevel problem, adversarial perturbations can be identified through a heuristic estimation of this impact. Thus, the design of data poisoning attacks requires an \emph{efficient} method to estimate the impact of a candidate perturbation on the adversarial attack objective $\atk(\adversarial{\set{G}}, \param)$.
    \item \emph{Combinatorial Search:} Given a measure of the perturbation impact, the perturbation with maximum impact needs to be selected. This combinatorial search over the candidate triples is tractable only for the adversarial deletions. Thus, the design of adversarial additions requires an \emph{efficient} method for the combinatorial search over the candidate perturbations.
\end{enumerate}

Along with the discussion in the previous section, these design requirements indicate that attack efficiency is a pre-requisite for the design of effective adversarial attacks against KGE models.
Thus, the aim of this research is to design efficient methods for data poisoning attacks that effectively degrade the ranks of the target triples predicted by the KGE model.
While the main consideration for attack design is attack effectiveness, designing \emph{effective} data poisoning attacks requires \emph{efficient} solutions.

\section{State-of-the-art Solutions}
\label{sec:sota_kgeattack}
Recently, some studies have attempted to investigate the adversarial vulnerabilities of representation learning for knowledge graphs. Of these, the studies most closely related to this thesis are \citet{zhang2019kgeattack} that proposes Direct Attacks, \citet{pezeshkpour2019criage} that proposes CRIAGE, and \citet{lawrence2021gradientrollback} that proposes Gradient Rollback.
Similar to this thesis, these studies focus on the task of predicting missing links in knowledge graphs using KGE models. Additionally, the attack threat model considered in these studies is similar to the threat model for this research (Section \ref{sec:problem_threatmodel}). The attacker is assumed to have white-box knowledge about the model parameters and aims to degrade the predictive performance by making edits to the training dataset. Thus, the adversarial attacks from these studies are used as baselines to evaluate the attacks proposed in the thesis.

\citet{zhang2019kgeattack} proposed methods for both adversarial deletions and adversarial additions, and used these to investigate the security vulnerabilities of KGE models.
\citet{pezeshkpour2019criage} also proposed both adversarial deletions and additions, but only used the adversarial additions for poisoning KGE models. The study used adversarial deletions as post-hoc explanations for KGE model predictions. The two tasks of adversarial deletions and post-hoc explanations are equivalent when the explanations for model predictions are counterfactual in nature and the adversarial objective is defined in terms of the model predictions.
This is because, both adversarial deletions for the adversarial objective \(
    \atk(\adversarial{\set{G}}, \widehat \param) \coloneqq \atkloss(\set{Z}, \widehat \param)
\)
and counterfactual explanations for model predictions, aim to identify the facts in the input knowledge graph whose removal would maximally impact the model predictions on the set of target triples $\set Z$. Thus, both adversarial deletions and additions proposed in \citet{pezeshkpour2019criage} are included in this research. 
Similar to CRIAGE, \citet{lawrence2021gradientrollback} proposed a method for post-hoc explanations of KGE models. This explanation method is also included for adversarial deletions in this thesis.

\subsection{Adversarial Deletions}
As discussed in the previous section, the main design requirement for adversarial deletions is a metric for the impact of a candidate triple's removal on the adversarial objective. 
The three methods for adversarial deletions in \citet{zhang2019kgeattack}, \citet{pezeshkpour2019criage} and \citet{lawrence2021gradientrollback} specify the adversarial objective in terms of the KGE model's predicted scores for the target triples. That is, for \(
    \atk(\adversarial{\set{G}}, \widehat \param) \coloneqq \atkloss(\set{Z}, \widehat \param)
\), the attack loss $\atkloss$ is a function of the KGE model predictions $f_z$ for target triples $\set Z$.
The methods differ in how the impact of perturbations on the KGE model predictions for target triples is estimated.

Direct Attack \citep{zhang2019kgeattack} uses the KGE model scores to measure the adversarial candidate's impact. 
Given a target triple, Direct Attack first identifies the perturbed embeddings that would degrade the KGE score predicted for the target triple. The attack then ranks the candidate deletions based on the difference in their KGE scores for original and perturbed embeddings. The intuition is to exploit the change in the candidate's score due to the embedding perturbation as an estimate of the candidate's impact on perturbing the embedding. Removing a candidate with larger difference in its score would have more impact on perturbing the embedding and in turn the target prediction. However, the experiments in Chapter \ref{ch:instance_attribution} show that Direct Attack is effective for additive KGE models only.

CRIAGE \citep{pezeshkpour2019criage} estimates the impact of candidate deletions by specializing the Influence Functions (IF) \citep{koh2017understanding} to KGE models. The influence expression from \citet{koh2017understanding} is derived for the specific loss function of Binary Cross Entropy (BCE) on the training triple. The KGE specialized IF expression is computationally more efficient than IF, but it limits the method's applicability to multiplicative KGE models that are trained using the BCE loss.

On the other hand, Gradient Rollback \citep{lawrence2021gradientrollback} estimates the influence of training triples on the learned parameters by keeping track of the gradient values during the training process. At the start of the training, the influence value for all training triples is initialized to zero. For each training step, the change in parameter values due to a triple is accumulated to the influence value of that triple. After training, the influence of a triple is obtained by subtracting its parameter updates from the model parameters.
In other words, the method makes separate copies of the parameter updates due to each training example, and estimates the impact of removing an example by subtracting its copy of parameter updates from the final model parameters. While this method provides theoretical bounds on the influence approximation, it requires additional storage for the separate parameter updates for all training triples. For knowledge graphs with millions of training triples, this overhead becomes expensive. Additionally, to track updates due to each triple, the method can only be applied for a batch size of 1.

In this thesis, the influence of candidate adversarial deletions is estimated through model-agnostic instance attribution methods. Unlike CRIAGE, these methods are applicable for all KGE models. Unlike Gradient Rollback, they do not require additional storage space. Data poisoning attacks using the instance attribution methods are discussed in Chapter \ref{ch:instance_attribution} of the thesis.

\subsection{Adversarial Additions}
As discussed in Section \ref{sec:problem_attack_design}, while the search space for adversarial deletions is limited to candidate triples in the training graph, it is computationally intractable to enumerate through all the possible candidate additions. Adversarial additions against KGE models have been proposed in Direct Attack \citep{zhang2019kgeattack} and CRIAGE \citep{pezeshkpour2019criage}. Both studies follow a similar two-step design approach for adversarial additions.

In the first step, the optimal gradient direction to perturb the latent space is computed. Direct Attack optimizes for the latent representation of the target entity in the target triple. CRIAGE optimizes for a function of the latent representations of entity-relation pair of the unknown adversarial addition. Both methods then determine discrete input perturbations from the perturbed latent space. Direct Attack randomly samples some candidate perturbations and scores them based on the difference of their KGE scores for original and perturbed embeddings. Thus, the attack is similar to the adversarial deletions with Direct Attack, and reduces the search space by random sampling. On the other hand, CRIAGE trains an auto-encoder to decode the entity-relation pair from the perturbed latent space. However, the auto-encoder uses a vanilla neural architecture which is not effective for generative modelling of the knowledge graphs and supports multiplicative KGE models only .

This thesis addresses the challenge of large candidate space for adversarial additions by breaking down the combinatorial search into three smaller steps over the entities and relations. This is achieved by a novel reformulation of the adversarial objective for poisoning KGE models through adversarial additions. The attacks proposed in Chapter \ref{ch:relation_inference} of the thesis exploit the inductive abilities of KGE models to select these adversarial additions.

\subsection*{Adversarial Attacks for Other Threat Models}
The adversarial attacks discussed above consider the same or similar threat model as the attack threat model in Section \ref{sec:problem_threatmodel}, and are thus directly comparable to the contributions of the thesis.
However, there is some research that investigates the security vulnerabilities of Machine Learning tasks for knowledge graphs, but focuses on a different attack setting or ML task than the one in this research. 

In parallel to the research in this thesis, \citet{banerjee2021stealthypoisoningKGE} studies risk aware adversarial attacks against KGE models for link prediction. The study aims to improve the stealth of an adversarial attack by reducing the exposure risk of the attack. It is not included in the evaluation of the proposed attacks, because the research in this thesis focuses on improving the attack effectiveness, instead of the attack stealth.

In another parallel work, \citet{zhang2021adversarialattackcrosslingualalignment} proposes adversarial attacks against the ML task of cross-lingual knowledge graph alignment. Due to differences in the neural architecture and the ML task, this study has not been included for the evaluation of the proposed attacks in the thesis.

Research in \citet{raman2021deceivingknowledgeaugmentedmodels} also generates perturbations to a knowledge graph. These graph perturbations are proposed as a method to understand the stability of KG-augmented NLP tasks like commonsense Question Answering (QA) and item recommendation. The aim of the study is to maximally perturb the graph structure while preserving the model's predictive performance on the graph. This perturbation goal is the opposite of the goal of adversarial attacks. Thus, the methods proposed in the study are not comparable to the contributions of this thesis.

Additional literature that does not focus on the adversarial robustness of KGE models, but improves our understanding of the model predictions through different methods is discussed in Section \ref{sec:related_understandingkge} of Chapter \ref{ch:related_work}.

\section{Proposed Attack Strategy}

As discussed in Section \ref{sec:problem_threatmodel}, the goal of the adversarial attacker is to degrade the rank of a highly plausible target triple by deleting or adding triples to the training knowledge graph.

The main requirement for designing data poisoning attacks by adversarial \emph{deletions} is to estimate the impact of a candidate adversarial deletion on the adversarial attack objective $\atk(\adversarial{\set{G}}, \widehat \param)$. Similar to state-of-the-art solutions, the adversarial objective for deletions in this research is defined in terms of an attack loss $\atkloss(\set{Z}, \widehat \param)$ on the target triples $\set Z$. A proxy for this loss is the score predicted for each target triple using the KGE model's scoring function $f_z$ where $z \in \set{Z}$. Thus, the aim of the adversarial attacker is to craft adversarial edits that would degrade the predicted score $f_z$ of the target triple $z$. 

To measure a candidate perturbation's impact on the predictions for target triples, this thesis proposes to use the model-agnostic instance attribution methods from Interpretable Machine Learning \citep{molnar2019interpretablemlbook}. These methods identify the training instances that are influential to model predictions, and thus quantify the impact of a candidate perturbation on the predicted score of the target triple. This solution is used to identify the adversarial deletions directly. But it requires a combinatorial search through the candidate perturbations, which does not scale to adversarial additions. Thus, a heuristic method is proposed to select adversarial additions by replacing one of the two entities of the influential triple with the most dissimilar entity in the embedding space. This solution is discussed further in Chapter \ref{ch:instance_attribution}.

It is noteworthy that the adversarial attacker's aim is to degrade the KGE model's predictive performance. The predictive performance for a set of target facts is measured using the ranks of the target facts, and the attacker aims to degrade the ranks predicted by the KGE model. As the rank of a target triple relies on the score predicted for the target triple, the solution for adversarial deletions defines the adversarial objective as degrading the predicted scores $f_z$. For this adversarial objective, the impact of candidate perturbations on the model predictions needs to be explicitly quantified. Since the impact for all candidates cannot be computed together (as discussed in Section \ref{sec:problem_bilevel}), a combinatorial search over the candidate space becomes necessary.

The large search space is the main design challenge for adversarial additions. The combinatorial search is made necessary by formulating the adversarial objective in terms of the model predictions $f_z$. Thus, the search space can be reduced by re-formulating the adversarial objective $\atk(\adversarial{\set{G}}, \widehat \param)$. The rank of a highly plausible target triple can be degraded by improving the rank of a less plausible \emph{decoy triple}, instead of degrading the predicted score of the target triple directly. Thus, for the adversarial additions, this thesis specifies the adversarial objective of degrading the ranks of the target triples as improving the ranks of the decoy triples.

For a target triple $z := (z_s, z_\mathtt{r}, z_o)$, the decoy triple for degrading the rank on the object side would be $(z_s, z_\mathtt{r}, z_o')$ and the decoy triple for degrading the rank on subject side would be $(z_s', z_\mathtt{r}, z_o)$. The aim of the attacker then, is to select decoy triples from the set of valid synthetic negatives and craft adversarial additions to improve their ranks. The attacker does not add the decoy triple itself as an adversarial edit, rather chooses the adversarial edits that would improve the rank of a missing decoy triple after training. As a measure of the candidate perturbations' impact on the ranks of the decoy triples, the attacker exploits the inductive abilities of KGE models. As discussed in Section \ref{sec:kge_inductive}, The inductive abilities of KGE models are expressed through the relation inference patterns like symmetry, inversion and composition. 

Chapter \ref{ch:relation_inference} proposes a collection of heuristic approaches to select the decoy triples and craft adversarial additions that use different inference patterns to improve the model's predictive performance on these decoy triples. To overcome the challenge of large candidate space, the search space is broken down into smaller steps - (i) determine the adversarial relations; (ii) determine the decoy entities that most likely violate an inference pattern; (iii) determine the remaining adversarial entities in the inference pattern that most likely improve the rank of decoy triples.

In summary, the contributions proposed in Chapter \ref{ch:instance_attribution} and \ref{ch:relation_inference} of the thesis differ in their specification of the adversarial attack objective for degrading the predictive performance of KGE models.

\section{Evaluation Protocol}
\label{sec:problem_evalprotocol}
The research contributions of the thesis are evaluated using state-of-the-art evaluation protocol for data poisoning attacks \citep{xu2020advgraphsurvey, biggio2018wild}. The aim of this evaluation is to measure the \emph{effectiveness} of the proposed attack strategies in degrading the ranks predicted by the KGE models for the target triples at inference time. Additionally, as discussed in Section \ref{sec:problem_attack_design}, the design of data poisoning attacks requires \emph{efficient} methods to estimate the impact of candidate perturbations and to reduce the combinatorial search space for adversarial additions. Thus, the runtime complexity of the proposed attacks is also analyzed. Furthermore, illustrative examples of the attacks are provided for qualitative analysis and to help understand the predictive behaviour of KGE models.

The predictive performance of KGE models is evaluated using state-of-the-art evaluation protocol for missing link prediction, which was discussed in Section \ref{sec:kge_eval_protocol}. The knowledge graph is split into training, validation and test facts; a KGE model is trained using the training triples; and evaluated on the test triples. The evaluation task involves an entity ranking procedure. Negative corruptions are generated for the test triple to be predicted, invalid negatives already existing in the training or validation set are filtered out \citep{bordes2013transe}, and the test triple is ranked against the negatives using the score predicted by the KGE model. The evaluation metrics reported over the entire test set are MRR and Hits@N.

The experimental setup for evaluating the adversarial attacks against KGE models is as follows. First, a victim KGE model is trained on the original training knowledge graph. This model is used to predict the ranks for the triples in the original test set of the knowledge graph. To assess the attack effectiveness in \emph{degrading} the predictive performance, triples for which the original KGE model has good predictive performance are needed. Thus, a subset of the test set triples that is ranked highly plausible by the original model is selected as target triples.
Both Hits@1 and Hits@10 are popular metrics for KGE evaluation, and thus triples with ranks $\leq$ 1 or ranks $\leq$ 10 can be considered as highly plausible. However, the adversarial attacks proposed in Chapter \ref{ch:instance_attribution} include computationally expensive methods. Thus, to avoid expensive attack computation for a large number of target triples, only the subset with ranks $\leq$ 1 is selected as targets for adversarial deletions. On the other hand, the contributions in Chapter \ref{ch:relation_inference} are designed to be efficient (because the main design requirement for adversarial additions is efficient combinatorial search). Thus, the proposed adversarial additions are evaluated using the subset of test set that is ranked $\leq$ 10.

Given the target triples, adversarial deletions or additions are generated for these target triples using one of the proposed or baseline attack strategies. The original training knowledge graph is then perturbed by removing or adding the adversarial triples to the graph. Finally, a new poisoned KGE model is trained on the perturbed knowledge graph. The hyperparameters for training the poisoned KGE model are the same as the original KGE model. This eliminates the effect of hyperparameter tuning on the predictive performance of the poisoned model. The difference between the predictive performance of the original and poisoned KGE models indicates the effectiveness of different adversarial attacks.

The software implementation for evaluating the research contributions from Chapter \ref{ch:instance_attribution} is publicly available on the GitHub at \url{https://github.com/PeruBhardwaj/AttributionAttack}. The implementation for the contributions from Chapter \ref{ch:relation_inference} is available on the GitHub at \url{https://github.com/PeruBhardwaj/InferenceAttack}.

\subsection{Evaluation Metrics}
The \emph{effectiveness} of adversarial attacks is measured as the difference between the ranks of the target triples predicted by the original KGE model and the poisoned KGE model. For the experiments in Chapter \ref{ch:instance_attribution} and Chapter \ref{ch:relation_inference}, this is reported as the MRR and the Hits@1 for the target triples due to the original KGE model, as well as the poisoned models obtained by different attack strategies. Lower values of MRR and Hits@1 then indicate the higher effectiveness of an adversarial attack. Additionally, the relative percentage difference between the MRR due to the original and poisoned KGE models is reported. This is measured as 

$\Delta_{MRR} = (MRR_{original} - MRR_{poisoned})/MRR_{original}*100$

\noindent
The more negative the value for $\Delta_{MRR}$, the higher is the attack effectiveness.

\bigskip
\noindent
As discussed in Section \ref{sec:problem_attack_design}, an efficient attack strategy is a pre-requisite for the design of effective adversarial attacks. Thus, while the primary focus of the evaluation is on attack effectiveness, \emph{efficiency} of the adversarial attacks is also analyzed. The efficiency of different attack strategies is measured as the time taken to generate the adversarial perturbations for all the target triples. Thus, given a set of target triples and the original KGE model, the absolute time taken to identify the complete set of adversarial perturbations for all of these target triples is recorded. This time (in seconds) is reported for the experiments in both Chapter \ref{ch:instance_attribution}  and Chapter \ref{ch:relation_inference}.

\subsection{Baselines}
The proposed adversarial attacks are evaluated against several baselines. These include the random edits to the training knowledge graph, as well as state-of-the-art data poisoning attacks discussed in Section \ref{sec:sota_kgeattack}. The baselines for attack evaluation are as follows.
\begin{enumerate}
    \item \emph{Random\_n} : For this baseline, random edits are made in the neighbourhood of the target triple. Based on the attack perturbations to be evaluated, the edits are deletions and additions in Chapter \ref{ch:instance_attribution} and Chapter \ref{ch:relation_inference} respectively. 
    \item \emph{Random\_g} : This baseline attack is obtained by making global random edits to the training knowledge graph. These random edits are not restricted to the neighbourhood of the target triple, and refer to deletions or additions based on the adversarial attacks to be evaluated in Chapters \ref{ch:instance_attribution} and \ref{ch:relation_inference}. %
    \item \emph{Direct Attack}: The data poisoning attacks from \citet{zhang2019kgeattack} are referred to as Direct-Del and Direct-Add for adversarial deletions and additions respectively. Since the study proposes both adversarial deletions and additions, these attacks are used as baselines in both Chapters \ref{ch:instance_attribution} and \ref{ch:relation_inference}.
    \item \emph{CRIAGE} : The data poisoning attack from \citet{pezeshkpour2019criage} is referred to as CRIAGE. Since the study proposes both adversarial deletions and additions, it is used as a baseline in both Chapters \ref{ch:instance_attribution} and \ref{ch:relation_inference}.
    \item \emph{Gradient Rollback (GR)}: The method for post-hoc explanations from \citet{lawrence2021gradientrollback} is referred to as Gradient Rollback (GR). It is used as a baseline for adversarial deletions in Chapter \ref{ch:instance_attribution}.
\end{enumerate}

The evaluation protocols used in \citet{zhang2019kgeattack}, \citet{pezeshkpour2019criage} and \citet{lawrence2021gradientrollback} differ with respect to their definition of the target triple's neighbourhood. Direct Attack \citep{zhang2019kgeattack} was evaluated for edits in the neighbourhood of the target subject entity only. By contrast, CRIAGE \citep{pezeshkpour2019criage} was proposed and evaluated for edits in the neighbourhood of the object entity only. While both these studies use the same evaluation metrics for attack effectiveness, GR \citep{lawrence2021gradientrollback} being a post-hoc explanation method uses different evaluation metrics. Further, all three studies used different KGE models and benchmark datasets for the attack evaluation. 

Thus, to ensure a fair evaluation of the research contributions for this thesis, all state-of-the-art methods are re-implemented with the same neighbourhood definition. As discussed in Section \ref{sec:problem_attack_design}, for this thesis, the target triple's neighbourhood is defined as the set of triples that have the same subject or the same object as the target triple. That is, for the target triple $z \coloneqq (z_s,z_\mathtt{r},z_o)$, the neighbourhood is computed as $\mathcal{X}:= \{x \coloneqq (x_s, x_\mathtt{r}, x_o)\, |\, x_s \in \{z_s, z_o\} \vee x_o \in \{z_s, z_o\}\}$. Additionally, the attacks are evaluated against the same KGE models for the same benchmark datasets. %
Publicly available software implementations for CRIAGE\footnote{\url{https://github.com/pouyapez/criage}} and Gradient Rollback\footnote{\url{https://github.com/carolinlawrence/gradient-rollback}} are used. Since an implementation of the Direct Attack is not publicly available, this method is re-implemented from scratch. Further details on the implementation of baselines are discussed in Chapters \ref{ch:instance_attribution} and \ref{ch:relation_inference}. 

\subsection{Models and Datasets}
The evaluation of the adversarial attacks is conducted against four KGE models -- DistMult \citep{yang2015distmult}, ComplEx \citep{trouillon2016complex}, ConvE \citep{dettmers2018conve} and TransE \citep{bordes2013transe}. 
These models are the most widely used models for representation learning on knowledge graphs. Recent studies by \citet{kadlec2017kgebaselines}, \citet{ruffinelli2020olddognewtricks} and \citet{ali2021bringinglightintodark} have shown that these models achieve state-of-the art predictive performance for missing link prediction. 
Additionally, as discussed in Chapter \ref{ch:background}, they have varied inductive abilities and are representative of the different families of KGE models - tensor factorization based models, translation based models and deep learning based models \citep{rossi2021kgesurvey, ali2021bringinglightintodark}. Thus, these four KGE models are suitable for the evaluation of the research contributions in this thesis.

The proposed adversarial attacks are evaluated on two publicly available\footnote{\url{https://github.com/TimDettmers/ConvE}} knowledge graphs -- WN18RR \citep{dettmers2018conve} and FB15k-237 \citep{toutanova2015observed}. These knowledge graphs are the benchmark datasets for the task of missing link prediction using Knowledge Graph Embeddings. As discussed in Section \ref{sec:kge_datasets}, they are both subsets of real-world knowledge graphs, namely WordNet and Freebase. Some of the earliest methods for KGE models were evaluated on these knowledge graphs. Despite the introduction of additional benchmark datasets, these remain state-of-the-art datasets for KGE model evaluation. Furthermore, they represent the knowledge from different domains -- while WN18RR is derived from a lexical ontology for English, FB15k-237 is a subset of a cross-domain knowledge graph created from crowdsourcing on the Web.
Hence, both of these datasets are representative knowledge graphs for the task of missing link prediction and are suitable for evaluating the contributions of this thesis.

\section{Summary}
As Knowledge Graph Embeddings are increasingly used in high-stakes domains, examining their security vulnerabilities has become critical. This chapter introduced the problem specification for designing adversarial attacks against KGE models.
A brief background on adversarial attacks from the filed of Adversarial Machine Learning was discussed, and a taxonomy for defining different attacks was introduced. The threat model for adversarial attacks proposed in this thesis was also introduced. This threat model provides a specification of the adversarial attacker's goals, their knowledge of the victim model and their capability for perturbing the input data. 

Based on the attack threat model, the mathematical formulation of data poisoning attacks was introduced. The attacks are expressed as a bilevel optimization problem, where the inner optimization represents the KGE learning process on the poisoned data and the outer optimization evaluates the impact of the poisoned dataset on the adversarial attack objective. Designing methods to solve this optimization problem is challenging. Thus, a set of heuristic design requirements for data poisoning attacks against KGE models was introduced. State-of-the-art poisoning attacks for addressing these requirements were discussed and the adversarial attack strategy proposed in this thesis was defined. Finally, the methodology for evaluating the research contributions of this thesis was discussed.

Building on the problem formulation in this chapter, the next chapter proposes and evaluates the methods for designing adversarial deletions against KGE models for missing link prediction in knowledge graphs.

\chapter{Poisoning via Instance Attribution Methods}
\label{ch:instance_attribution}
As discussed in Section \ref{sec:problem_attack_design}, the main design requirement for adversarial deletions against KGE models is a measure for the impact of adversarial candidates on the adversarial objective for the target facts. Let the \emph{target triples} be denoted as \(\mathcal{Z} \coloneqq \{z \coloneqq (z_s, z_\mathtt{r}, z_o)\}\). The aim of the attacker is to degrade the ranks predicted for these target triples. Since the rank of a target triple is assigned based on the scores predicted by the KGE model, the adversarial objective can be specified in terms of the KGE model predictions for the target triples. The attack strategy then is to measure the impact of candidate perturbations on the KGE model predictions for target facts and select the candidate perturbation with maximum impact. This chapter investigates the instance attribution metrics from the research field of Interpretable Machine Learning \citep{molnar2019interpretablemlbook} for designing the adversarial deletions against KGE models. The majority of the contents of this chapter are taken verbatim from the author's publication \citet{bhardwaj2021instanceattribution}.

\section{Threat Model}
\label{sec:ia_threatmodel}
The attack threat model used in this chapter is the same as the threat model defined in Section \ref{sec:problem_threatmodel}. This is also the same threat model as the state-of-the-art data poisoning attacks against KGE models \citep{pezeshkpour2019criage, zhang2019kgeattack, lawrence2021gradientrollback}. 
To ensure reliable vulnerability analysis, the attack setting being considered is a white-box attack setting. In this setting, the attacker has full knowledge of the victim model architecture and access to the learned embeddings. However, they cannot perturb the architecture or the embeddings directly. Rather, the learned model can be affected only through perturbations in the training data. The focus of this chapter are \emph{adversarial deletions}, but a heuristic method to generate \emph{adversarial additions} from these deletions is also proposed. For both types of perturbations, the attacker is restricted to making only one edit in the neighbourhood of the target triple. The neighbourhood of the target triple $z \coloneqq (z_s,z_\mathtt{r},z_o)$ is the set of triples that have the same subject or the same object as the target triple, i.e. $\mathcal{X}:= \{x \coloneqq (x_s, x_\mathtt{r}, x_o)\, |\, x_s \in \{z_s, z_o\} \vee x_o \in \{z_s, z_o\}\}$.

\section{Instance Attribution Methods}
For adversarial deletions, training triples that have influenced the model's prediction on the target triple need to be identified. Deleting these influential triples from the training set will likely degrade the prediction on the target triple. Thus, an \emph{influence score} \(\phi (z,x) : \mathcal{T} \times \mathcal{T} \rightarrow \mathbb{R} \) is defined for the pairs of triples \((z,x) \in \mathcal{T} \times \mathcal{T}\). This score indicates the influence of training triple $x$ on the prediction of target triple $z$. 
Larger values of the influence score \(\phi (z,x)\) indicate that removing $x$ from the training data would cause larger reduction in the predicted score on $z$. 
Thus, the training triple with the highest influence sore is selected as the adversarial deletion.
Figure \ref{fig:ia_examples} illustrates an example - to degrade the prediction for $(Karl,\mathtt {works\_with}, Joe)$, the influence scores for neighbouring training triples are computed; $(Karl,\mathtt {deposits}, Acc)$ is identified as the most influential triple, and selected as adversarial deletion.

Trivially, the influence score for a training triple $x$ can be computed by removing $x$ from the training set and re-training the KGE model on the knowledge graph without $x$. However, this is a prohibitively expensive step that requires re-training a new KGE model for every candidate influential triple.
The instance attribution methods from the literature on Interpretable Machine Learning \citep{molnar2019interpretablemlbook} provide a more efficient solution. These methods estimate the influence of training triples on the model predictions without re-training the model. They have been used to provide post-hoc explanations for neural networks for images \citep{koh2017understanding, hanawa2021evaluationsimilaritymetrics, charpiat2019inputsimilarity} and for text \citep{han2020influencenlp, han2020influenceveiledtoxicity, pezeshkpour2021instanceattribution}. Since these methods are post-hoc in nature, they are agnostic of the model architectures and the input data modalities. Yet, despite their widespread use for neural network explainability, they have not been used for KGE model architectures. Three types of instance attribution methods have been used in the previous literature. These are -- Instance Similarity metrics based on the feature representations of the target and the training instances \citep{hanawa2021evaluationsimilaritymetrics, charpiat2019inputsimilarity}, Gradient Similarity metrics based on the gradients of the model's loss function for the target and the training instances \citep{hanawa2021evaluationsimilaritymetrics, charpiat2019inputsimilarity} and Influence Function \citep{koh2017understanding} that estimates the effect of leave-one-out training. All of these methods for instance attribution are used to the select the most influential training triples as adversarial deletions against KGE models. Besides the Influence Function proposed in \citet{koh2017understanding}, prior literature has proposed Relative Influence Function \citep{barshan2020relatif} and Fisher Kernel \citep{khanna2019fisherkernels} methods for estimating the leave-one-out effect. However, these methods require very expensive computation of the full Hessian matrix or the Fisher Information matrix, and are therefore, not included in this research. The methods used for adversarial attacks against KGE model are discussed below.

\subsubsection{Instance Similarity}
The influence of training triple $x$ on the prediction of target triple $z$ is estimated as the similarity of their feature representations. The intuition behind these metrics is to identify the training triples that a KGE model has learnt to be similar to the target triple and thus (might) have influenced the model's prediction on the target triple.
Computing this similarity between triples requires feature vector representations for the triples. On the other hand, standard KGE models provide representations for the entities and relations only and the standard KGE scoring functions assign a scalar score to the triples (Section \ref{sec:kge_scoring_functions}). However, it is noteworthy that this scalar value is obtained by reducing over the embedding dimension. For example, in the tri-linear dot product for DistMult \citep{yang2015distmult}, the embeddings of subject, relation and object are multiplied element-wise and then the scalar score for the triple is obtained by summing over the embedding dimension, i.e. $f_t \coloneqq \langle \vec{e}_s, \vec{e}_\mathtt{r}, \vec{e}_o\rangle \coloneqq \sum_{i=1}^k \vec{e}_{s_i} \vec{e}_{\mathtt{r}_i} \vec{e}_{o_i}$ where $k$ is the embedding dimension. 

Thus, to obtain feature vector representations for the triples \(\vec{f}_t : \mathcal{E} \times \mathcal{R} \times \mathcal{E} \rightarrow \mathbb{R}^k \), the state-of-the-art KGE scoring functions are used without reduction over the embedding dimension. For the DistMult model, the triple feature vector is \(\vec{f} \coloneqq \vec{e}_s \circ \vec{e}_\mathtt{r} \circ \vec{e}_o\) where $\circ$ is the Hadamard (element-wise) product. Table \ref{tab:ia_scoring_functions_featurevectors} shows the feature vector scores for different KGE models used in this research.

Given the feature vectors for target triples $\vec{f}(z)$ and the feature vectors for training triples $\vec{f}(x)$, the following instance similarity metrics are defined based on the research in \citet{hanawa2021evaluationsimilaritymetrics}.

\paragraph{Dot Metric:} This metric computes the similarity between the target and training instances as the dot product of their feature vectors.
That is, \break
\(
 \phi_{dot} (z,x) \coloneqq \langle \vec{f}(z), \vec{f}(x) \rangle
\)

\paragraph{$\bm{\ell_2}$ Metric: } This metric computes similarity as the negative Euclidean distance between the feature vectors of target instance and test instance. 
That is, \break
\(
\phi_{\ell_2} (z,x) \coloneqq - \norm{\vec{f}(z) - \vec{f}(x)}_2
\)

\paragraph{Cosine Metric:} This metric computes similarity as the dot product between $\bm{\ell_2}$ normalized feature vectors of target and test instance, i.e. it ignores the magnitude of the vectors and only relies on the angle between them.
That is, \break
\(
\phi_{cos} (z,x) \coloneqq \cos{(\vec{f}(z), \vec{f}(x))}
\)

Here, the dot product for two vectors $\vec{a}$ and $\vec{b}$ is denoted as $\langle \vec{a}, \vec{b} \rangle \coloneqq \sum_{i=1}^p a_i b_i$; the $\bm{\ell_2}$ norm of a vector is denoted as $\norm{\vec{a}}_2 \coloneqq \sqrt{\langle \vec{a}, \vec{a} \rangle}$; and the cos similarity between vectors $\vec{a}$ and $\vec{b}$ is denoted as $\cos(\vec{a}, \vec{b}) \coloneqq \nicefrac{\langle \vec{a}, \vec{b} \rangle}{\norm{\vec{a}}_2 \norm{\vec{b}}_2}$.

\begin{table}
    \centering
    \small
    \begin{tabular}{l c c}
    \toprule
    
     \multicolumn{1}{c}{\textbf{Model}} & \textbf{Scoring Function} & \textbf{Feature Vectors}  \\
     \midrule
       DistMult  &  $ \tdot{\vec{e}_s}{\vec{e}_{\mathtt{r}}}{\vec{e}_o}$  & $ \vec{e}_s \circ \vec{e}_{\mathtt{r}} \circ \vec{e}_o$  \\ [2pt]
       ComplEx  &  $ \Re(\tdot{\vec{e}_s}{\vec{e}_{\mathtt{r}}}{\overline{\vec{e}_o}})$   &   $ \Re(\vec{e}_s \circ \vec{e}_{\mathtt{r}} \circ \overline{\vec{e}_o})$ \\ [2pt]
       ConvE    &  $ \langle (\vec{e}_s * \vec{e}_{\mathtt{r}}), \vec{e}_o \rangle$  &  $ (\vec{e}_s * \vec{e}_{\mathtt{r}}) \circ \vec{e}_o $ \\ [2pt]
       TransE   &  $- \norm{\vec{e}_s + \vec{e}_{\mathtt{r}} - \vec{e}_o}_p$  &  $- (\vec{e}_s + \vec{e}_{\mathtt{r}} - \vec{e}_o)$ \\ [2pt]
       \bottomrule
    \end{tabular}
    \caption{Scoring functions \(f_{s\mathtt{r}o}\) and the proposed Triple Feature Vectors \(\vec{f}_{s\mathtt{r}o}\) of the KGE models used in this research. For ComplEx, $\vec{e}_s, \vec{e}_\mathtt{r}, \vec{e}_o \in \mathbb{C}^k$; for the remaining models $\vec{e}_s, \vec{e}_\mathtt{r}, \vec{e}_o \in \mathbb{R}^k$. Here, $\langle \cdot \rangle$ denotes the tri-linear dot product; $\circ$ denotes the element-wise Hadamard product; $\overline{\ \cdot\ }$ denotes conjugate for complex vectors; $\norm{\cdot}_p$ denotes l-p norm; $*$ is the neural architecture in ConvE, i.e. \(\vec{e}_s * \vec{e}_\mathtt{r} \coloneqq \sigma(\mathrm{vec}(\sigma([ \overline{\vec{e}_\mathtt{r}}, \overline{\vec{e}_s}] \ast \boldsymbol{\Omega})) \mat{W})\) where $\sigma$ denotes sigmoid activation, $\ast$ denotes 2D convolution; $\overline{\ \cdot\ }$ denotes 2D reshaping of real vectors, $\boldsymbol{\Omega}$ and $\boldsymbol{\mat{W}}$ are the additional model parameters.}
    \label{tab:ia_scoring_functions_featurevectors}
\end{table}

\subsubsection{Gradient Similarity}
The gradient of the loss for triple $z$ w.r.t. model parameters $\widehat{\param}$ is denoted as \(\vec{g}(z, \widehat{\param}) := \nabla_{\param}{\loss (z,\widehat{\param)}} \). Gradient similarity metrics compute similarity between the gradients due to target triple $z$ and the gradients due to training triple $x$. The intuition is to assign higher influence to training triples that have similar effect on the model's parameters as the target triple; and are therefore likely to impact the prediction on target triple \citep{charpiat2019inputsimilarity}. 
Thus, using the same similarity functions as Instance Similarity, the following three metrics for gradient similarity are defined - Gradient Dot (GD), Gradient $\bm{\ell_2}$ (GL) and Gradient Cosine (GC).

\paragraph{GD(dot): } \(\phi_{\mathrm{GD}} (z,x) \coloneqq \langle \, \vec{g}(z, \widehat{\param})\, , \, \vec{g}(x, \widehat{\param}) \, \rangle \)

\paragraph{GL ($\bm{\ell_2}$):} \(\phi_{\mathrm{GL}} (z,x) \coloneqq - \norm{\vec{g}(z, \widehat{\param}) - \vec{g}(x, \widehat{\param})}_2 \)

\paragraph{GC(cos):} \(\phi_{\mathrm{GC}} (z,x) \coloneqq \cos{( \, \vec{g}(z, \widehat{\param}) \, , \, \vec{g}(x, \widehat{\param}) \, )} \)

\subsubsection{Influence Function}
Influence Function (IF) is a classic technique from robust statistics and was introduced to explain the predictions of black-box models in \citet{koh2017understanding}. To estimate the effect of a training point on a model's predictions, it first approximates the effect of removing the training point on the learned model parameters. To do this, it performs a first order Taylor expansion around the learned parameters $\widehat{\param}$ at the optimality conditions. 

Following the derivation in \citet{koh2017understanding}, the effect of removing the training triple $x$ on $\widehat{\param}$ is given by 
\(
\nicefrac{d\widehat{\param}}{d\epsilon_i} = \mat{H}_{\widehat{\param}}^{-1} \ \vec{g}(x, \widehat{\param})
\). 
Here, $\mat{H}_{\widehat{\param}}$ denotes the Hessian of the loss function \(
\mat{H}_{\widehat{\param}} \coloneqq \nicefrac{1}{n} \sum_{t \in \mathcal{T}} \nabla_{\param}^{2}{\loss (t, \widehat{\param})}
\). 
Using the chain rule then, the influence of removing $x$ on the model's prediction at $z$ is approximated as $\langle \vec{g} (z,\widehat{\param}) \ ,  \nicefrac{d\widehat{\param}}{d\epsilon_i} \rangle$. Thus, the influence score using IF is defined as - 

\paragraph{IF: } \, 
\(\phi_{\mathrm{IF}} (z,x) := \langle \, \vec{g}(z, \widehat{\param}) \, , \, \mat{H}_{\widehat{\param}}^{-1} \vec{g}(x, \widehat{\param}) \, \rangle 
\)

\bigskip
\noindent
Computing the IF for deep learning based models poses two challenges - (i) storing and inverting the Hessian matrix is computationally too expensive for a large number of parameters; (ii) the Hessian is not guaranteed to be positive definite and thus, invertible because deep learning models are non-convex models. Since KGE models are learned from deep neural architectures, both of these challenges hold for KGE models as well. The state-of-the-art solution to address these challenges has been proposed in the original work by \citet{koh2017understanding}. 

To address the first challenge, instead of computing the exact Hessian matrix, a Hessian-vector product (HVP) of the model parameters with the target triple's gradient is estimated. That is, for every target triple $z$, the value \(
\mat{H}_{\widehat{\param}}^{-1} \ \vec{g} (z, \widehat{\param})
\) is pre-computed. Then, for each neighbourhood triple $x$ in the training set, the influence score $\phi_{\mathrm{IF}} (z,x)$ is computed as the dot product between the pre-computed HVP and the gradient due to $x$, that is, \( \vec{g} (x, \widehat{\param})\) . 
Furthermore, as suggested in \citet{koh2017understanding}, the stochastic estimator LiSSA \citep{agarwal2017lissa} is used to estimate the HVP in linear time by sampling the triples from the training data.
For the second issue of non-convexity, a "damping" term is added to the Hessian matrix so that it is positive definite and invertible. This term is a hyperparameter that is tuned to ensure that all eigenvalues of the Hessian matrix are positive, i.e. the Hessian matrix is positive definite. Further discussion on the validity of Influence Functions for non-convex settings is available in \citet{koh2017understanding}.

\begin{figure}[]
    \centering
    \begin{subfigure}[htb]{1\textwidth}
        \centering
        \includegraphics[scale=1,width=1\textwidth]{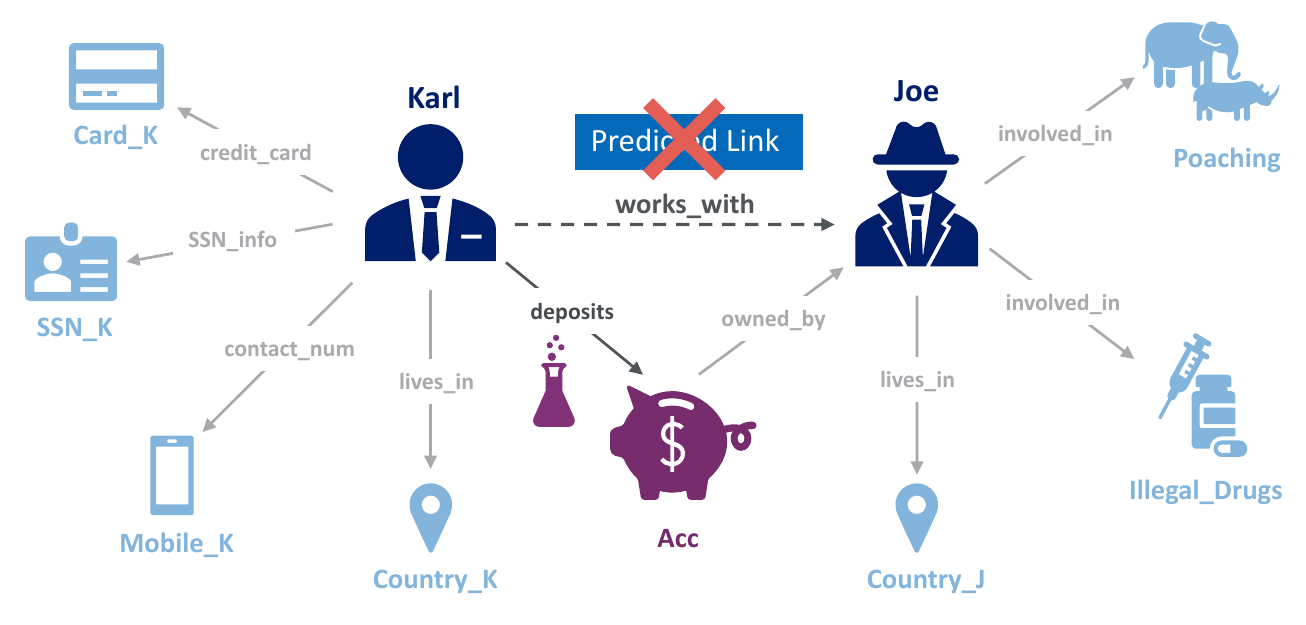}
    \caption{Adversarial Deletion}
    \end{subfigure}
    \newline
    \newline
    \newline
    \newline
    \begin{subfigure}[htb]{1\textwidth}
        \includegraphics[scale=1, width=1\textwidth]{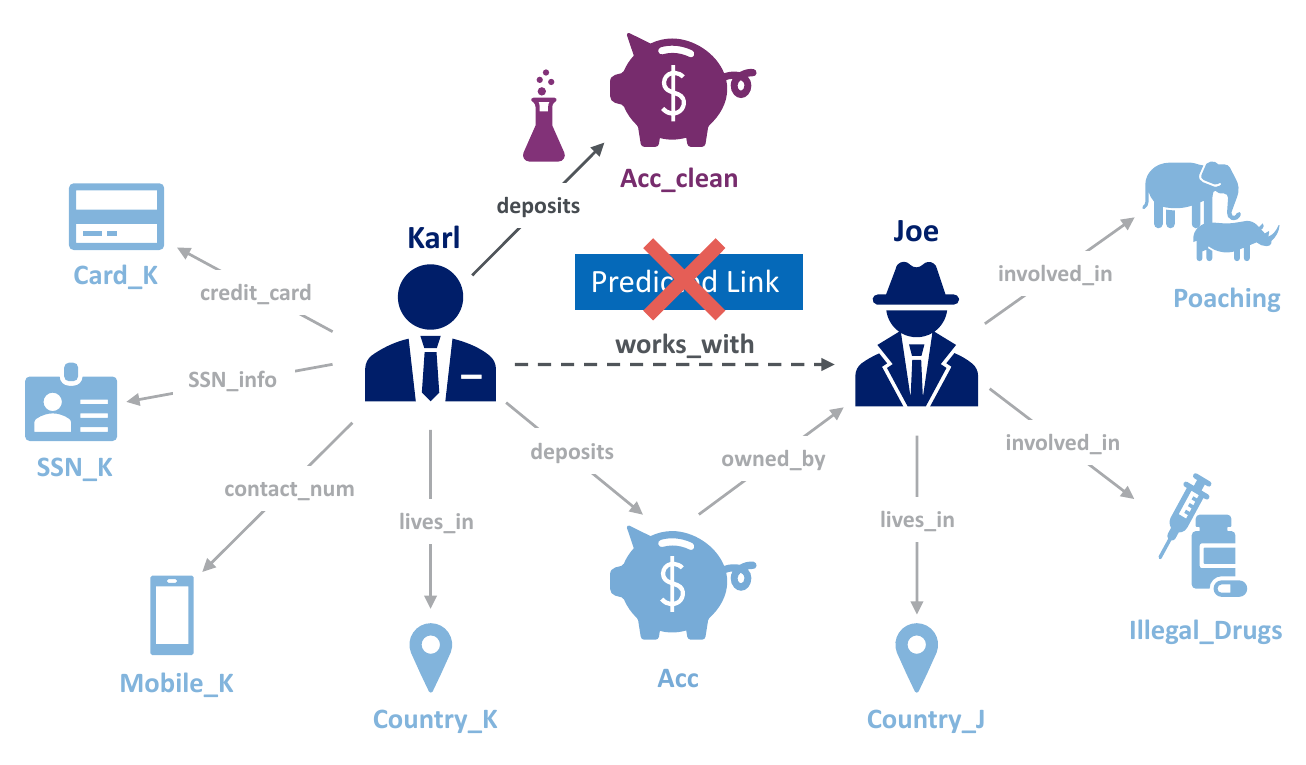}
    \caption{Adversarial Addition}
    \end{subfigure} 
    
    \caption{Illustrative examples for adversarial attacks based on instance attribution methods. To degrade the prediction for target triple $(Karl,\mathtt {works\_with}, Joe)$, the attacker can (a) delete the most influential triple $(Karl,\mathtt {deposits}, Acc)$, or (b) add the triple $(Karl,\mathtt {deposits}, Acc\_clean)$ that reduces the influence of the influential triple by selecting the entity $Acc\_clean$ that is dissimilar to $Acc$.}
    \label{fig:ia_examples}
\end{figure}

\section{Adversarial Additions}
In this attack setting, the adversarial attacker can only \emph{add} triples to the neighbourhood of target triple.
The Instance Attribution metrics from the previous section are used to select the training triple $x := (x_s, x_\mathtt{r}, x_o)$ in the neighbourhood of the target triple $z := (z_s, z_\mathtt{r}, z_o)$ that is most influential to the prediction of $z$. For brevity, lets assume $x_s = z_s$, i.e. the influential and target triples have the same subject. To generate adversarial addition using the influential triple, $x_o$ is replaced with another entity $\adversarial{x_o}$, which is most dissimilar to $x_o$ in the latent feature space. Since the adversarial triple $\adversarial{x} := (x_s, x_\mathtt{r}, \adversarial{x_o})$ has the same subject and relation as the influential triple but a different object, it should reduce the influence of the influential triple on the target triple's prediction. This in turn should degrade the model's prediction on the target triple. Consider the knowledge graph in Figure \ref{fig:ia_examples} as an example. To degrade the prediction for $(Karl,\mathtt {works\_with}, Joe)$, the most influential triple identified by an instance attribution method is $(Karl,\mathtt {deposits}, Acc)$. The influence of this influential triple is reduced by adding $(Karl,\mathtt {deposits}, Acc\_clean)$, where $Acc$ has been replaced with $Acc\_clean$.

For multiplicative models, the dissimilar entity $\adversarial{x_o}$ is selected using the cosine similarity between the embedding of $x_o$ and the embeddings of all entities $\mathcal{E}$. For additive models, the $\bm{\ell_2}$ similarity between the embedding of $x_o$ and the embeddings of all entities $\mathcal{E}$ is used.


\section{Experimental Setup}
\label{sec:ia_evaluation}
As discussed in Section \ref{sec:problem_evalprotocol}, the aim of the evaluation is to measure the effectiveness of the proposed attack strategies in degrading the KGE model's predictions on target triples at test time.
For this, the state-of-the-art evaluation protocol for data poisoning attacks \citep{xu2020advgraphsurvey} has been used. First, a victim KGE model is trained on the original knowledge graph. Next, adversarial deletions or additions are generated using one of the proposed attack strategies and the original knowledge graph is perturbed. Finally, a new KGE model is trained on the perturbed knowledge graph.
The same hyperparameters are used for training the original and poisoned KGE models.
The source code implementation of all experiments conducted in this chapter is available at \url{https://github.com/PeruBhardwaj/AttributionAttack}.

\subsection{Datasets}
The proposed adversarial attacks are evaluated on four state-of-the-art KGE models - DistMult, ComplEx, ConvE and TransE; on two publicly available benchmark datasets for link prediction\footnote{https://github.com/TimDettmers/ConvE}- WN18RR and FB15k-237. 
To evaluate the predictive performance of original and poisoned KGE models, the standard KGE evaluation protocol is used (Section \ref{sec:kge_eval_protocol}). In addition, following \citet{bordes2013transe} and as is the standard practice, triples from the validation and test set that contain unseen entities are filtered out.

To assess the attack effectiveness in \emph{degrading} performance on triples predicted as True, a set of triples that are predicted as True by the victim model need to be selected.
Thus, for this evaluation, a subset of the benchmark test set is selected that has been ranked the best (i.e. ranks=1) by the victim KGE model. If this subset has more than 100 triples, 100 triples are randomly sampled as the \emph{target triples}; otherwise all the triples are used as target triples. This pre-processing step is done to avoid the expensive Hessian inverse computation in the Influence Functions (IF) for a large number of target triples - for each target triple, estimating the Hessian inverse (as an HVP) using the LissA algorithm requires one training epoch.

Table \ref{tab:ia_data} shows the dataset statistics and the number of triples which are ranked best by the different KGE models. 

\begin{table}[h]
\centering
\footnotesize
\setlength{\tabcolsep}{5pt}
\begin{tabular}{c  l ll}
    \toprule           
    \multicolumn{2}{l}{} & \textbf{WN18RR} &  \textbf{FB15k-237} \\ 
    \midrule
    \multicolumn{2}{l}{Entities}                  &  40,559   & 14,505 \\ 
    \multicolumn{2}{l}{Relations}                 &  11       & 237 \\ 
    \multicolumn{2}{l}{Training}                  & 86,835    &  272,115            \\ 
    \multicolumn{2}{l}{Validation}                &  2,824    &  17,526   \\ 
    \multicolumn{2}{l}{Test}                      &   2,924   &  20,438    \\ 
    \midrule
    \multirow{4}{*}{\shortstack[l]{Subset \\ with \\ Best Ranks} }
    & DistMult  &   1,109   &  1,183    \\
    & ComplEx   & 1,198  &  1,238    \\
    & ConvE     &    1,106   &  901    \\
    & TransE    &    15   &   1223   \\
    \bottomrule
\end{tabular}
\caption{Statistics for WN18RR and FB15k-237.
Triples from the validation and test set that contained unseen entities were removed to ensure that new entities are not added as adversarial edits. 
The numbers above (including the number of entities) reflect this filtering.}
\label{tab:ia_data}
\end{table}

\subsection{Baselines}
The proposed attacks are evaluated against baseline methods based on random edits and the state-of-art data poisoning attacks against KGE models. 
\emph{Random\_n} is the baseline attack based on random edits that adds or removes a random triple from the neighbourhood of the target triple. \emph{Random\_g} adds or removes a random triple globally and is not restricted to the target's neighbourhood. 
\emph{Direct-Del} and \emph{Direct-Add} are the adversarial deletion and addition attacks proposed in \citet{zhang2019kgeattack}. \emph{CRIAGE} is the data poisoning attack from \citet{pezeshkpour2019criage} and is a baseline for both deletions and additions. 
\emph{GR} (Gradient Rollback) \citep{lawrence2021gradientrollback} uses influence estimation to provide post-hoc explanations for KGE models and can also be used to generate adversarial deletions. Thus, this method is included as a baseline for adversarial deletions only. 

The attack evaluations in \citet{zhang2019kgeattack, pezeshkpour2019criage, lawrence2021gradientrollback} differ with respect to the definition of their \emph{neighbourhood}. Thus, to ensure fair evaluation for this thesis, all the state-of-art methods are implemented with the same neighbourhood - triples that are linked to the subject or object of the target triple (Section \ref{sec:ia_threatmodel}). For this, the publicly available implementations for CRIAGE\footnote{https://github.com/pouyapez/criage} and Gradient Rollback\footnote{https://github.com/carolinlawrence/gradient-rollback} were used and the Direct-Del and Direct-Add were implemented by the author herself. Additional details on the implementation of KGE models and baselines; and the computing resources used are available in the Appendix \ref{apx:instance_attribution_training_details}.

\begin{table}
\centering
\fontsize{10}{10}\selectfont
\setlength{\tabcolsep}{3.75pt}
\begin{tabular}{  l  ll  ll   ll  lr }

\toprule
    
      & \multicolumn{2}{c}{\textbf{DistMult}} & \multicolumn{2}{c}{\textbf{ComplEx}} & \multicolumn{2}{c}{\textbf{ConvE}} & \multicolumn{2}{c}{\textbf{TransE}} \\
   \cmidrule(lr){2-3}  \cmidrule(lr){4-5}  \cmidrule(lr){6-7} \cmidrule(lr){8-9} 
     & \textbf{MRR}   & \textbf{H@1}  & \textbf{MRR}   & \textbf{H@1} & \textbf{MRR}   & \textbf{H@1} & \textbf{MRR}   & \textbf{H@1} \\
\midrule
    
     \textbf{Original}      & 1.00            &  1.00     &  1.00             & 1.00      &  1.00          &   1.00    & 1.00           &  1.00 \\
\midrule
     \multicolumn{9}{l}{\textbf{Baseline Attacks}} \\
\midrule
    
     Random\_n              & 0.87 (-13\%) &  0.82     & 0.85 (-15\%)   & 0.80        & 0.82 (-18\%)         & 0.79        &    0.82 (-18\%) &  0.70   \\
     Random\_g              & 0.97         &  0.95     & 0.96           & 0.93        &  0.99                & 0.98       &   0.93           &  0.87   \\
\midrule
     Direct-Del             & 0.88         &  0.77     &  0.86 (-14\%)  &  0.77        &  0.71 (-29\%)        & 0.64        &  0.54 \textbf{(-46\%)}  & 0.37    \\
     CRIAGE                 & 0.73 (-27\%) &  0.66     &   -             &  -          &  Er                  & Er        &   -          &  -   \\
    GR                      & 0.92        &  0.87     &  0.86           &  0.81       & 0.94                 & 0.91      &   0.84           &  0.73  \\
\midrule
    \multicolumn{9}{l}{\textbf{Proposed Attacks}} \\
\midrule
    
     Dot Metric               & 0.89                  &  0.82     & 0.85           & 0.79         & 0.84 (-16\%)      & 0.80       &  0.77     & 0.60   \\
     $\bm{\ell_2}$ Metric     & \textbf{0.25 (-75\%)} &  0.16      & 0.29 (-71\%)  & 0.20         &  0.88      &  0.78       &   0.62    & 0.50   \\
     Cos Metric               & \textbf{0.25 (-75\%)} & 0.16      & 0.29 (-71\%)   &  0.20        & 0.87        & 0.76      &  0.56 (-44\%)    & 0.40   \\
\midrule
     GD (dot)                  &   0.28 (-72\%) &  0.19        &  0.29                    & 0.21        &  0.25          & 0.21        &  0.71 (-29\%)   & 0.57    \\
     GL ($\bm{\ell_2}$)        &  0.30          &  0.20        &  0.28 \textbf{(-72\%)}   &  0.19       &  0.17 \textbf{(-83\%)}      & 0.12        &   0.72     & 0.60    \\
     GC (cos)                  &  0.29          &  0.19        &  0.29                    & 0.21        & 0.20   & 0.16        & 0.71 (-29\%)    & 0.57   \\
\midrule
    
     IF                       &  0.28 (-72\%)   & 0.19        & 0.29 (-71\%)    & 0.20        & 0.22 (-78\%)          & 0.17     &  0.71 (-29\%)     &  0.57  \\
     
\bottomrule    

\end{tabular}
\caption{\small Reduction in MRR and Hits@1 due to \textbf{adversarial deletions on target triples in WN18RR}. Lower values indicate better results; best results for each model are in bold. First block of rows are the baseline attacks with random edits; second block is state-of-the-art attacks; remaining are the proposed attacks. For each block, the \emph{best} reduction in percentage relative to the original MRR; computed as $(poisoned - original)/original*100$ is reported. }
\label{tab:ia_deletion_WN18RR}
\end{table}

\section{Evaluation of Attack Effectiveness}

For WN18RR and FB15k-237 respectively, Tables \ref{tab:ia_deletion_WN18RR} and \ref{tab:ia_deletion_FB15k-237} show the degradation in MRR and Hits@1 due to adversarial deletions; and Tables \ref{tab:ia_addition_WN18RR} and \ref{tab:ia_addition_FB15k-237} due to adversarial additions for state-of-the-art KGE models. 
Different patterns observed in these results are discussed below.

\emph{Note - }For the results reported in the publication associated with this chapter \citep{bhardwaj2021instanceattribution}, the implementation of Gradient Rollback (GR) contained a software error. This has been corrected in commit $62a39e7$ of the software, and updated results have been included in Tables \ref{tab:ia_deletion_WN18RR} and \ref{tab:ia_deletion_FB15k-237} of this chapter.

\begin{table}
\centering
\fontsize{10}{10}\selectfont
\setlength{\tabcolsep}{3.75pt}
\begin{tabular}{  l  ll  ll   ll  lr }

\toprule
    
      & \multicolumn{2}{c}{\textbf{DistMult}} & \multicolumn{2}{c}{\textbf{ComplEx}} & \multicolumn{2}{c}{\textbf{ConvE}} & \multicolumn{2}{c}{\textbf{TransE}} \\
   \cmidrule(lr){2-3}  \cmidrule(lr){4-5}  \cmidrule(lr){6-7} \cmidrule(lr){8-9} 
     & \textbf{MRR}   & \textbf{H@1}  & \textbf{MRR}   & \textbf{H@1} & \textbf{MRR}   & \textbf{H@1} & \textbf{MRR}   & \textbf{H@1} \\
\midrule
    
     \textbf{Original}      & 1.00            &  1.00     &  1.00             & 1.00      &  1.00          &   1.00    & 1.00           &  1.00 \\
\midrule
     \multicolumn{9}{l}{\textbf{Baseline Attacks}} \\
\midrule
     Random\_n              &  0.66 (-34\%)  &  0.52      &   0.65 (-35\%)   & 0.51           & 0.62 (-38\%)    &  0.46       &   0.71 (-29\%)   &  0.56      \\
     Random\_g              &  0.68          &  0.53      &  0.65 (-35\%)    &  0.51         &  0.63            &  0.50       &   0.75           &   0.61     \\
\midrule
     Direct-Del             & 0.59 (-41\%)     &  0.42     &   0.62 (-38\%)    &  0.47     &   0.57 (-43\%)          &  0.41       &   0.62 \textbf{(-38\%)}          &   0.45      \\
     CRIAGE                 & 0.62             & 0.47      &   -               &  -        &      Er        &    Er     &       -       &     -    \\
    GR           & 0.67             & 0.53      &  0.63             &  0.48     &  0.59            & 0.42        &   0.72           &   0.58      \\
\midrule
    \multicolumn{9}{l}{\textbf{Proposed Attacks}} \\
\midrule
     Dot Metric               & 0.63                  & 0.47        & 0.64                  & 0.49        & 0.60        & 0.44       &   0.74        &  0.62       \\
     $\bm{\ell_2}$ Metric     & 0.58                  & 0.41        & 0.56 \textbf{(-44\%)} &  0.40       & 0.53 \textbf{(-47\%)}       & 0.35        &   0.63 (-37\%)      &  0.46        \\
     Cos Metric               & 0.56 \textbf{(-44\%)}  & 0.39       & 0.57                  & 0.40        & 0.55        & 0.38        &  0.63 (-37\%)       &  0.45        \\
\midrule
     GD (dot)           &   0.60          & 0.44         &  0.60         & 0.45        & 0.55 (-45\%)     & 0.37        &  0.65           &   0.49              \\
     GL ($\bm{\ell_2}$)   &   0.62          &  0.45        &  0.60         &  0.45       & 0.56      & 0.41        &   0.70          &   0.58              \\
     GC (cos)           &   0.58 (-42\%)  &  0.42        &  0.57 (-43\%) & 0.39        & 0.57      & 0.40        &   0.64 (-36\%)         &  0.48               \\
\midrule
    
     IF       &  0.60 (-40\%)        &  0.44      &  0.60 (-40\%)   & 0.45         & 0.58 (-42\%)          & 0.43     &  0.66 (-34\%)     &    0.52               \\
\bottomrule    

\end{tabular}
\caption{\small Reduction in MRR and Hits@1 due to \textbf{adversarial deletions on target triples in FB15k-237}. Lower values indicate better results; best results for each model are in bold. First block of rows are the baseline attacks with random edits; second block is state-of-the-art attacks; remaining are the proposed attacks. For each block, the \emph{best} reduction in percentage relative to the original MRR; computed as $(poisoned - original)/original*100$ is reported.}
\label{tab:ia_deletion_FB15k-237}
\end{table}

\begin{table}
\centering
\fontsize{10}{10}\selectfont
\setlength{\tabcolsep}{3.6pt}
\begin{tabular}{  l  ll  ll   ll  lr }

\toprule
    
      & \multicolumn{2}{c}{\textbf{DistMult}} & \multicolumn{2}{c}{\textbf{ComplEx}} & \multicolumn{2}{c}{\textbf{ConvE}} & \multicolumn{2}{c}{\textbf{TransE}} \\
   \cmidrule(lr){2-3}  \cmidrule(lr){4-5}  \cmidrule(lr){6-7} \cmidrule(lr){8-9} 
     & \textbf{MRR}   & \textbf{H@1}  & \textbf{MRR}   & \textbf{H@1} & \textbf{MRR}   & \textbf{H@1} & \textbf{MRR}   & \textbf{H@1} \\
\midrule
    
     \textbf{Original}      & 1.00            &  1.00     &  1.00             & 1.00      &  1.00          &   1.00    & 1.00           &  1.00 \\
\midrule
     \multicolumn{9}{l}{\textbf{Baseline Attacks}} \\
\midrule
     Random\_n              & 0.99 (-1\%)       & 0.98       & 0.97 (-3\%)    & 0.94           &  0.99 (-1\%)           & 0.98       &   0.76 \textbf{(-24\%)}          &  0.57      \\
     Random\_g              & 0.99 (-1\%)     & 0.97        &  0.97 (-3\%)    & 0.95          &   0.99 (-1\%)          &  0.98       &   0.93           &  0.87      \\
\midrule
     Direct-Add             & 0.98 (-2\%)     &  0.96      &  0.95 (-5\%)    &  0.92        &  0.99 (-1\%)           &  0.98       &   0.81 (-19\%)          &   0.67      \\
     CRIAGE                 & 0.98 (-2\%)     & 0.97       &  -               &  -         &    Er          &  Er                  &     -         &    -     \\
\midrule
    \multicolumn{9}{l}{\textbf{Proposed Attacks}} \\
\midrule
     Dot Metric               &  0.97                      & 0.93        & 0.95         & 0.90        &   0.95 (-5\%)    &  0.91      &  0.95         & 0.90        \\
     $\bm{\ell_2}$ Metric     &  0.89 \textbf{(-11\%)}     & 0.78        &  0.88        &  0.77       &   0.98           &   0.96      &  0.87 (-13\%)       & 0.83         \\
     Cos Metric               & 0.89 \textbf{(-11\%)}      & 0.78        &  0.87 (-13\%) & 0.77        &   0.99          &   0.98      &  0.87 (-13\%)       &  0.83        \\
\midrule
     GD (dot)            &  0.90                     & 0.79         & 0.89                    & 0.79        &  0.92     & 0.85                 &   0.80 (-20\%)         &  0.73               \\
     GL ($\bm{\ell_2}$)  &  0.89 \textbf{(-11\%)}    & 0.79         & 0.86 \textbf{(-14\%)}   &  0.73       &  0.88 \textbf{(-12\%)} & 0.77    &   0.89          &  0.83               \\
     GC (cos)           &  0.90                     & 0.80         &  0.87                   &  0.76       &  0.91     &  0.82                &   0.80 (-20\%)         &  0.73               \\
\midrule
     IF           &  0.90 (-10\%)          & 0.79       &  0.89 (-11\%)              &  0.79       &  0.91 (-8.9\%)         & 0.82     &   0.77 (-23\%)    &  0.67                 \\
\bottomrule    

\end{tabular}
\caption{\small Reduction in MRR and Hits@1 due to \textbf{adversarial additions on target triples in WN18RR}. Lower values indicate better results; best results for each model are in bold. First block of rows are the baseline attacks with random edits; second block is state-of-the-art attacks; remaining are the proposed attacks. For each block, the \emph{best} reduction in percentage relative to the original MRR; computed as $(poisoned - original)/original*100$ is reported.}
\label{tab:ia_addition_WN18RR}
\end{table}

\begin{table}
\centering
\fontsize{10}{10}\selectfont
\setlength{\tabcolsep}{3.75pt}
\begin{tabular}{  l  ll  ll   ll  lr }

\toprule

      & \multicolumn{2}{c}{\textbf{DistMult}} & \multicolumn{2}{c}{\textbf{ComplEx}} & \multicolumn{2}{c}{\textbf{ConvE}} & \multicolumn{2}{c}{\textbf{TransE}} \\
   \cmidrule(lr){2-3}  \cmidrule(lr){4-5}  \cmidrule(lr){6-7} \cmidrule(lr){8-9} 
     & \textbf{MRR}   & \textbf{H@1}  & \textbf{MRR}   & \textbf{H@1} & \textbf{MRR}   & \textbf{H@1} & \textbf{MRR}   & \textbf{H@1} \\
\midrule
    
     \textbf{Original}      & 1.00            &  1.00     &  1.00             & 1.00      &  1.00          &   1.00    & 1.00           &  1.00 \\
\midrule
     \multicolumn{9}{l}{\textbf{Baseline Attacks}} \\
\midrule
     Random\_n              &  0.65 (-34\%)  & 0.50      &  0.69              &  0.57          &  0.61 (-39\%)   & 0.46        &  0.74            &  0.62      \\
     Random\_g              &  0.66         &  0.52       &  0.66 (-34\%)     &  0.52         &  0.63            &  0.50       &   0.73 (-27\%)          &  0.61      \\
\midrule
     Direct-Add             &  0.64 (-36\%)    &   0.48     &  0.66 (-34\%)              &  0.52     &  0.60 (-40\%)           &  0.45       &  0.72 (-28\%)           &   0.59      \\
     CRIAGE                 &  0.66     &  0.50      &  -               &   -       &     Er         &  Er       &     -         &    -     \\
\midrule
    \multicolumn{9}{l}{\textbf{Proposed Attacks}} \\
\midrule
     Dot Metric               & 0.67            & 0.54        &  0.65        & 0.50        &  0.61          &  0.46      &   0.74 (-26\%)       &  0.62       \\
     $\bm{\ell_2}$ Metric     & 0.64            & 0.50        &  0.66        &  0.52       &  0.59 (-41\%)  &  0.43       &  0.74 (-26\%)       & 0.62         \\
     Cos Metric               & 0.63 (-37\%)    & 0.49        & 0.63 (-37\%) & 0.47        &  0.60          &  0.43       &   0.74 (-26\%)      & 0.61         \\
\midrule
     GD (dot)           &  0.61 \textbf{(-39\%)}    & 0.45         &  0.65                   &  0.50       & 0.62           &  0.46       &  0.71 \textbf{(-29\%)}   &  0.58    \\
     GL ($\ell_2$)      &  0.63                     & 0.48         &  0.67                   &   0.53      &  0.61 (-39\%)  &  0.45       &   0.74                   &  0.60     \\
     GC (cos)           &  0.62                     &  0.46        &  0.64 \textbf{(-36\%)}  &  0.49       &  0.61 (-39\%)  &  0.45       &   0.71 \textbf{(-29\%)}  &   0.56     \\
\midrule
     IF       &  0.61 \textbf{(-39\%)}         &  0.45      &   0.65 (-35\%)             & 0.50        &  0.58 \textbf{(-42\%)}     & 0.42     &  0.71 \textbf{(-29\%)}  &   0.58    \\
\bottomrule    

\end{tabular}
\caption{\small Reduction in MRR and Hits@1 due to \textbf{adversarial additions on target triples in FB15k-237}. Lower values indicate better results; best results for each model are in bold. First block of rows are the baseline attacks with random edits; second block is state-of-the-art attacks; remaining are the proposed attacks. For each block, the \emph{best} reduction in percentage relative to the original MRR; computed as $(poisoned - original)/original*100$ is reported.}
\label{tab:ia_addition_FB15k-237}
\end{table}

\subsection{Comparison with Baselines}
\label{sec:ia_eval_baseline}
It is observed that the proposed strategies for adversarial deletions and adversarial additions successfully degrade the predictive performance of KGE models. 
On the other hand, the state-of-the-art attacks are ineffective or only partially effective. Adversarial deletions from Gradient Rollback perform similar to random baselines; likely because this method estimates the influence of a training triple as the sum of its gradients over the training process. In this way, it does not account for the target triple in the influence estimation. The method is also likely to be effective only for a KGE model that is trained with a batch size of 1 because it needs to track the gradient updates for each triple.

The CRIAGE baseline is only applicable to DistMult and ConvE. However, for the experiments here, the attack ran into \texttt{\small numpy.linalg.LinAlgError: Singular matrix} error for ConvE; because the Hessian matrix computed from the victim model embeddings was non-invertible\footnote{This issue might be resolved by changing the hyperparameters of the victim KGE model so that the Hessian matrix from the victim embeddings is invertible. But there is no strategic way to make such changes.}. For adversarial deletions on DistMult, the baseline works better than random edits but not the proposed attacks \footnote{Since the influence estimation in CRIAGE uses BCE loss, the attack performance was also compared for DistMult trained with BCE in Appendix \ref{apx:criage_bce}, but the results are similar.}.
It is also ineffective against adversarial additions. 

On the other hand, the Direct-Del is effective on TransE, but not on multiplicative models. 
This is likely because it estimates the influence of a candidate triple as the \emph{difference} in the triple's score when the neighbour entity embedding is perturbed. The additive nature of this influence score might make it more suitable for additive models. Furthermore, the Direct-Add works similar to random additions, likely because it uses random down-sampling.

The proposed attacks based on instance attribution methods consistently outperform random baselines for adversarial additions and deletions. One exception to this pattern are adversarial additions against TransE on WN18RR. In this case, no influence metric performs better than random neighbourhood edits, though they are all effective for adversarial deletions. One possible reason is that the TransE model is designed to learn hierarchical relations like $\mathtt{\_has\_part}$. On further investigation, it was found that the target triples ranked highest by the model have such hierarchical relations; and the influential triple for them has the same relation. That is, the triple $(s_1, \mathtt{\_has\_part}, s)$ is the influential triple for $(s, \mathtt{\_has\_part}, o)$. Removing this influential triple breaks the hierarchical link between $s_1$ and $s$; and degrades TransE predictions on the target. But adding the triple $(s_2, \mathtt{\_has\_part}, s)$ still preserves the hierarchical structure which TransE can use to score the target triple correctly.

The Instance Similarity method cos metric was selected for further analysis. It performs the best of all instance attribution methods for adversarial deletions, but performs worse than random neighbourhood edits for adversarial additions. 
Table \ref{tab:ia_wn18rr_transe} shows the relations in the target triples and the influential triples (i.e. adversarial deletions) selected by cos metric.
It is seen that the target triples contain mostly hierarchical relations like $\mathtt{\_synset\_domain\_topic\_of}$ and $\mathtt{\_has\_part}$. Also, the cos metric identifies influential triples with same relations. Since the proposed adversarial additions are only based on modifying the entity in the influential triple, these edits improve the hierarchy structure of the graph instead of breaking it. Thus, these edits perform well for adversarial deletions, but not for additions.

\begin{table}
    \centering
    \small
    \begin{tabular}{l l l}
    \toprule
        \textbf{Target Relation} &  \textbf{Influential Relation} \\
        \hline
       $\mathtt{has\_part}$  & $\mathtt{has\_part}$ \\
       $\mathtt{synset\_domain\_topic\_of}$ & $\mathtt{synset\_domain\_topic\_of}$ \\
       $\mathtt{has\_part}$ & $\mathtt{has\_part}$ \\
       $\mathtt{synset\_domain\_topic\_of}$   & $\mathtt{synset\_domain\_topic\_of}$  \\
       $\mathtt{synset\_domain\_topic\_of}$  &  $\mathtt{synset\_domain\_topic\_of}$   \\
       $\mathtt{synset\_domain\_topic\_of}$  & $\mathtt{synset\_domain\_topic\_of}$  \\
       $\mathtt{instance\_hypernym}$ &   $\mathtt{instance\_hypernym}$  \\
       $\mathtt{synset\_domain\_topic\_of}$   &  $\mathtt{synset\_domain\_topic\_of}$  \\
       $\mathtt{instance\_hypernym}$  &  $\mathtt{synset\_domain\_topic\_of}$ \\
       $\mathtt{synset\_domain\_topic\_of}$  &   $\mathtt{synset\_domain\_topic\_of}$ \\
       $\mathtt{member\_meronym}$  &  $\mathtt{derivationally\_related\_form}$   \\
       $\mathtt{synset\_domain\_topic\_of}$ &  $\mathtt{synset\_domain\_topic\_of}$  \\
       $\mathtt{has\_part}$  &  $\mathtt{has\_part}$  \\
       $\mathtt{member\_meronym}$   & $\mathtt{member\_meronym}$   \\
       $\mathtt{synset\_domain\_topic\_of}$   &  $\mathtt{synset\_domain\_topic\_of}$ \\
      \bottomrule
    \end{tabular}
    \caption{Relations from the target triples and influential triples (adversarial deletions) for the cos metric on WN18RR-TransE. This combination has 15 target triples and the table shows the relations for all of them.}
    \label{tab:ia_wn18rr_transe}
\end{table}

\subsection{Comparison across Instance Attribution Metrics }
\label{sec:ia_eval_if}
Among the different metrics for instance attribution, the IF and Gradient Similarity metrics show similar degradation in predictive performance. This indicates that the computationally expensive Hessian inverse in the IF can be avoided and simpler metrics can identify influential triples with comparable effectiveness. 
Furthermore, cos and $\bm{\ell_2}$ based Instance Similarity metrics outperform all other methods for adversarial deletions on DistMult, ComplEx and TransE. This effectiveness of naive metrics indicates the high vulnerability of shallow KGE architectures to data poisoning attacks in practice.
In contrast to this, the Input Similarity metrics are less effective in poisoning ConvE, especially significantly on WN18RR. This is likely because the triple feature vectors for ConvE are based on the output from a deeper neural architecture than the Embedding layer alone.
Within Instance Similarity metrics, it is observed that the dot metric is not as effective as others. This could be because the dot product does not normalize the triple feature vectors. Thus, training triples with large norms are prioritized over relevant influential triples \citep{hanawa2021evaluationsimilaritymetrics}.

\subsection{Comparison of Datasets}
\label{sec:ia_eval_data}
\begin{figure}[h]
    \centering
    \includegraphics[scale=1,width=0.7\textwidth]{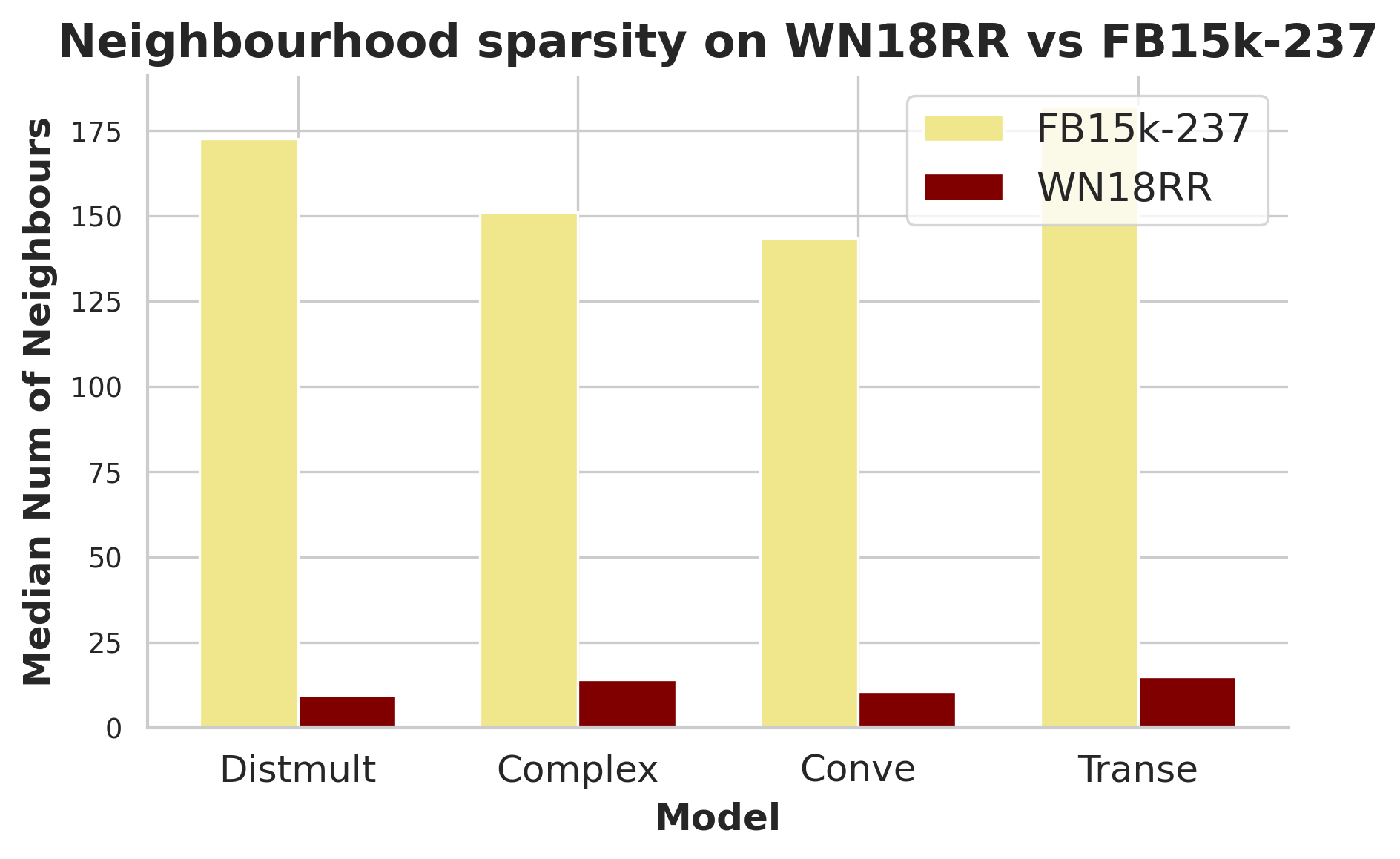}
    \caption{Comparison of the median number of neighbouring triples of target triples from WN18RR and FB15k-237 for DistMult, ComplEx, ConvE and TransE.}
    \label{fig:ia_neighbourhood_sparsity}
\end{figure}

Among the different datasets, it is noteworthy that the degradation in predictive performance is more significant on WN18RR than on FB15k-237. This is likely due to the sparser graph structure of WN18RR, i.e. there are fewer neighbours per target triple in WN18RR than in FB15k-237. The graph in Figure \ref{fig:ia_neighbourhood_sparsity} shows the median number of neighbours of the target triples for WN18RR and FB15k-237. The median value is reported instead of mean because of a large standard deviation in the number of target triple neighbours for FB15k-237.

Based on this graph, the target triple's neighbourhood for WN18RR is significantly sparser than the neighbourhood for FB15k-237.
Since the KGE model predictions are learned from fewer triples for WN18RR, it is also easier to perturb these results with fewer adversarial edits. Thus, removing only 1 neighbour in WN18RR significantly degrades the model's predictions on the target triple.

On the other hand, because of more neighbours in FB15k-237, the model predictions are likely influenced by a \emph{group} of training triples. Such group effect of training instances on the model parameters has been studied in \citet{koh2019groupinfluence} and \citet{basu2020groupinfluence}. These methods will be investigated for \emph{KGE models} on FB15k-237 as part of the future work.


\section{Additional Evaluation}
\subsection{Analysis of Runtime Efficiency}
\label{sec:ia_runtime_analysis}

In addition to the effectiveness in the degradation of KGE model performance, the runtime efficiency of baseline and proposed attack methods for adversarial deletions is also analyzed. 
For all the target triples in the test set, the triples in their neighbourhood are pre-computed. Given the original KGE model, the target triples and their neighbourhood, the absolute time taken to select the influential triples for all the target triples is recorded. Table \ref{tab:ia_runtime} shows the time taken in seconds to select the complete set of adversarial deletions for DistMult model on WN18RR and FB15k-237. For brevity, only the attacks on DistMult model are reported, but the results on other models show similar time scales.

\begin{table}[h]
\centering
\small
\begin{tabular}{c  l  ll  }

\toprule
     & & \textbf{WN18RR} & \textbf{FB15k-237} \\
   
\midrule
    \multirow{6}{*}{\shortstack[l]{\textbf{Baseline} \\ \textbf{Attacks}}}
    & Random\_n              &  0.024    & 0.057         \\
    & Random\_g              &  0.002     &  0.002        \\
\cline{2-4} \\[-12pt]
    & Direct-Del            &  0.407     &  0.272          \\
    & CRIAGE                 &  2.235     &  75.117       \\
    & GR     & 29.919     &  174.191        \\
\midrule
    \multirow{7}{*}{\shortstack[l]{\textbf{Proposed} \\ \textbf{Attacks}}}
    & Dot Metric               & 0.288            & 0.342              \\
    & $\bm{\ell_2}$ Metric     & 0.057            & 0.067               \\
    & Cos Metric               & 0.067            & 0.148        \\
\cline{2-4} \\[-12pt] 
    & GD (dot)           & 7.354           & 109.015               \\
    & GL ($\bm{\ell_2}$)           & 8.100           & 120.659                      \\
    & GC (cos)           &  9.478           &  141.276                    \\
\cline{2-4} \\[-12pt]
    
    & IF       &  4751.987           &  4750.404                 \\
     
\bottomrule    

\end{tabular}
\caption{Runtime efficiency of the baseline and proposed adversarial attacks for DistMult on WN18RR and FB15k-237. The absolute time taken in seconds to generate the complete set of influential triples for all target triples is reported. The neighbourhood triples of the target triples were pre-computed. The runtime for GR does not include the time taken to compute the influence map during model training.}
\label{tab:ia_runtime}
\end{table}

The time scales show that the Instance Similarity metrics (dot metric, $\bm \ell_2$ metric, cos metric) are more efficient than the state-of-the-art attacks (Direct-Del, CRIAGE and GR). Furthermore, the $\bm \ell_2$ metric is almost as quick as random triple selection.
The efficiency of the Gradient Similarity metrics is also better than or equivalent to CRIAGE and GR.
Only the attack method based on IF is much slower than any other method. This is because estimating the Hessian inverse in IF requires one training epoch for every target triple. That is, for the reported experiments, 100 training epochs are done to get the influential triples for 100 target triples. However, based on the results in Section \ref{sec:ia_eval_if}, this expensive computation does not provide improved adversarial deletions, and thus, might be unnecessary to select the influential triples for KGE models.

Comparing the efficiency of attribution attacks for different datasets, the proposed methods are more efficient for the WN18RR dataset than FB15k-237. Since the same number of target triples were used for both datasets (i.e. 100), the difference in efficiency is explained by the difference in neighbourhood sizes of WN18RR and FB15k-237. As Figure \ref{fig:ia_neighbourhood_sparsity} from the previous section shows, the neighbourhood of a target triple in FB15k-237 contains more triples than in WN18RR. Thus, enumerating the neighbours to select the adversarial deletions in WN18RR is more efficient than FB15k-237. The difference in efficiency is more evident for the gradient similarity based attribution methods than the instance similarity based methods. This is because the software implementation for instance similarity methods processes the entire neighbourhood of each triple as a single batch. On the other hand, gradient based methods require enumerating the neighbours to compute the gradient vector for each neighbour individually. Thus, the effect of number of neighbours is more evident for gradient similarity methods than for instance similarity. The reason for similar runtime performance of IF for both datasets is not immediately evident and might be attributed to the hyperparameter setting for computing the IF values. However, for models other than DistMult, there is an observable difference between the IF efficiency for WN18RR and FB15k-237, with better runtime for WN18RR. These values in seconds are -- $(1305.315\ , 5894.426)$ for ComplEx, $(12513.344\ , 17433.215)$ for ConvE and $(2613.325\ , 7384.060)$ for TransE.

\subsection{Qualitative Analysis}
\label{sec:ia_qualitative}
\begin{table}[]
    \centering
    \small
     \setlength{\tabcolsep}{6.0pt}
    \begin{tabularx}{0.9\textwidth}{l >{\raggedright\arraybackslash}X}
    \toprule
       \textbf{Attack}  &   \multicolumn{1}{c}{\textbf{Target and Influential Triples}}  \\
    \midrule
    \multirow{2}{*}{\shortstack[l]{\textbf{Dot} \\ \textbf{Metric}}}
       &     \emph{dynamical\_JJ\_1} ,\  \emph{similar\_to} ,\  \emph{hold-down\_NN\_1}  \\
       &     \emph{hold-down\_NN\_1} ,\  \emph{similar\_to} ,\  \emph{dynamical\_JJ\_1}  \\ [7pt]
       
    \multirow{2}{*}{\shortstack[l]{$\bm{\ell_2}$ \\ \textbf{Metric}}}
       &    \emph{departed\_NN\_1} ,\  \emph{derivationally\_related\_form} ,\  \emph{snuff\_it\_VB\_1}  \\
       &    \emph{snuff\_it\_VB\_1} ,\  \emph{derivationally\_related\_form} ,\  \emph{departed\_NN\_1}  \\ [7pt]
    
    \multirow{2}{*}{\shortstack[l]{\textbf{Cos} \\ \textbf{Metric}}}
       &    \emph{level\_NN\_6} ,\  \emph{derivationally\_related\_form} ,\  \emph{tear\_down\_VB\_1}  \\
       &    \emph{tear\_down\_VB\_1} ,\  \emph{derivationally\_related\_form} ,\  \emph{level\_NN\_6}  \\
    
    \midrule
    
    \multirow{2}{*}{\shortstack[l]{\textbf{GD} \\ \textbf{(dot)}}}
       &    \emph{personal\_identity\_NN\_1} ,\  \emph{derivationally\_related\_form} ,\  \emph{place\_VB\_8}  \\
       &    \emph{place\_VB\_8} ,\  \emph{derivationally\_related\_form} ,\  \emph{personal\_identity\_NN\_1}  \\ [7pt]
       
    \multirow{2}{*}{\shortstack[l]{\textbf{GL} \\ ($\bm{\ell_2}$)}}
       &    \emph{dynamitist\_NN\_1} ,\  \emph{derivationally\_related\_form} ,\  \emph{dynamite\_NN\_1}  \\
       &    \emph{dynamite\_NN\_1} ,\  \emph{derivationally\_related\_form} ,\  \emph{dynamitist\_NN\_1}  \\ [7pt]
       
    \multirow{2}{*}{\shortstack[l]{\textbf{GC} \\ \textbf{(cos)}}}
       &    \emph{departed\_NN\_1} ,\  \emph{derivationally\_related\_form} ,\  \emph{snuff\_it\_VB\_1}  \\
       &    \emph{snuff\_it\_VB\_1} ,\  \emph{derivationally\_related\_form} ,\  \emph{departed\_NN\_1}  \\
       
    \midrule
       
    \multirow{2}{*}{ \textbf{IF}}
       &    \emph{departed\_NN\_1} ,\  \emph{derivationally\_related\_form} ,\  \emph{snuff\_it\_VB\_1}  \\
       &    \emph{snuff\_it\_VB\_1} ,\  \emph{derivationally\_related\_form} ,\  \emph{departed\_NN\_1}  \\
       
    \bottomrule
    \end{tabularx}
    \caption{Examples of target triples from WN18RR ComplEx with maximum change in rank due to adversarial deletions, and their corresponding influential triples.}
    \label{tab:ia_examples_wn18rr_complex_max}
\end{table}

This section provides a qualitative analysis of the influential triples selected using the different instance attribution methods for the ComplEx model trained on WN18RR and FB15k-237. Tables \ref{tab:ia_examples_wn18rr_complex_max} and \ref{tab:ia_examples_wn18rr_complex_min} show these examples for target triples that exhibit the maximum and minimum change in the ranks respectively due to the adversarial deletions for WN18RR. Since the entities in the benchmark splits are represented as numerical IDs, their string representation are obtained using the \emph{definitions.txt} file. This file is available for download from the original website of WN18RR - \url{https://everest.hds.utc.fr/doku.php?id=en:smemlj12}. The values for maximum change in ranks due to different attacks for WN18RR are -- $(35962.5, 6653.5, 39971.5)$ due to instance similarity metrics, $(28011, 37084.5, 36741)$ due to gradient similarity metrics and $39978.5$ due to Influence Function.
Tables \ref{tab:ia_examples_fb15k_complex_max} and \ref{tab:ia_examples_fb15k_complex_min} focus on the target triples with maximum and minimum change in ranks respectively for FB15k-237. The IDs in the benchmark split are converted into string representations using the browser at  \url{https://freebase.toolforge.org/}. The maximum change in ranks is -- $(38.5, 31, 80)$ due to instance similarity metrics, $(61.5, 114.5, 149.5)$ due to gradient similarity metrics and $114.5$ due to Influence Function. The minimum change in ranks for both datasets is $0$.

Among the target triples with maximum change for WN18RR (Table \ref{tab:ia_examples_wn18rr_complex_max}), $\bm{\ell_2}$ metric and GC (cos) behave similar to Influence Function. These methods identify the same influential triple (\emph{snuff\_it\_VB\_1} ,\  \emph{derivationally\_related\_form} ,\  \emph{departed\_NN\_1}) for the same target triple (\emph{departed\_NN\_1} ,\  \emph{derivationally\_related\_form} ,\  \emph{snuff\_it\_VB\_1}). The entity \emph{departed\_NN\_1} is the noun form to refer to a person who is no longer alive, while the entity \emph{snuff\_it\_VB\_1} refers to the act of passing away from physical life. The two words have similar semantic meaning but different parts-of-speech. Thus, it makes sense for them to be linked by the relation \emph{derivationally\_related\_form}. This indicates that Instance and Gradient Similarity metrics are effective approximations of Influence Function as they are able to identify an influential triple correctly.

Among the triples with no change in ranks for WN18RR (Table \ref{tab:ia_examples_wn18rr_complex_min}), GL ($\bm{\ell_2}$) identifies (\emph{the\_netherlands\_NN\_1} ,\  \emph{derivationally\_related\_form} ,\  \emph{netherlander\_NN\_1}) as the influential triple for (\emph{north\_atlantic\_treaty\_organization\_NN\_1} ,\  \emph{member\_meronym} ,\  \emph{the\_netherlands\_NN\_1}). The target triple indicates that the country Netherlands is a member of NATO. However, the identified influential triple does not provide meaningful attribution for this target prediction, and is thus, ineffective. Some other training triples like (\emph{north\_atlantic\_treaty\_organization\_NN\_1} ,\  \emph{member\_meronym} ,\  \emph{italy\_NN\_1}) or (\emph{europe\_NN\_1} ,\  \emph{has\_part} ,\  \emph{the\_netherlands\_NN\_1}) or (\emph{europe\_NN\_1} ,\  \emph{has\_part} ,\  \emph{italy\_NN\_1}) would be more meaningful influential triples for this target. On the other hand, some influential triples identified by the attribution methods are meaningful, but still ineffective in degrading the target prediction. Consider the GC (cos) attribution method with target triple (\emph{love\_VB\_3} ,\  \emph{derivationally\_related\_form} ,\  \emph{passion\_NN\_6}). While the influential triple (\emph{passion\_NN\_6} ,\  \emph{derivationally\_related\_form} ,\  \emph{love\_VB\_3}) is meaningful, it is not effective. This is because the words 'love' and 'passion' have multiple meanings which are represented as different entities in the knowledge graph. For example, there are 3 entities \emph{love\_VB\_1}, \emph{love\_VB\_2}, \emph{love\_VB\_3} for love, and 2 entities \emph{passion\_NN\_4}, \emph{passion\_NN\_6} for passion. Furthermore, these entities are linked to each other via \emph{derivationally\_related\_form}. Thus, the target triple's prediction is influenced by multiple training triples, and deleting one influential triple is insufficient to perturb its prediction.

\begin{table}[h]
    \centering
    \small
     \setlength{\tabcolsep}{6.0pt}
    \begin{tabularx}{0.9\textwidth}{l >{\raggedright\arraybackslash}X}
    \toprule
       \textbf{Attack}  &   \multicolumn{1}{c}{\textbf{Target and Influential Triples}}  \\
    \midrule
    \multirow{2}{*}{\shortstack[l]{\textbf{Dot} \\ \textbf{Metric}}}
       &     \emph{tug\_VB\_2} ,\  \emph{derivationally\_related\_form} ,\  \emph{undertaking\_NN\_1}  \\
       &     \emph{toil\_NN\_1} ,\  \emph{derivationally\_related\_form} ,\  \emph{tug\_VB\_2}  \\ [7pt]
       
    \multirow{2}{*}{\shortstack[l]{$\bm{\ell_2}$ \\ \textbf{Metric}}}
       &    \emph{genus\_alnus\_NN\_1} ,\  \emph{member\_meronym} ,\  \emph{red\_alder\_NN\_1}  \\
       &    \emph{red\_alder\_NN\_1} ,\  \emph{hypernym} ,\  \emph{alder\_tree\_NN\_1}  \\ [7pt]
    
    \multirow{2}{*}{\shortstack[l]{\textbf{Cos} \\ \textbf{Metric}}}
       &    \emph{suddenness\_NN\_1} ,\  \emph{derivationally\_related\_form} ,\  \emph{sudden\_JJ\_1}  \\
       &    \emph{sudden\_JJ\_1} ,\  \emph{derivationally\_related\_form} ,\  \emph{suddenness\_NN\_1}  \\
    
    \midrule
    
    \multirow{2}{*}{\shortstack[l]{\textbf{GD} \\ \textbf{(dot)}}}
       &    \emph{stock\_VB\_2} ,\  \emph{derivationally\_related\_form} ,\  \emph{stock\_NN\_3}  \\
       &    \emph{stock\_NN\_3} ,\  \emph{hypernym} ,\  \emph{hold\_NN\_8}  \\ [7pt]
       
    \multirow{2}{*}{\shortstack[l]{\textbf{GL} \\ ($\bm{\ell_2}$)}}
       &    \emph{north\_atlantic\_treaty\_organization\_NN\_1} ,\  \emph{member\_meronym} ,\  \emph{the\_netherlands\_NN\_1}  \\
       &    \emph{the\_netherlands\_NN\_1} ,\  \emph{derivationally\_related\_form} ,\  \emph{netherlander\_NN\_1}  \\ [7pt]
       
    \multirow{2}{*}{\shortstack[l]{\textbf{GC} \\ \textbf{(cos)}}}
       &    \emph{love\_VB\_3} ,\  \emph{derivationally\_related\_form} ,\  \emph{passion\_NN\_6}  \\
       &    \emph{passion\_NN\_6} ,\  \emph{derivationally\_related\_form} ,\  \emph{love\_VB\_3}  \\ 
       
    \midrule
       
    \multirow{2}{*}{ \textbf{IF}}
       &    \emph{suddenness\_NN\_1} ,\  \emph{derivationally\_related\_form} ,\  \emph{sudden\_JJ\_1}  \\
       &    \emph{fast\_JJ\_1} ,\  \emph{also\_see} ,\  \emph{sudden\_JJ\_1}  \\
       
    \bottomrule
    \end{tabularx}
    \caption{Examples of target triples from WN18RR ComplEx with no change in rank due to adversarial deletions, and their corresponding influential triples.}
    \label{tab:ia_examples_wn18rr_complex_min}
\end{table}

For FB15k-237 dataset, it can be observed that some of the target triples themselves are unintuitive. For example, from Table \ref{tab:ia_examples_fb15k_complex_max}, the target triple (\emph{University of Louisiana at Monroe}  ,   \emph{/common/ topic/ webpage./ common/ webpage/ category}    ,  \emph{Official Website}) indicates that the University of Louisiana (a topic) has a webpage which is the official website for this topic. Similarly, the triple (\emph{The Reader}   ,   \emph{/film/ film/ other\_crew./ film/ film\_crew\_gig/ film\_crew\_role}    ,  \emph{Sound Mixer}) indicates that the film The Reader has some crewmembers whose job is sound mixing. However, instead of linking the crew member's role to an entity for the member, the role is linked directly to the movie. Such unintuitive relation links exist in FB15k-237 because the original schema of Freebase has Compound Value Type (CVT) entities \citep{tanon2016freebasetowikidata} which represent n-ary relations between normal entities. These CVT nodes were removed while preparing the flat graph structure for FB15k and FB15k-237 \citep{bordes2013transe}. Thus, the entities linked to CVT nodes in original Freebase data are linked directly in FB15k-237. Further, the multiple hop links between these entities are concatenated into single relations. This leads to semantically incoherent triples, whose predictions are difficult to explain. In addition to the impact of number of neighbours on attack performance (Section \ref{sec:ia_eval_data}), the incoherence of triples in FB15k-237 is also likely responsible for lower efffectiveness of the instance attribution methods for this dataset. 

On the other hand, looking at target triples without CVT based relations, some influential triples being identified are meaningful, but ineffective in degrading the target triple's rank. Consider the attacks Cos metric and GL ($\bm{\ell_2}$) from Table \ref{tab:ia_examples_fb15k_complex_min}. For the target triple (\emph{Harry Potter and the Half Blood Prince}   ,   \emph{/film/ film/ story\_by}    ,  \emph{JK Rowling}), the influential triples being selected are (\emph{Harry Potter and the Order of the Phoenix} ,  \emph{/film/ film/ story\_by} ,  \emph{JK Rowling}) and (\emph{Harry Potter and the Half Blood Prince} ,  \emph{/film/ film/ prequel} ,  \emph{Harry Potter and the Order of the Phoenix}). Both of these are meaningful influential triples for the target triple, but do not lead to a degradation of rank. This is explained by the training graph for movies \emph{Harry Potter and the Half Blood Prince} and \emph{Harry Potter and the Order of the Phoenix} - both of them share the same acting cast, have the same genres and have been nominated for similar awards. Thus, the prediction for the target triple is not influenced by a single training triple, but multiple ones. This observation motivates the need to define and investigate methods to determine the influence of multiple triples collectively instead of individual triples separately.

\begin{table}
    \centering
     \fontsize{10}{10}\selectfont
     \setlength{\tabcolsep}{4.0pt}
  \begin{tabular}{  m{0.1\textwidth}  >{\raggedright\arraybackslash}p{0.16\textwidth}  >{\raggedright\arraybackslash}p{0.5\textwidth} >{\raggedright\arraybackslash}p{0.16\textwidth}  } 
  \toprule
    \textbf{Attack} & \multicolumn{3}{c}{\textbf{Target and Influential Triples}} \\ 
    \midrule
    \multirow{2}{*}{\shortstack[l]{\textbf{Dot} \\ \textbf{Metric}}}
    &   \emph{The Reader}   &   \emph{/film/ film/ other\_crew./ film/ film\_crew\_gig/ film\_crew\_role}    &  \emph{Sound Mixer}   \\
    &     \emph{The Devil's Double}  &  \emph{/film/ film/ other\_crew./ film/ film\_crew\_gig/ film\_crew\_role}  &  \emph{Sound Mixer} \\
    
    &   & &   \\
    
    \multirow{2}{*}{\shortstack[l]{$\bm{\ell_2}$ \\ \textbf{Metric}}}
    &   \emph{United States of America}   &   \emph{/location/ country/ second\_level\_divisions}    &  \emph{Pasco County}   \\
    &     \emph{Pasco County} &  \emph{/location/ statistical\_region/ rent50\_2./ measurement\_unit/ dated\_money\_value/ currency} &  \emph{United States Dollar} \\
    
    &   & &   \\
    
    \multirow{2}{*}{\shortstack[l]{\textbf{Cos} \\ \textbf{Metric}}}
    &   \emph{Piper Laurie}   &   \emph{/people/ person/ profession}    &  \emph{Actor}   \\
    &     \emph{Patty Duke} &  \emph{/people/ person/ profession} &  \emph{Actor} \\
    
    \midrule
    \multirow{2}{*}{\shortstack[l]{\textbf{GD} \\ \textbf{(dot)}}}
    &   \emph{University of Louisiana at Monroe}   &   \emph{/common/ topic/ webpage./ common/ webpage/ category}    &  \emph{Official Website}   \\
    &     \emph{United States of America} &  \emph{/location/ location/ contains} &  \emph{University of Louisiana at Monroe} \\
    
    &   & &   \\
    
    \multirow{2}{*}{\shortstack[l]{\textbf{GL} \\ ($\bm{\ell_2}$)}}
    &   \emph{University of Louisiana at Monroe}   &   \emph{/common/ topic/ webpage./ common/ webpage/ category}    &  \emph{Official Website}   \\
    &     \emph{University of Louisiana at Monroe} &  \emph{/education/ university/ local\_tuition./ measurement\_unit/ dated\_money\_value/ currency} &  \emph{United States Dollar} \\
    
    &   & &   \\
    
    \multirow{2}{*}{\shortstack[l]{\textbf{GC} \\ \textbf{(cos)}}}
    &   \emph{Piper Laurie}   &   \emph{/people/ person/ profession}    &  \emph{Actor}   \\
    &     \emph{Piper Laurie} &  \emph{/people/ person/ spouse\_s./ people/ marriage/ type\_of\_union} &  \emph{Marriage} \\
    
    \midrule
       
    \multirow{2}{*}{ \textbf{IF}}
    &   \emph{University of Louisiana at Monroe}   &   \emph{/common/ topic/ webpage./ common/ webpage/ category}    &  \emph{Official Website}   \\
    &     \emph{United States of America} &  \emph{/location/ location/ contains} &  \emph{University of Louisiana at Monroe} \\

    \bottomrule
  \end{tabular}
  \caption{Examples of target triples from FB15k-237 ComplEx with maximum change in rank due to adversarial deletions, and their corresponding influential triples.}
    \label{tab:ia_examples_fb15k_complex_max}
\end{table}

\begin{table}
    \centering
     \fontsize{10}{10}\selectfont
     \setlength{\tabcolsep}{4.0pt}
  \begin{tabular}{  m{0.1\textwidth}  >{\raggedright\arraybackslash}p{0.2\textwidth}  >{\raggedright\arraybackslash}p{0.4\textwidth} >{\raggedright\arraybackslash}p{0.2\textwidth}  } 
  \toprule
    \textbf{Attack} & \multicolumn{3}{c}{\textbf{Target and Influential Triples}} \\ 
    \midrule
    \multirow{2}{*}{\shortstack[l]{\textbf{Dot} \\ \textbf{Metric}}}
    &   \emph{Pompano Beach}   &   \emph{/location/ hud\_county\_place/ place}    &  \emph{Pompano Beach}   \\
    &     \emph{Pompano Beach}  &  \emph{/location/hud\_foreclosure\_area/ estimated\_number\_of\_mortgages./ measurement\_unit/ dated\_integer/ source}  &  \emph{United States Department of Housing and Urban Development} \\
    
    &   & &   \\
    
    \multirow{2}{*}{\shortstack[l]{$\bm{\ell_2}$ \\ \textbf{Metric}}}
    &   \emph{Indian Idol}   &   \emph{/tv/ tv\_program/ country\_of\_origin}    &  \emph{India}   \\
    &     \emph{Shatrughan Sinha} &  \emph{/people/ person/ nationality} &  \emph{India} \\
    
    &   & &   \\
    
    \multirow{2}{*}{\shortstack[l]{\textbf{Cos} \\ \textbf{Metric}}}
    &   \emph{Harry Potter and the Half Blood Prince}   &   \emph{/film/ film/ story\_by}    &  \emph{JK Rowling}   \\
    &     \emph{Harry Potter and the Order of the Phoenix} &  \emph{/film/ film/ story\_by} &  \emph{JK Rowling} \\
    
    \midrule
    \multirow{2}{*}{\shortstack[l]{\textbf{GD} \\ \textbf{(dot)}}}
    &   \emph{Indian Idol}   &   \emph{/tv/ tv\_program/ country\_of\_origin}    &  \emph{India}   \\
    &     \emph{Malayali} &  \emph{/people/ ethnicity/ geographic\_distribution} &  \emph{India} \\
    
    &   & &   \\
    
    \multirow{2}{*}{\shortstack[l]{\textbf{GL} \\ ($\bm{\ell_2}$)}}
    &   \emph{Harry Potter and the Half Blood Prince}   &   \emph{/film/ film/ story\_by}    &  \emph{JK Rowling}   \\
    &     \emph{Harry Potter and the Half Blood Prince} &  \emph{/film/ film/ prequel} &  \emph{Harry Potter and the Order of the Phoenix} \\
    
    &   & &   \\
    
    \multirow{2}{*}{\shortstack[l]{\textbf{GC} \\ \textbf{(cos)}}}
    &   \emph{Reed College}   &   \emph{/education/ educational\_institution\_campus/ educational\_institution}    &  \emph{Reed College}   \\
    &     \emph{Reed College} &  \emph{/organization/ organization/ headquarters./ location/ mailing\_address/ citytown} &  \emph{Portland} \\
    
    \midrule
       
    \multirow{2}{*}{ \textbf{IF}}
    &   \emph{Indian Idol}   &   \emph{/tv/ tv\_program/ country\_of\_origin}    &  \emph{India}   \\
    &    \emph{Malayali} &  \emph{/people/ ethnicity/ geographic\_distribution} &  \emph{India} \\

    \bottomrule
  \end{tabular}
  \caption{Examples of target triples from FB15k-237 ComplEx with no change in rank due to adversarial deletions, and their corresponding influential triples.}
    \label{tab:ia_examples_fb15k_complex_min}
\end{table}


\section{Summary}
This chapter proposed the use of instance attribution methods to select adversarial deletions against the KGE models. Additionally, using the influential triples identified from the instance attribution methods, a heuristic approach was proposed to select adversarial additions. 
The evaluation of the effectiveness of the proposed attacks showed that they outperform the baseline attacks on two benchmark datasets against four KGE models. It was also observed that the attacks are particularly effective when the KGE model relies on fewer training instances to make predictions, i.e. when the input graph is sparse. For the sparse graphs, the proposed attacks are also more efficient than highly connected graphs because they need to enumerate through a smaller set of training triples to select the influential triple.
Additionally, shallow neural architectures like DistMult, ComplEx and TransE are vulnerable to simpler attacks based on Instance Similarity than the computationally expensive Influence Function.

While in this chapter, the instance attribution methods were used to select both adversarial deletions and additions, the proposed method for adversarial additions is heuristic in nature. This is because it overcomes the challenge of enumerating through all possible adversarial additions, but still relies on enumerating the neighbourhood triples to select the influential triple. 
The next chapter proposes an attack strategy for adversarial additions that breaks down the large combinatorial search space of adversarial additions into smaller search steps over the entities and relations in the knowledge graph.
To achieve this objective, instead of specifying the adversarial attack objective in terms of the model predictions for the target triples, a reformulation of the attack objective is proposed. This allows the search space to be reduced to three smaller steps.

\chapter{Poisoning via Relation Inference Patterns}
\label{ch:relation_inference}
This chapter proposes methods to select adversarial additions against the KGE models. As discussed in Section \ref{sec:problem_attack_design}, in addition to the design requirement for the impact of candidate perturbations, adversarial additions need to account for the large combinatorial search space. 
This search space is unavoidable when the adversarial attack objective is defined in terms of the KGE model's predictions for the target facts. This objective formulation requires the impact of a perturbation to be quantified explicitly. However, the impact of a perturbation on the ranks predicted for target facts can also be defined implicitly through the impact on synthetic negative triples against which the target fact is ranked. This chapter formulates the adversarial attack objective of degrading the predicted ranks of target facts as the task of improving the predicted ranks of decoy triples.

The notation $(s,\mathtt{r},o)$ is used for the \emph{target triple}; in this case, $s,o$ are the \emph{target entities} and $\mathtt{r}$ is the \emph{target relation}. 
The rank of a highly plausible target triple can be degraded by improving the rank of less plausible \emph{decoy triples}. For a target triple $(s,\mathtt{r},o)$, the decoy triple for degrading the rank on object side would be $(s,\mathtt{r},o')$ and the decoy triple for degrading the rank on subject side would be $(s',\mathtt{r},o)$. 
Thus, the aim of the adversarial attacker in this chapter, is to select decoy triples from the set of valid synthetic negatives and craft \emph{adversarial additions} to improve their ranks. The attacker does not add the decoy triple itself as an adversarial edit, rather chooses the adversarial edits that would improve the rank of a missing decoy triple through an inference pattern. The majority of contents in this chapter are taken verbatim from the author's publication \citet{bhardwaj2021relationinferencepatterns}.

\section{Threat Model}
The threat model for this chapter is the same as the threat model defined in Section \ref{sec:problem_threatmodel} and the prior state-of-the-art attacks against KGE models \citep{pezeshkpour2019criage, zhang2019kgeattack}.
The attacks are designed for a white-box attack setting. In this attack setting, the attacker has full knowledge of the target KGE model and the training dataset \citep{joseph_nelson_rubinstein_tygar_2019}. However, they cannot manipulate the model architecture or the learned embeddings directly; but only through \emph{addition} of triples to the training knowledge graph. %
Additionally, the attacker is restricted to making edits only in the neighbourhood of target entities. They are also restricted to 1 decoy triple for each entity of the target triple.  
Furthermore, because of the use of filtered settings for KGE evaluation \citep{bordes2013transe}, the attacker cannot add the decoy triple itself to the training data (which intuitively would be a way to improve the decoy triple's rank).


\section{Relation Inference Patterns}
As discussed in Section \ref{sec:kge_inductive}, the general intuition behind the design of the scoring functions of KGE models is to capture logical properties between relations from the observed facts in the knowledge graph. These logical properties or \emph{relation inference patterns} can then be used to make downstream inferences about entities and relations. For example, the relation $ \mathtt{is\_owned\_by}$ is inverse of the relation $ \mathtt {owns}$, and when the fact $(Account,\mathtt{is\_owned\_by}, Karl)$ is true, then the fact $(Karl,\mathtt {owns}, Account)$ is also true and vice versa. A model that can capture inversion pattern can thus predict missing facts about $\mathtt {owns}$ based on observed facts about $\mathtt{is\_owned\_by}$. The most studied inference patterns in the current literature are symmetry, inversion and composition since they occur very frequently in real-world knowledge graphs \citep{abboud2020boxe, ali2021bringinglightintodark}. In this chapter, the relation inference patterns are used to improve the ranks of the decoy triples and in turn, degrade the ranks of the target triples.

Since the inference patterns on the knowledge graph specify a logic property between the relations, they can be expressed as Horn Clauses which is a subset of the First Order Logic (FOL) formulae. For example, a property represented in the form  $\forall x,y : (x, \mathtt{owns}, y) \Rightarrow (y, \mathtt{is\_owned\_by}, x)$ means that two entities linked by the relation $\mathtt{owns}$ are also likely to be linked by the inverse relation $\mathtt{is\_owned\_by}$. In this expression, the right hand side of the implication $\Rightarrow$ is referred to as the \emph{head} and the left hand side as the \emph{body} of the clause. 
Using such expressions, the three inference patterns used in this research are defined as follows.

\begin{definition}
    The \emph{symmetry} pattern $\mathcal{P}_s$ is expressed as \(
        \forall x,y : (x, \mathtt{r}, y) \Rightarrow (y, \mathtt{r}, x)
    \). Here, the relation $\mathtt{r}$ is symmetric relation.
\end{definition}
\begin{definition}
    The \emph{inversion} pattern $\mathcal{P}_i$ is expressed as 
    \(
        \forall x,y : (x, \mathtt{r_i}, y) \Rightarrow (y, \mathtt{r}, x)
    \). Here, the relations $\mathtt{r_i}$ and $\mathtt{r}$ are inverse of each other.
\end{definition}
\begin{definition}
    The \emph{composition} pattern $\mathcal{P}_c$ is expressed as 
    \(
        \forall x,y,z : (x, \mathtt{r_1}, z) \wedge (z, \mathtt{r_2}, y) \Rightarrow (x, \mathtt{r}, y)
    \). Here, the relation $\mathtt{r}$ is a composition of $\mathtt{r_1}$ and $\mathtt{r_2}$ ; and the $\wedge$ is the conjunction operator from relational logic.
\end{definition}

The mapping \(\mathcal{G}: \mathcal{V \rightarrow \mathcal{E}} \) of variables $\mathcal{V}$ in the above expressions to entities $\mathcal{E}$ is called a grounding.
For example, the logic expression $\forall x,y : (x, \mathtt{owns}, y) \Rightarrow (y, \mathtt{is\_owned\_by}, x)$ can be mapped to the grounding $(Karl, \mathtt{owns}, Account) \Rightarrow (Account, \mathtt{is\_owned\_by}, Karl)$. Thus, a KGE model that captures the inversion pattern will assign a high prediction confidence to the head atom when the body of the clause exists in the knowledge graph.

\section{Steps for Adversarial Attack}
\begin{figure}[]
    \centering
    \begin{subfigure}[htb]{0.75\textwidth}
        \includegraphics[width=1\textwidth]{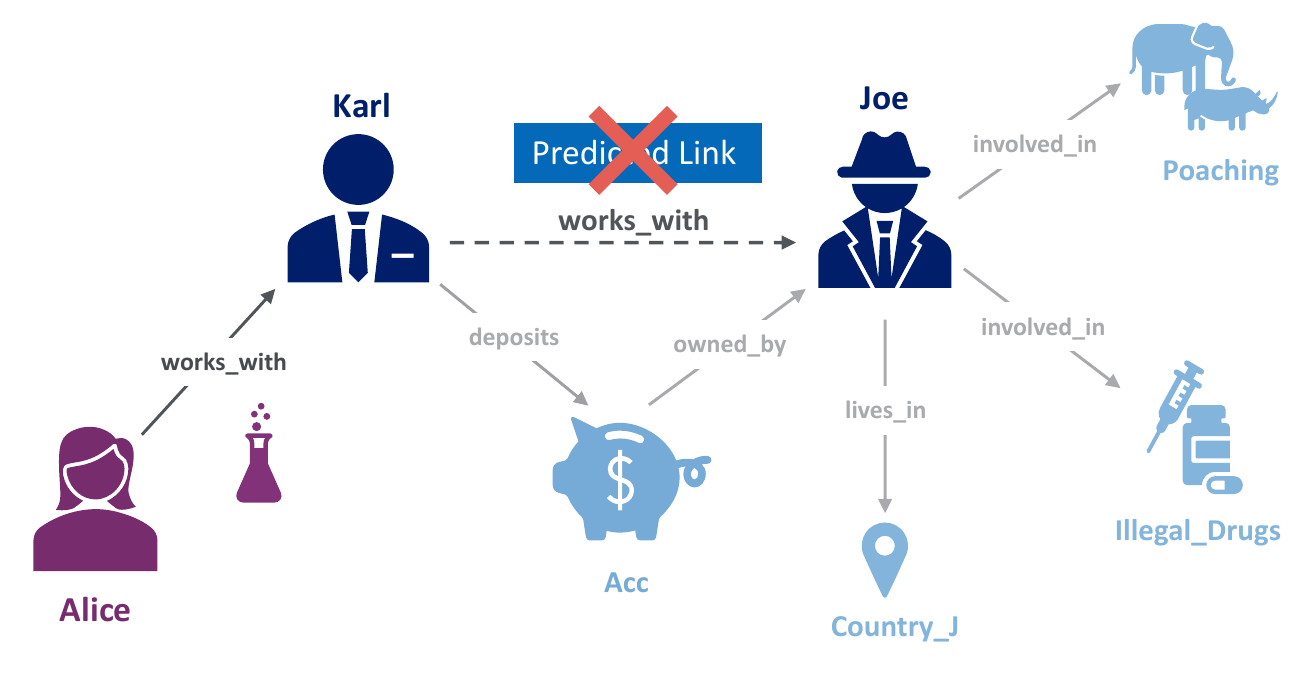}
    \caption{Symmetry Attack}
    \end{subfigure}
    \newline
    \newline
    \newline
    \begin{subfigure}[htb]{0.75\textwidth}
        \includegraphics[width=1\textwidth]{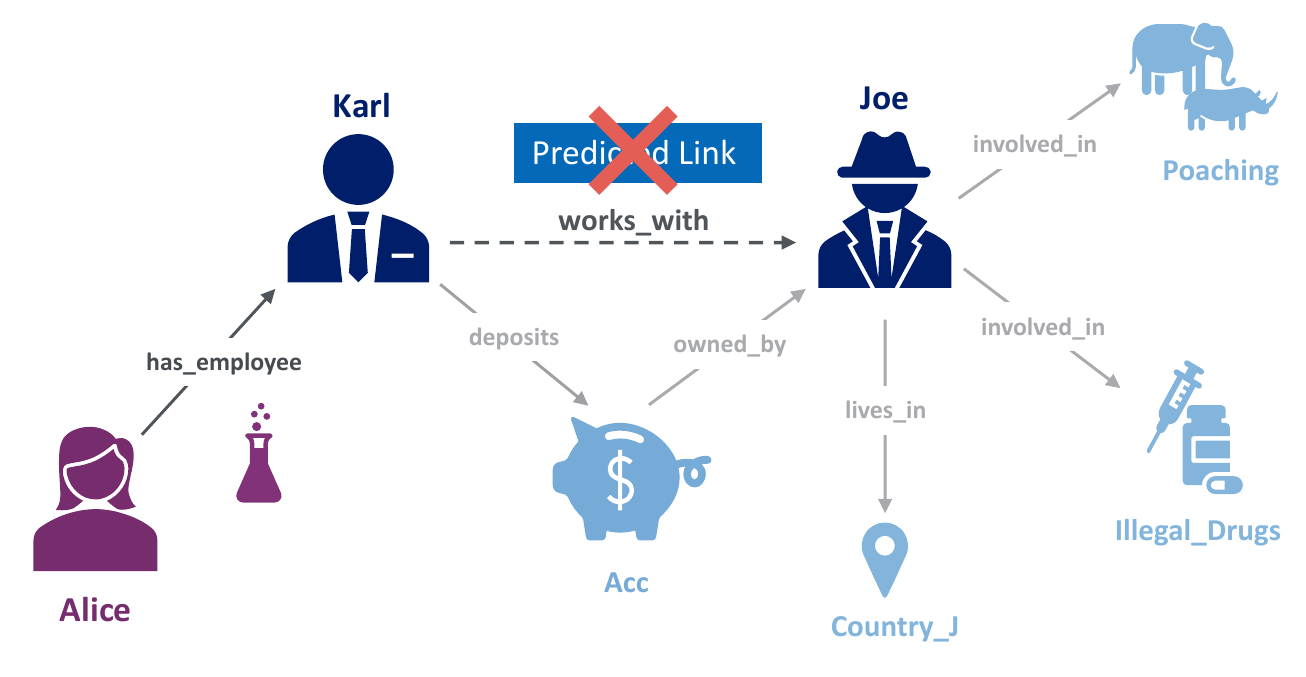}
    \caption{Inversion Attack}
    \end{subfigure}
    \newline
    \newline
    \newline
    \begin{subfigure}[htb]{0.8\textwidth}
        \includegraphics[width=1\textwidth]{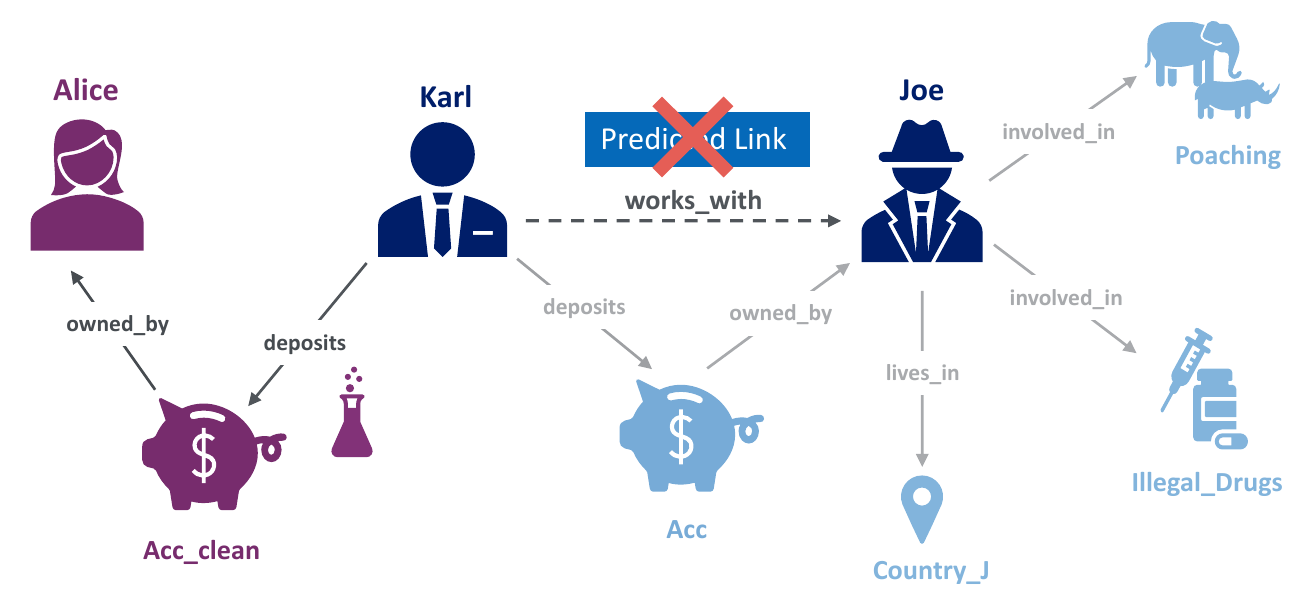}
    \caption{Composition Attack}
    \end{subfigure} 
    
    \caption{Illustrative examples for adversarial additions based on relation inference patterns. The target triple for attack is $(Karl,\mathtt {works\_with}, Joe)$, and the decoy triple is $(Karl,\mathtt {works\_with}, Alice)$. The adversarial addition based on symmetry pattern is $(Alice,\mathtt {works\_with}, Karl)$, based on inversion pattern $(Alice,\mathtt {has\_employee}, Karl)$, and based on composition pattern $(Karl,\mathtt {deposits}, Acc\_clean) \wedge (Acc\_clean, \mathtt{owned\_by}, Alice)$.}
    \label{fig:rip_examples}
\end{figure}

The predictive performance of KGE models is determined by ranking a given fact against the synthetic corruptions of its subject and object entities (Section \ref{sec:kge_eval_protocol}). Thus, the model's prediction for a target fact can be degraded by carefully selecting a decoy triple from the possible corruptions and adding adversarial triples that improve the rank assigned to the decoy triple. Consider the example of financial knowledge graph from Figure \ref{fig:knowledge_graph} in Chapter \ref{ch:background}. Adversarial additions based on the relation inference patterns for this knowledge graph are shown in Figure \ref{fig:rip_examples}. The target triple whose rank the attacker wants to degrade is $(Karl,\mathtt {works\_with}, Joe)$. To degrade the rank for object-side corruptions, the attacker aims to improve the rank of decoy triple $(Karl,\mathtt {works\_with}, Alice)$. The decoy triple's rank can be improved through the symmetry pattern by adding $(Alice,\mathtt {works\_with}, Karl)$. Similarly, the inversion pattern can be exploited to improve the rank by adding $(Alice,\mathtt {has\_employee}, Karl)$. Furthermore, the rank of $(Karl,\mathtt {works\_with}, Alice)$ can also be improved by exploiting the composition pattern to add the triples $(Karl,\mathtt {deposits}, Acc\_clean)$ and $(Acc\_clean, \mathtt{owned\_by}, Alice)$. 

Given this strategy, how should the adversarial attacker select the decoy triple and the corresponding adversarial triples to exploit a specific relation inference pattern?
For the logic expressions in the previous section, the decoy triple becomes the head atom and adversarial edits are the triples in the body of the expression.
Since the decoy triple is an object or subject side negative of the target triple, the attacker already knows the relation in the head atom. They now want to determine  (i) the adversarial relations in the body of the expression; (ii) the decoy entities which will most likely violate the inference pattern for the chosen relations and; (iii) the remaining entities in the body of the expression which will improve the prediction on the chosen decoy triple. Notice that the attacker needs all three steps for composition pattern only; for inversion pattern, only the first two steps are needed; and for symmetry pattern, only the second step is needed. 
Each of these steps is described in detail below.

\subsection{Step1: Determine Adversarial Relations}
Expressing the relation patterns as logic expressions is based on relational logic and assumes that the relations are constants. Thus, an algebraic approach is used to determine the relations in the head and body of a clause.
Given the target relation $\mathtt{r}$, the adversarial relations are determined using an algebraic model of inference \citep{yang2015distmult}. The model is intuitively similar to the prior literature on knowledge graph traversal in latent vector space, used to extract rules from knowledge graphs in \citet{yang2015distmult}, and to answer path queries in knowledge graphs in \citet{guu2015traversingknowledgegraphsvectorspace}, \citet{sun2020FaithfulEmbeddingsKnowledgeBaseQueries} and \citet{arakelyan2021complexqueryanswering}.

\paragraph{Inversion:}
If an atom $(x, \mathtt{r}, y)$ holds true, then for the learned embeddings in multiplicative models, it can be assumed that $\vec{e}_x \circ \vec{e}_{\mathtt{r}} \approx \vec{e}_y$; where $\circ$ denotes the Hadamard (element-wise) product. If the atom $(y, \mathtt{r_i}, x)$ holds true as well, then it can also be assumed that $\vec{e}_y \circ \vec{e}_{\mathtt{r_i}} \approx \vec{e}_x$. Thus, \(\vec{e}_{\mathtt{r}} \circ \vec{e}_{\mathtt{r_i}} \approx \vec{1}\) for inverse relations $\mathtt{r}$ and $\mathtt{r_i}$ when the embeddings are learned from multiplicative models. A similar expression is obtained  \(\vec{e}_{\mathtt{r}} + \vec{e}_{\mathtt{r_i}} \approx 0\) when the embeddings are learned from additive models.

Thus, to determine adversarial relations for \emph{inversion} pattern, the pre-trained embeddings are used to select $\mathtt{r_i}$ that minimizes $\abs{\vec{e}_{\mathtt{r_i}} \vec{e}_{\mathtt{r}}^T - 1}$ for the multiplicative models; and $\mathtt{r_i}$ that minimizes $\abs{\vec{e}_{\mathtt{r_i}} + \vec{e}_{\mathtt{r}}}$ for the additive models.

\paragraph{Composition:}
If two atoms $(x, \mathtt{r_1}, y)$ and $(y, \mathtt{r_2}, z)$ hold true, then for multiplicative models, \(\vec{e}_x \circ \vec{e}_{\mathtt{r_1}} \approx \vec{e}_y\) and \(\vec{e}_y \circ \vec{e}_{\mathtt{r_2}} \approx \vec{e}_z \). Therefore, \(\vec{e}_x \circ (\vec{e}_{\mathtt{r_1}} \circ \vec{e}_{\mathtt{r_2}}) \approx \vec{e}_z\). Hence, relation $\mathtt{r}$ is a composition of $\mathtt{r_1}$ and $\mathtt{r_2}$ if $\vec{e}_{\mathtt{r_1}} \circ \vec{e}_{\mathtt{r_2}} \approx \vec{e}_{\mathtt{r}}$. Similarly, for embeddings from the additive models, the composition pattern is modeled as $\vec{e}_{\mathtt{r_1}} + \vec{e}_{\mathtt{r_2}} \approx \vec{e}_{\mathtt{r}}$.

Thus, to determine adversarial relations for \emph{composition} pattern, the pre-trained embeddings are used to obtain all possible compositions of ($\mathtt{r_1}, \mathtt{r_2}$). For the multiplicative models, these compositions are obtained using $\vec{e}_{\mathtt{r_1}} \circ \vec{e}_{\mathtt{r_2}}$ and for additive models, using $\vec{e}_{\mathtt{r_1}} + \vec{e}_{\mathtt{r_2}}$. From these compositions, that relation pair is chosen for which the Euclidean distance between the composed relation embeddings and the target relation embedding $\vec{e}_{\mathtt{r}}$ is minimum.


\subsection{Step2: Determine Decoy Entities}
To select the decoy entity, three different heuristic approaches are considered - soft truth score which is inspired by the research in \citet{guo2016kale} and \citet{guo2018ruge}, ranks predicted by the KGE model and cosine distance. The choice of these three approaches is based on the author's intuition for the most promising ways to select the decoy entities that will violate the relation inference patterns. While other advanced methods might be possible, these simple methods can be implemented readily and efficiently, and serve as good starting points.

\paragraph{Soft Logical Modelling of Inference Patterns:}
Once the adversarial relations are determined, it is possible to express the grounding for symmetry, inversion and composition patterns for the decoy triples. Only the object side decoy triple is discussed below for brevity - 
\begin{align*}
    \mathcal{G}_s: (o', \mathtt{r}, s) &\Rightarrow (s, \mathtt{r}, o') \\
    \mathcal{G}_i: (o', \mathtt{r_i},s) &\Rightarrow (s, \mathtt{r}, o') \\
    \mathcal{G}_c: (s, \mathtt{r_1}, o'') \wedge (o'', \mathtt{r_2}, o') &\Rightarrow (s, \mathtt{r}, o')
\end{align*}
If the model captures the patterns $\mathcal{P}_s$, $\mathcal{P}_i$ or $\mathcal{P}_c$ to assign high rank to the target triple, then the head atom $(s, \mathtt{r}, o')$ of a grounding that violates this pattern is a suitable decoy triple. Adding the body of this grounding to the knowledge graph would improve the model performance on decoy triple through $\mathcal{P}_s$, $\mathcal{P}_i$ or $\mathcal{P}_c$. 

To determine the decoy triple this way, there is need for a measure of the degree to which a grounding satisfies an inference pattern.
This measure is called the \emph{soft truth score} $\phi : \mathcal{G} \rightarrow [0,1]$ - it provides the truth value of a logic expression indicating the degree to which the expression is true \citep{guo2016kale, guo2018ruge}. The soft truth score of grounded patterns is modeled using the t-norm based fuzzy logics \citep{hajek1998tnormfuzzylogics}.

The score $f_{sro}$ of an individual atom (i.e. triple) is computed using the KGE model's scoring function. The sigmoid function $\sigma(x)=1/(1+\exp(-x))$ is used to map this score to a continuous truth value in the range $(0,1)$. Hence, the soft truth score for an individual atom is \(\phi(s,\mathtt{r},o) = \sigma(f_{sro})\).
The soft truth score for the grounding of a pattern can then be expressed through logical composition (e.g. $\wedge$ and $\Rightarrow$) of the scores of individual atoms in the grounding. Following the research in \citet{guo2016kale, guo2018ruge}, the following compositions are defined for logical conjunction ($\wedge$), disjunction ($\vee$), and negation ($\neg$):
\begin{align*}
  \phi(a \wedge b) & = \phi(a) \cdot \phi(b), \\
  \phi(a \vee b)   & = \phi(a) + \phi(b) - \phi(a) \cdot \phi(b), \\
  \phi(\neg a)     & = 1 - \phi(a).
\end{align*}

Here, $a$ and $b$ are two logical expressions, which can either be single triples or be constructed by combining triples with logical connectives. If $a$ is a single triple $(s,\mathtt{r},o)$, we have $\phi(a)=\phi(s,\mathtt{r},o)$. Given these compositions, the truth value of any logical expression can be calculated recursively \citep{guo2016kale, guo2018ruge}. 
 
Thus, the following soft truth scores are obtained for the groundings of symmetry, inversion and composition patterns $\mathcal{G}_s$, $\mathcal{G}_i$ and $\mathcal{G}_c$ - 
\begin{align*}
    \phi(\mathcal{G}_s) &= \phi(o',\mathtt{r},s) \cdot \phi(s,\mathtt{r},o') - \phi(o',\mathtt{r},s) + 1  \\
    \phi(\mathcal{G}_i) &= \phi(o',\mathtt{r_i},s) \cdot \phi(s,\mathtt{r},o') - \phi(o',\mathtt{r_i},s) + 1. \\
    \phi(\mathcal{G}_c) &= \phi(s,\mathtt{r_1},o'') \cdot \phi(o'',\mathtt{r_2},o') \cdot \phi(s,\mathtt{r},o') - \phi(s,\mathtt{r_1},o'') \cdot \phi(o'',\mathtt{r_2},o') + 1
\end{align*}

To select the decoy triple $(s,\mathtt{r},o')$ for symmetry and inversion, all possible groundings are scored using $\phi(\mathcal{G}_s)$ and $\phi(\mathcal{G}_i)$. The head atom of the grounding with minimum score is chosen as the decoy triple.

For the composition pattern, the soft truth score $\phi(\mathcal{G}_c)$ for candidate decoy triples $(s,\mathtt{r},o')$ contains two entities $(o',o'')$ to be identified. Thus, a greedy approach is used to select the decoy entity $o'$. In this approach, the pre-trained embeddings are used to group the entities $o''$ into $k$ clusters using K-means clustering and a decoy entity with minimum soft truth score is determined for each cluster. Then, the decoy entity $o'$ with minimum score across the $k$ clusters is selected.

\paragraph{KGE Ranks:} 
For this attack heuristic, the ranking protocol from KGE evaluation is used to rank the target triple against the valid subject and object side negatives $(s',\mathtt{r},o)$ and $(s,\mathtt{r},o')$. For each side, that negative triple is selected which is ranked just worse than the target triple (that is, $negative\_rank = target\_rank + 1$). These are suitable as decoy because their predicted scores are likely not very different from the target triple's score. Thus, the model's prediction confidence for these triples might be effectively manipulated through adversarial additions. This is in contrast to very low ranked triples as decoy; where the model has likely learnt a low score with high confidence.

\paragraph{Cosine Distance:}
A high rank for the target triple $(s,\mathtt{r},o)$ against queries $(s,\mathtt{r},?)$ and $(?,\mathtt{r},o)$ indicates that $\vec{e}_s, \vec{e}_o$ are similar to the embeddings of other subjects and objects related by $\mathtt{r}$ in the training data. Thus, a suitable heuristic for selecting decoy entities $s'$ and $o'$ is to choose ones whose embeddings are dissimilar to $\vec{e}_s, \vec{e}_o$. Since these entities are not likely to occur in the neighbourhood of $o$ and $s$, they will act adversarially to reduce the rank of target triple. 
Thus, for this attack heuristic, decoy entities $s'$ and $o'$ that have maximum cosine distance from target entities $s$ and $o$ respectively are selected.


\subsection{Step3: Determine Adversarial Entities}
This step is only needed for the composition pattern because the body for this pattern has two adversarial triples.
Given the decoy triple in the head of the composition expression, the body of the expression that would maximize the rank of the decoy triple needs to be selected.
For this step, the soft-logical model defined in Step 2 for selecting the decoy triples is used again. 
The soft truth score for composition grounding of the decoy triple is given by $\phi(\mathcal{G}_t) = \phi(s,\mathtt{r_1},o'') \cdot \phi(o'',\mathtt{r_2},o') \cdot \phi(s,\mathtt{r},o') - \phi(s,\mathtt{r_1},o'') \cdot \phi(o'',\mathtt{r_2},o') + 1$. The entity $o''$ with maximum score is selected as the adversarial entity. This is because this entity satisfies the composition pattern for the decoy triple and is thus likely to improve the decoy triple's ranks on addition to the knowledge graph.

\begin{table}
    \centering
    \small
    \begin{tabular}{c c c c }
    \toprule
          \textbf{Adversarial Attack Step}    &      \textbf{Sym}   &   \textbf{Inv}   &   \textbf{Com}\\
    \midrule
        {Determine Adversarial Relations}  &   n/a       &   Alg   &   Alg  \\
    \midrule
         \multirow{3}{*}{\shortstack[l]{Determine Decoy Entities}}  &   Sft    &    Sft   &   Sft   \\
                                                &     Rnk   &    Rnk   &   Rnk  \\
                                                &     Cos   &   Cos   &   Cos  \\
    \midrule
        Determine Adversarial Entities                    &   n/a     &   n/a      &      Sft  \\
    \bottomrule
    \end{tabular}
    \caption{A summary of the heuristic approaches used for different steps of the adversarial attack with symmetry (Sym), inversion (Inv) and composition (Com) pattern. Alg denotes the algebraic model for inference patterns; Sft denotes the soft truth score; Rnk denotes the KGE ranks; and Cos denotes the cosine distance.}
    \label{tab:rip_attack_summary}
\end{table}


\section{Computational Complexity Analysis}
\label{sec:rip_complexity_analysis}
The main challenge in designing adversarial additions against KGE models is the large combinatorial search space of the candidate triples. To overcome this challenge, the attacks based on relation inference patterns break down the search space into three smaller steps. The computational complexity of the proposed attacks is discussed below.
Lets say \( \mathcal{E} \) is the set of entities and \( \mathcal{R} \) is the set of relations. 
The number of target triples to attack is $t$ and the specific target triple is $(s,\mathtt{r},o)$.

\paragraph{Determine Adversarial Relations:} This step determines the inverse relation or the composition relation of a target triple. To select the inverse relation, $\mathcal{R}$ computations are needed for every target triple. On the other hand, selecting composition relation requires the composition operation $\mathcal{R}^2$ times per target triple. To avoid the repetition of these computations for every target triple, the inverse and composition relations are pre-computed for all relations in the knowledge graph. This gives the complexity $\mathcal{O}(\mathcal{R}^2)$ for the inverse relation. For the composition relation, compositions of all relation pairs are computed and the adversarial pair is selected by comparison with the target relation. This gives $\mathcal{O}(\mathcal{R}^2 + \mathcal{R})$ complexity for composition. The computational complexity for this step is independent of the number of target triples $t$.

\paragraph{Determine Decoy Entity:} For a given target triple, the three heuristics to select the decoy entity are soft-truth score, KGE ranks and cosine distance. The expression for computing the soft-truth scores requires the scores for individual atoms in the expression for all possible entity substitutions. Using a pre-trained KGE model, one atomic score value for all possible entity substitutions can be computed by one forward call to the model. This way, for symmetry and inversion, computing the soft truth score values requires 2 forward calls to the KGE model because these expressions contain 2 different atomic expressions. 
For the composition pattern, the expression for soft truth scores contains three different atomic expressions. Furthermore, since the expression contains an entity $o''$ in addition to the decoy entity, the score needs to be computed for $k$ different clusters into which the entities are grouped. Thus, computing the soft truth values of the composition pattern for one target triple requires $3k$ forward calls to the model. %

Selection of the decoy entity based on KGE ranks requires the computation of one atomic expression for all possible entity substitutions. This requires one forward call to the pre-trained KGE model. The atomic scores further need to be ranked using Pytorch's $\mathtt{torch.sort}$ function. In general, the sorting operation has a time complexity of $\mathcal{O}(\mathcal{E} log(\mathcal{E}))$. Thus, the expected computational complexity for selecting the decoy entity for one target triple using the ranks from the KGE model is $\mathcal{O}(\mathcal{E} log(\mathcal{E}))$.

For cosine distance, the similarity of $s$ and $o$ to all entities is computed via two calls to Pytorch's $\mathtt{F.cosine\_similarity}$. Once the heuristic scores are computed, there is an additional complexity of $\mathcal{O}(\mathcal{E})$ to select the entity with minimum score. These computations need to be repeated for all target triples $t$. Thus, the complexity for decoy selection is $\mathcal{O}(t\mathcal{E})$ for all heuristics except two cases - soft truth score on composition where it is $\mathcal{O}(kt\mathcal{E})$ and scores based on KGE ranks where it is $\mathcal{O}(t (\mathcal{E} log(\mathcal{E}) + \mathcal{E}))$, which can be simplified to $\mathcal{O}(t \mathcal{E} log(\mathcal{E}))$. %

\paragraph{Determine Adversarial Entity:} This step requires the computation of ground truth scores for the composition pattern which contain three different atomic expressions. Thus, the step requires three forward calls to the KGE model. The computational complexity of this step for all target triples is $\mathcal{O}(t\mathcal{E})$.

Based on the above discussion, the computational complexity for different attacks is as follows - 
\begin{enumerate}
    \item For symmetry attacks based on soft-truth scores and cosine similarity, the complexity is $\mathcal{O}(t\mathcal{E})$; and for the symmetry attacks based on KGE ranks, the complexity is $\mathcal{O}(t\mathcal{E} log(\mathcal{E}))$
    
    \item For inversion attacks based on soft-truth scores and cosine similarity, the complexity is $\mathcal{O}(\mathcal{R}^2 + t\mathcal{E})$; and for the inversion attacks based on KGE ranks, the complexity is $\mathcal{O}(\mathcal{R}^2 + t\mathcal{E} log(\mathcal{E}))$
    
    \item For composition attacks, the complexity is $\mathcal{O}(\mathcal{R}^2 + \mathcal{R} + kt\mathcal{E})$ for soft truth score. On simplification, this complexity expression becomes $\mathcal{O}(\mathcal{R}^2 + kt\mathcal{E})$. The complexity for KGE ranks is $\mathcal{O}(\mathcal{R}^2 + \mathcal{R} + t\mathcal{E} log(\mathcal{E}))$, which on simplification becomes $\mathcal{O}(\mathcal{R}^2 + t\mathcal{E} log(\mathcal{E}))$. For composition attacks based on cosine distance, the time complexity is $\mathcal{O}(\mathcal{R}^2 + \mathcal{R} + t\mathcal{E})$, which can be simplified to $\mathcal{O}(\mathcal{R}^2 + t\mathcal{E})$.
\end{enumerate}

It is noteworthy that the above expressions for computational complexity indicate that all the proposed attack strategies are more efficient than the naive method of searching through the combinatorial space of all possible adversarial additions. For $t$ target triples, the computational complexity for the enumerative search would be $\mathcal{O}(t \mathcal{R} \mathcal{E})$. This expression does not account for the complexity of computing the metric for the impact of a candidate on the target triple. Despite this, the computational complexity of the complete attack strategies discussed above is more efficient than the naive methods. By breaking down the search space into smaller steps, the proposed attack strategies reduce the multiplicative components of the search space complexity into additive components.

\section{Experimental Setup}
As discussed in Section \ref{sec:problem_evalprotocol}, the aim of the evaluation is to assess the effectiveness of proposed attacks in \emph{degrading} the predictive performance of KGE models on missing triples that are predicted true.
For this, the state-of-the-art evaluation protocol for data poisoning attacks \citep{xu2020advgraphsurvey} has been used. First, a clean model is trained on the original data; then the adversarial edits are generated and added to the training knowledge graph; and finally a new model is re-trained on this poisoned graph. All the hyperparameters for training the KGE model on original and poisoned datasets remain the same. 

\subsection{Datasets}
The proposed attack strategies are evaluated on four models with varying inductive abilities - DistMult, ComplEx, ConvE and TransE; on two publicly available benchmark datasets for link prediction\footnote{https://github.com/TimDettmers/ConvE}- WN18RR and FB15k-237. 
To evaluate the predictive performance of the KGE models, standard KGE evaluation protocol is used (Section \ref{sec:kge_eval_protocol}). In addition, following \citet{bordes2013transe} and as is the accepted state-of-the-art practice, triples from the validation and test set that contain unseen entities are filtered out for the KGE model evaluation.

To assess the attack effectiveness in \emph{degrading} performance on triples predicted as True, a set of triples that are predicted as True by the model are needed.
Thus, a subset of the original test set is selected as \emph{target triples}, where each triple is ranked $\leq$ 10 by the original model. That is, the subset of test set that is ranked the best by the original model is selected as target triples for the evaluation.
Table \ref{tab:rip_data} provides an overview of dataset statistics and the number of target triples selected.

\begin{table}[]
\centering
\small
\setlength{\tabcolsep}{5pt}
\begin{tabular}{c  l ll}
    \toprule           
    \multicolumn{2}{l}{} & \textbf{WN18RR} &  \textbf{FB15k-237} \\ 
    \midrule
    \multicolumn{2}{l}{Entities}                  &  40,559   & 14,505 \\ 
    \multicolumn{2}{l}{Relations}                 &  11       & 237 \\ 
    \multicolumn{2}{l}{Training}                  & 86,835    &  272,115            \\ 
    \multicolumn{2}{l}{Validation}                &  2,824    &  17,526   \\ 
    \multicolumn{2}{l}{Test}                      &   2,924   &  20,438    \\ 
    \midrule
    \multirow{4}{*}{Target} 
    & DistMult  &   1,315   &  3,342    \\
    & ComplEx   & 1,369  &  3,930    \\
    & ConvE     &    1,247   &  4,711    \\
    & TransE    &    1,195   &   5,359   \\
    \bottomrule
\end{tabular}
\caption{\small Statistics for the datasets WN18RR and FB15k-237.
Triples from the validation and test set that contained unseen entities were removed to ensure that new entities are not added as adversarial edits to the training graph. 
The numbers above (including the number of entities) reflect this filtering.}
\label{tab:rip_data}
\end{table}

\subsection{Baselines}
The proposed attack strategies are compared against the following baselines based on random edits as well as state-of-art data poisoning attacks. 

\emph{Random\_n}: Random edits in the neighbourhood of each entity of the target triple.

\emph{Random\_g1}: Global random edits in the knowledge graph which are not restricted to the neighbourhood of entities in the target triple and have 1 edit per decoy triple (like symmetry and inversion).

\emph{Random\_g2}: Global random edits in the knowledge graph which are not restricted to the neighbourhood of entities in the target triple and have 2 edits per decoy triple (like composition).

\emph{Direct-Add}: Poisoning attack from \cite{zhang2019kgeattack} for edits in the neighbourhood of subject of the target triple. The method is extended to both subject and object side neighbours to match the evaluation protocol in this chapter. Further implementation details of this attack are available in Appendix \ref{apx:inference_pattern_ijcai_baseline}.

\emph{CRIAGE}: Poisoning attack from \citet{pezeshkpour2019criage}. The publicly available implementation of the attack and the default attack settings\footnote{https://github.com/pouyapez/criage} are used for this research. The method was proposed for edits in the neighbourhood of object of the target triple. It is extended to the neighbourhood of both subject and object entities to match the evaluation protocol in this research and to ensure fair evaluation.

\subsection{Implementation}
For every attack, the adversarial edit candidates that already exist in the graph are filtered out. This post-processing step is done to ensure that the poisoned training graph does not contain duplicate triples.
Similarly, duplicate adversarial edits for different targets are also removed before adding them to the original knowledge graph. 
For Step 2 of the composition attack with ground truth, the elbow method is used to determine the number of clusters for each model-data combination. The elbow method is a heuristic approach to select the optimal number of clusters in the k-means clustering, that is, the value of k. %

Further details on KGE model training, computing resources and number of clusters used are available in Appendix \ref{apx:inference_pattern_implementation}. The source code to reproduce the experimental results reported in this Chapter is available on GitHub at \url{https://github.com/PeruBhardwaj/InferenceAttack}.


\section{Evaluation of Attack Effectiveness}
\label{sec:rip_results}

\begin{table}
\centering
\fontsize{10}{10}\selectfont
\setlength{\tabcolsep}{3.45pt}
\begin{tabular}{  l  ll  ll   ll  lr }

\toprule
    
      & \multicolumn{2}{c}{\textbf{DistMult}} & \multicolumn{2}{c}{\textbf{ComplEx}} & \multicolumn{2}{c}{\textbf{ConvE}} & \multicolumn{2}{c}{\textbf{TransE}} \\
   \cmidrule(lr){2-3}  \cmidrule(lr){4-5}  \cmidrule(lr){6-7} \cmidrule(lr){8-9} 
     & \multicolumn{1}{l}{\textbf{MRR}}    & \multicolumn{1}{l}{\textbf{H@1}}  & \multicolumn{1}{l}{\textbf{MRR}}    & \multicolumn{1}{l}{\textbf{H@1}} & \multicolumn{1}{l}{\textbf{MRR}}    & \multicolumn{1}{l}{\textbf{H@1}} & \multicolumn{1}{l}{\textbf{MRR}}    & \multicolumn{1}{l}{\textbf{H@1}} \\
\midrule
    
     \textbf{Original}      & 0.90            &  0.85     &  0.89             &  0.84      &  0.92           &   0.89    & 0.36           &  0.03 \\
\midrule
     \multicolumn{9}{l}{\textbf{Baseline Attacks}} \\
\midrule
     Random\_n     & 0.86 (-4\%)     &  0.83     &  0.84 (-6\%)      &    0.80    &  0.90 (-2\%)    &   0.88    &  0.28 (-20\%)  &   0.01 \\
     Random\_g1    & 0.88            &  0.83     &  0.88             &   0.83     &  0.92           &   0.89    &  0.35          &   0.02 \\
     Random\_g2    & 0.88            &  0.83     &  0.88             &   0.83     &  0.91           &    0.89   &  0.34          &   0.02 \\
\midrule
     Direct-Add     & 0.82 (-8\%)    &  0.81     &  0.76 (-14\%)    &    0.74    & 0.90 (-2\%)     &   0.87    &  \textbf{0.24 (-33\%)}     &   \textbf{0.01} \\
     CRIAGE        & 0.87          &  0.84     &  -                &   -        & 0.90         &   0.88    &  -            &   - \\
\midrule
    \multicolumn{9}{l}{\textbf{Proposed Attacks}} \\
\midrule
     Sym\_truth    & 0.66           &  0.40     &  \textbf{0.56 (-37\%)}  &  \textbf{0.24}      & \textbf{0.61 (-34\%)}   &    \textbf{0.28}    & 0.57      &  0.36 \\
     Sym\_rank     & 0.61           &  0.32     &  \textbf{0.56 (-37\%)}  &  \textbf{0.24}      & 0.62                    &    0.31    & 0.25               &  0.02 \\
     Sym\_cos      & \textbf{0.57 (-36\%)} &  \textbf{0.32}  &  0.62     &   0.43     &  0.67        &    0.44                        & \textbf{0.24 (-33\%)}  &  \textbf{0.01} \\
\midrule
     Inv\_truth   & 0.87           &   0.83     &  0.86                  &   0.80     & 0.90          &    0.87    &  0.34          &  0.03 \\
     Inv\_rank    & 0.86           &   0.83     &  0.85                  &   0.80     & 0.89 (-4\%)   &    0.85    &  0.25          &  0.02 \\
     Inv\_cos     & 0.83 (-8\%)     &  0.82     &  0.80 (-10\%)          &   0.79     & 0.90          &    0.88    &  0.25 (-30\%)  &  0.01 \\
\midrule
    
     Com\_truth   & 0.86           &  0.83      &  0.86                &   0.81     &  0.89           &    0.86   &   0.53 (+49\%)    &  0.27 \\
     Com\_rank    & 0.85 (-5\%)      &  0.80      &  0.83                &   0.77     &  0.89           &    0.84   &   0.57            &  0.32 \\
     Com\_cos     & 0.86           &  0.77      &  0.82 (-8\%)         &   0.70     &  0.88(-4\%)     &    0.83   &   0.53 (+49\%)     &   0.27 \\
    
\bottomrule    

\end{tabular}
\caption{\small Reduction in MRR and Hits@1 due to different attacks on the \textbf{target split} of WN18RR. 
For each block of rows, the \emph{best} relative percentage difference from original MRR; computed as $(original-poisoned)/original*100$ is reported. Lower values indicate better results; best results for each model are in bold. Statistics on the target split are in Table \ref{tab:rip_data}.}
\label{tab:rip_benchmark_mrr_WN18RR}
\end{table}

\begin{table}[]
\centering
\fontsize{10}{10}\selectfont
\setlength{\tabcolsep}{3.5pt}
\begin{tabular}{  l  ll  ll   ll  lr }
\toprule
     & \multicolumn{2}{c}{\textbf{DistMult}} & \multicolumn{2}{c}{\textbf{ComplEx}} & \multicolumn{2}{c}{\textbf{ConvE}} & \multicolumn{2}{c}{\textbf{TransE}} \\
   \cmidrule(lr){2-3}  \cmidrule(lr){4-5}  \cmidrule(lr){6-7} \cmidrule(lr){8-9} 
     & \textbf{MRR}   & \textbf{H@1}  & \textbf{MRR}   & \textbf{H@1} & \textbf{MRR}   & \textbf{H@1} & \textbf{MRR}   & \textbf{H@1} \\
\midrule
     \textbf{Original}   & 0.61              &  0.38     & 0.61       &   0.45             & 0.61              &   0.45      &    0.63               &  0.48   \\
\midrule
     \multicolumn{9}{l}{\textbf{Baseline Attacks}} \\
\midrule
     Random\_n  & 0.54 (-11\%)    &  0.40     & 0.54  (-12\%)  &    0.40          & 0.56 (-8\%)       &   0.41       &   0.60 (-4\%)       & 0.45   \\
     Random\_g1 & 0.54            &  0.40     & 0.55           &   0.41           & 0.57              &   0.43       &    0.62             &  0.46   \\
     Random\_g2  & 0.55           &  0.41     & 0.55           &   0.40           & 0.57              &   0.42        &  0.61             &  0.46   \\
\midrule
     Direct-Add   & 0.53 (-13\%)     &  0.39     & 0.51 (-16\%)    &  0.38            & 0.54  (-11\%)            &  0.39       &  0.57 (-10\%)       &  0.42  \\
     CRIAGE     & 0.54     &  0.41     &  -              &  -               & 0.56        &  0.41       & -                     &  -   \\
\midrule
    \multicolumn{9}{l}{\textbf{Proposed Attacks}} \\
\midrule
     Sym\_truth &  0.51            & 0.36       &  0.56         &   0.41             & \textbf{0.51} (\textbf{-17\%})       &   \textbf{0.34}      & 0.62          &  0.48  \\
     Sym\_rank  &  0.53            &  0.39      &  0.53         & \textbf{0.38}      &  0.55          &  0.38        & \textbf{0.53} (\textbf{-16\%})       &   \textbf{0.36} \\
     Sym\_cos   &  \textbf{0.46} (\textbf{-25\%}) &  \textbf{0.31} & \textbf{0.51} (\textbf{-17\%})  &  \textbf{0.38}     
               & 0.52            &   0.37        & 0.55         &  0.40  \\
\midrule
     Inv\_truth  & 0.55            & 0.41       &   0.54          &   0.40        &  0.56              &  0.41        & 0.62               & 0.46   \\
     Inv\_rank   & 0.56            &  0.43      &   0.55          &   0.40        &  0.55  (-9\%)      &   0.40       & 0.58 (-8\%)       &  0.42  \\
     Inv\_cos    & 0.54 (-11\%)      &  0.40     &  0.53 (-14\%)  &   0.39        &  0.56              &   0.42        & 0.59              &  0.44   \\
\midrule
     Com\_truth  & 0.56          &  0.42      &  0.55             &   0.41       &   0.57            &    0.43       & 0.65                &  0.51  \\
     Com\_rank   & 0.56 (-8\%)      & 0.42      & 0.55 (-11\%)    &   0.40       &   0.56 (-8\%)     &   0.41        & 0.69                &  0.48   \\
     Com\_cos    & 0.56 (-8\%)      &  0.43     & 0.56            &   0.42      &    0.56            &   0.42        & 0.63 (0\%)       &  0.49   \\
\bottomrule    

\end{tabular}
\caption{\small Reduction in MRR and Hits@1 due to different attacks on the \textbf{target split} of FB15k-237. 
For each block of rows, the \emph{best} relative percentage difference from original MRR; computed as $(original-poisoned)/original*100$ is reported. Lower values indicate better results; best results for each model are in bold. Statistics on the target split are in Table \ref{tab:rip_data}.}
\label{tab:rip_benchmark_mrr_FB15k-237}
\end{table}

Table \ref{tab:rip_benchmark_mrr_WN18RR} and \ref{tab:rip_benchmark_mrr_FB15k-237} show the reduction in MRR and Hits@1 due to different attacks on the WN18RR and FB15k-237 datasets. 
These results show that the proposed adversarial attacks outperform the random baselines and the state-of-art data poisoning attacks against all KGE models on both datasets. 

\subsection{Comparison across Relation Inference Patterns}
It is observed that the attacks based on symmetry inference pattern perform the best across all model-dataset combinations. This indicates the sensitivity of KGE models to symmetry pattern. For DistMult, ComplEx and ConvE, this sensitivity can be explained by the symmetric nature of the scoring functions of these models. That is, the models assign either equal or similar scores to triples that are symmetric opposite of each other. In the case of TransE, the model's sensitivity to symmetry pattern is explained by the translation operation in scoring function. The score of target $(s,\mathtt{r},o)$ is a translation from subject to object embedding through the relation embedding. Symmetry attack adds the adversarial triple $(o',\mathtt{r},s)$ where the relation is same as the target relation, but target subject is the object of adversarial triple. Now, the model learns the embedding of $s$ as a translation from $o'$ through relation $\mathtt{r}$. This adversarially modifies the embedding of $s$ and in turn, the score of $(s,\mathtt{r},o)$. 

Furthermore, the inversion and composition based attacks perform better than the baselines in most cases, but not as good as symmetry. For the composition pattern, it is likely that the model has stronger bias for shorter and simpler patterns like symmetry and inversion than for composition. This makes it harder to deceive the model through composition than through symmetry or inverse. 

\subsection{Comparison of Datasets}
The above observation for the inversion and composition pattern is particularly true for FB15k-237 where the performance for these patterns is similar to random baselines. 
One possible reason for the low effectiveness on FB15k-237 dataset could be the high connectivity of this dataset \citep{dettmers2018conve}. This means that the KGE model relies on a high number of triples to learn the target triples' ranks. Thus, effectively poisoning the KGE models for FB15k-237 will likely require more adversarial triples per target triple than that considered in this research.

The inversion pattern is likely ineffective on the benchmark datasets because these datasets do not have any inverse relations \citep{dettmers2018conve,toutanova2015observed}. This implies that the attacks cannot identify the inverse of the target triple's relation in Step 1 and a KGE model trained on these clean datasets would not be vulnerable to inversion attacks. To investigate this hypothesis further, the attacks are evaluated on WN18 dataset where the inverse relations have not been filtered out. This means that the KGE model can learn the inversion pattern and the inversion attacks can identify the inverse of the target relation. 

Table \ref{tab:rip_benchmark_mrr_WN18} shows the results for different adversarial attacks on WN18. In this setting, it is observed that the inversion attacks outperform other attacks against ComplEx on WN18, indicating the sensitivity of ComplEx to the inversion pattern when the dataset contains inverse relations. It is also observed that the symmetry based attack is the most effective for DistMult, ConvE and TransE. This indicates the sensitivity of these models to the symmetry pattern even when the inverse relations are present in the knowledge graph. For DistMult and ConvE, this is likely due to the symmetric nature of their scoring functions; and for TransE, this is likely because of the translation operation as discussed earlier.

\begin{table}[]
\centering
\fontsize{10}{10}\selectfont
\setlength{\tabcolsep}{3.5pt}
\begin{tabular}{  l  ll  ll   ll  lr }
\toprule
     & \multicolumn{2}{c}{\textbf{DistMult}} & \multicolumn{2}{c}{\textbf{ComplEx}} & \multicolumn{2}{c}{\textbf{ConvE}} & \multicolumn{2}{c}{\textbf{TransE}} \\
   \cmidrule(lr){2-3}  \cmidrule(lr){4-5}  \cmidrule(lr){6-7} \cmidrule(lr){8-9}
     & \textbf{MRR}   & \textbf{H@1}  & \textbf{MRR}   & \textbf{H@1} & \textbf{MRR}   & \textbf{H@1} & \textbf{MRR}   & \textbf{H@1} \\
\midrule
     \textbf{Original}   & 0.82              &  0.67     & 0.99       &   0.99             & 0.80              &   0.63      &    0.65               &  0.45   \\
\midrule
     \multicolumn{9}{l}{\textbf{Baseline Attacks}} \\
\midrule
     Random\_n  & 0.80 (-2\%)    &  0.63     & 0.99  (0\%)  &    0.98          & 0.79 (-2\%)       &   0.61       &   0.46 (-29\%)       & 0.18   \\
     Random\_g1 & 0.82            &  0.66     & 0.99           &   0.98           & 0.80              &   0.62       &    0.57             &  0.33   \\
     Random\_g2  & 0.81           &  0.65     & 0.99           &   0.98           & 0.79              &   0.62        &  0.50             &  0.22   \\
\midrule
     Direct-Add   & 0.77 (-6\%)     &  0.59     & 0.97 (-3\%)    &  0.95            & 0.77  (-3\%)            &  0.61       &  0.43 (-33\%)       &  0.16  \\
     CRIAGE     & 0.78     &  0.61     &  -              &  -               & 0.78        &  0.63       & -                     &  -   \\
\midrule
    \multicolumn{9}{l}{\textbf{Proposed Attacks}} \\
\midrule
     Sym\_truth &  0.62            & 0.30       &  0.90         &   0.82             & \textbf{0.58} (\textbf{-17\%})       &   \textbf{0.27}      & 0.74          &  0.60  \\
     Sym\_rank  &  0.59            &  0.27      &  0.89 (-10\%)        & 0.79      &  0.62          &  0.33        & 0.52       &   0.34 \\
     Sym\_cos   &  \textbf{0.50} (\textbf{-38\%}) &  \textbf{0.17} &  0.92   &  0.85   & 0.60            &   0.35        & \textbf{0.41 (-37\%)}         &  \textbf{0.13}  \\
\midrule
     Inv\_truth  & 0.81            & 0.66       &   0.86          &   0.74        &  0.78 (-3\%)      &  0.61        & 0.59               & 0.34   \\
     Inv\_rank   & 0.82            &  0.66      &   \textbf{0.84 (-16\%)}   &   \textbf{0.68}         &  0.79      &   0.61       & 0.55       &  0.34  \\
     Inv\_cos    & 0.79 (-3\%)      &  0.64     &  0.87   &   0.75        &  0.80              &   0.63        & 0.51 (-22\%)             &  0.25   \\
\midrule
     Com\_truth  & 0.79          &  0.62      &  0.98             &   0.97       &   0.77            &    0.62       & 0.53 (-18\%)               &  0.25  \\
     Com\_rank   & 0.80          & 0.64      & 0.98               &   0.96       &   0.75 (-6\%)     &   0.58        & 0.67                &  0.47   \\
     Com\_cos    & 0.78 (-5\%)      &  0.61     & 0.97 (-2\%)      &   0.95      &    0.77            &   0.62        & 0.58        &  0.32   \\
\bottomrule    

\end{tabular}
\caption{\small Reduction in MRR and Hits@1 due to different attacks on the \textbf{target split} of WN18. For each block of rows, the \emph{best} relative percentage difference from original MRR; computed as $(original-poisoned)/original*100$ are reported. Lower values indicate better results; best results for each model are in bold.}
\label{tab:rip_benchmark_mrr_WN18}
\end{table}

\subsection{Analysis on Decoy Triples}
An exception in the results is the composition pattern on TransE where the model performance improves instead of degrading on the target triples. This is likely due to the model's sensitivity to composition pattern such that adding this pattern improves the performance on all triples, including target triples. To verify this, the change in ranks of the decoy triples was investigated.

\begin{figure}[]
    \centering
    \begin{subfigure}[htb]{0.47\textwidth}
        \includegraphics[width=1\textwidth]{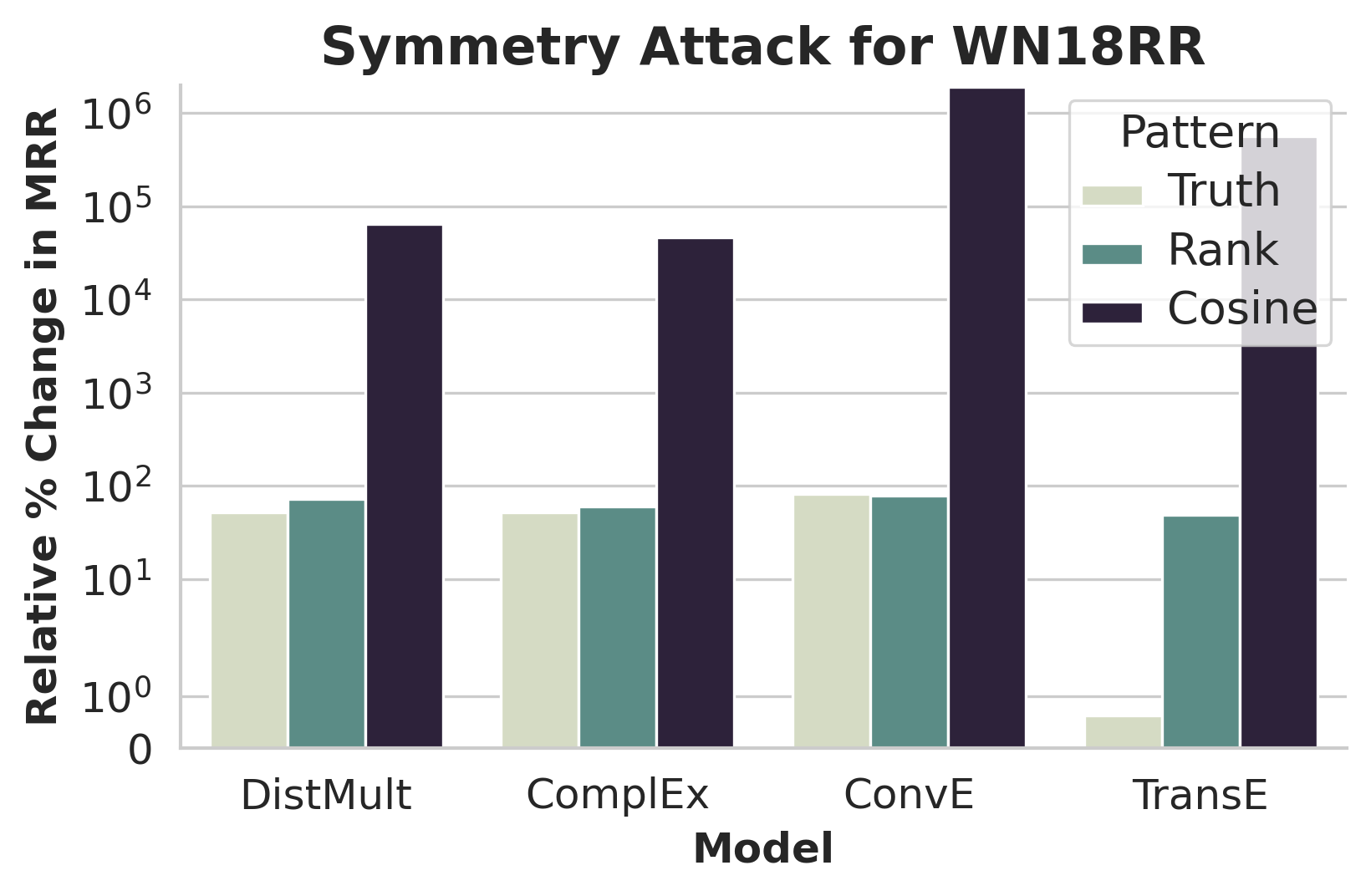}
    \end{subfigure}
    \quad
    \begin{subfigure}[htb]{0.47\textwidth}
        \includegraphics[scale=0.5,width=1\textwidth]{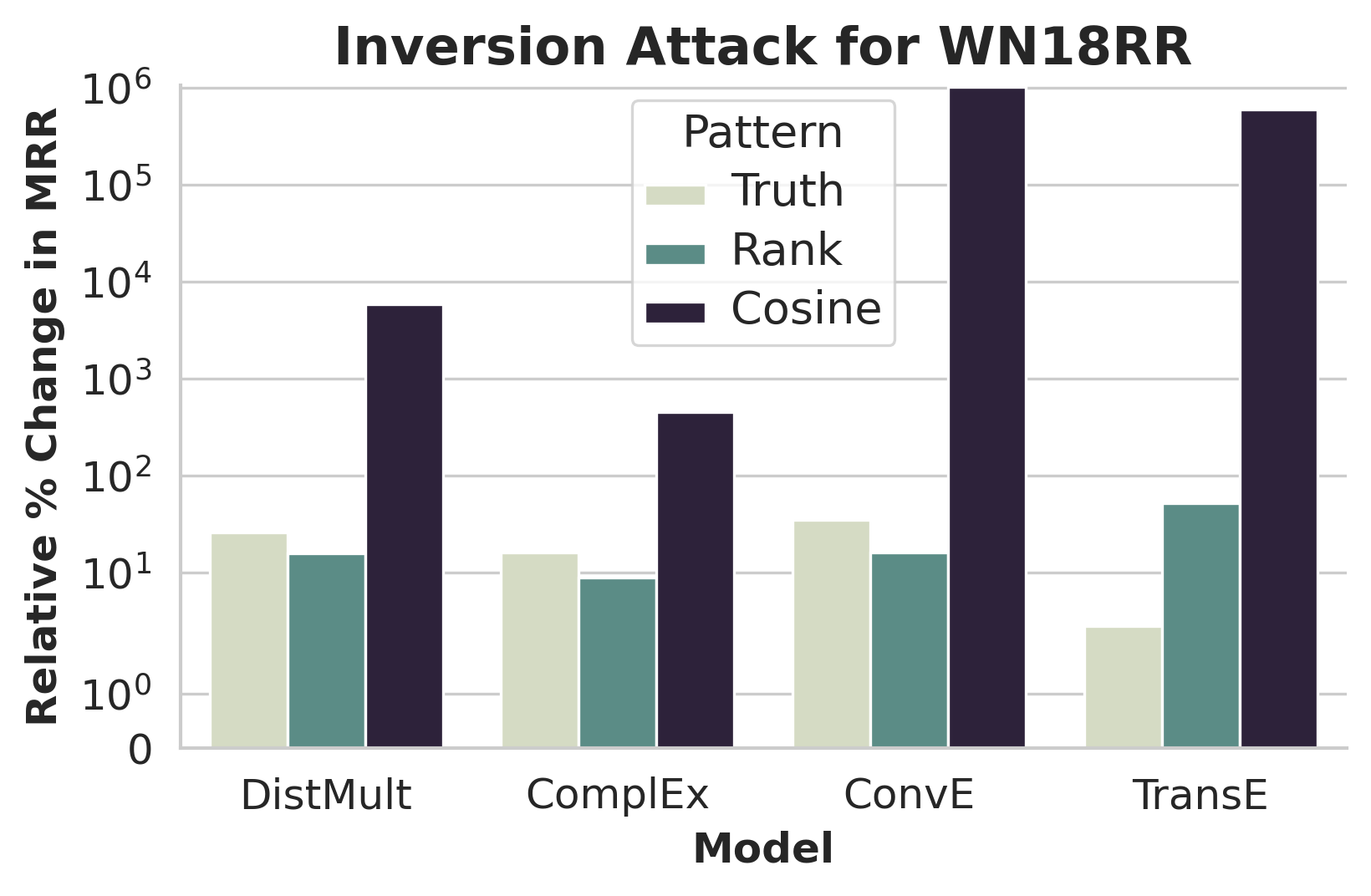}
    \end{subfigure}
    \newline
    \newline
    \newline
    \begin{subfigure}[htb]{0.47\textwidth}
        \includegraphics[width=1\textwidth]{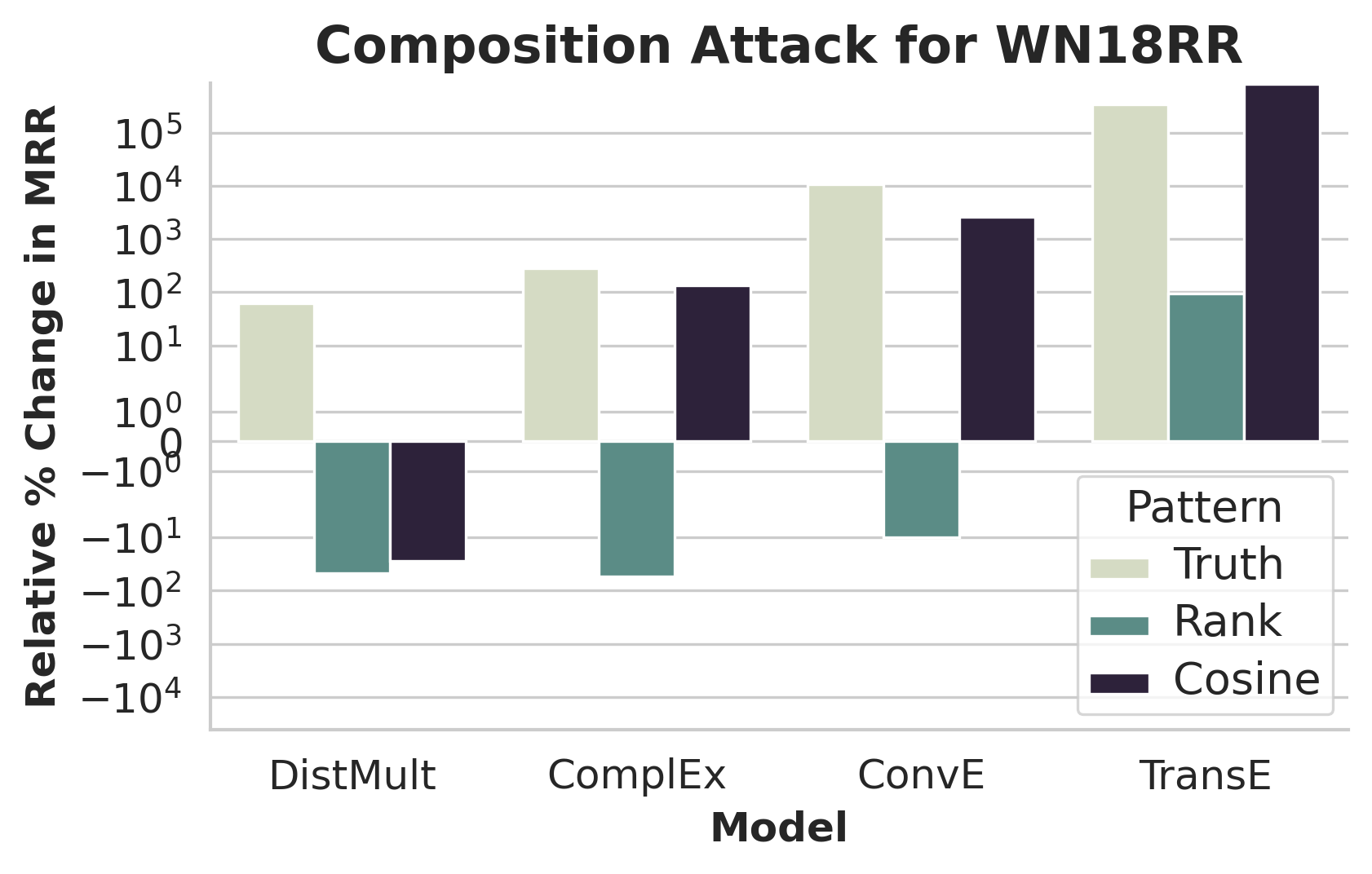}
    \end{subfigure} 
    \quad
    \begin{subfigure}[htb]{0.47\textwidth}
        \includegraphics[width=1\textwidth]{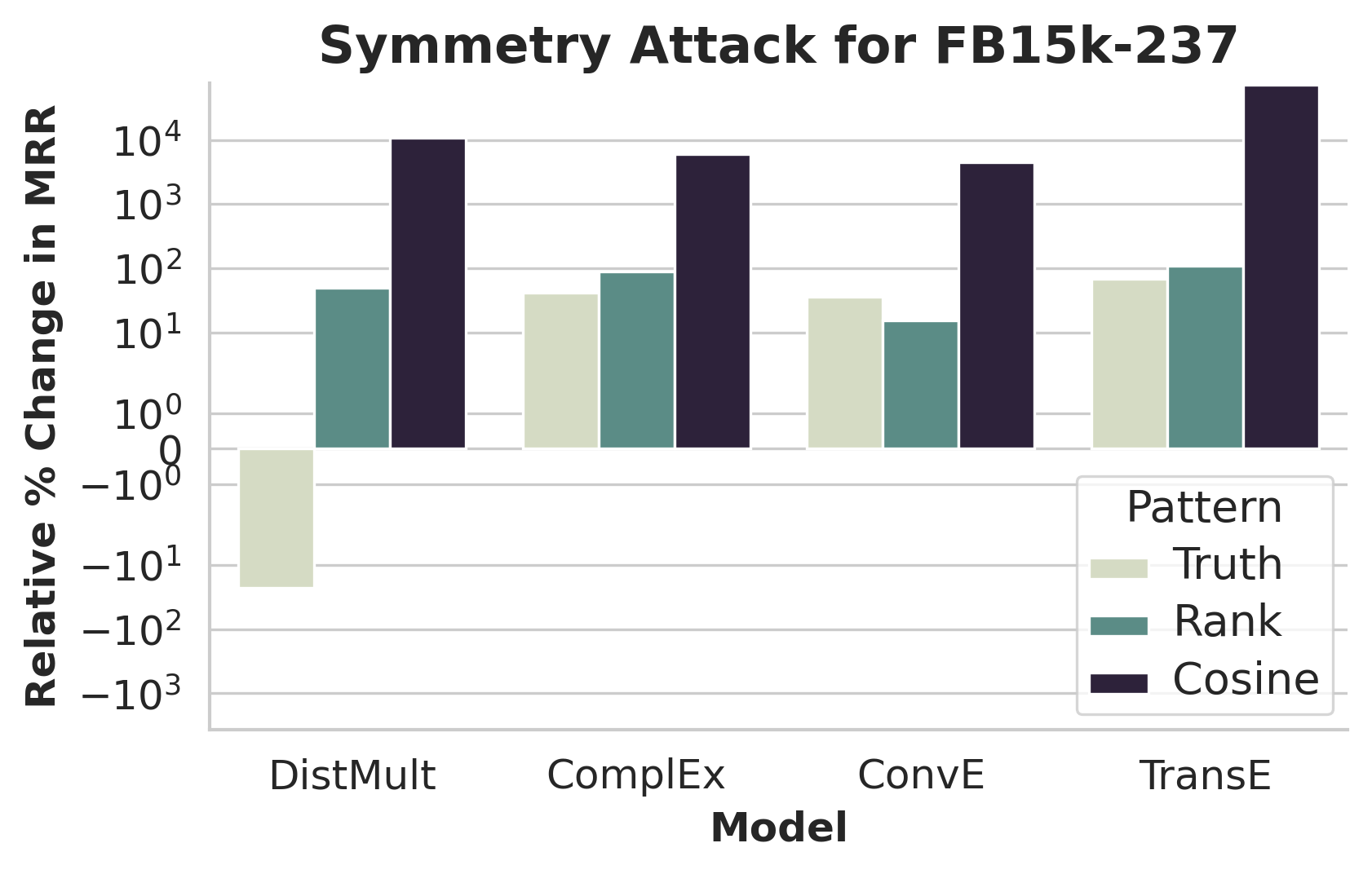}
    \end{subfigure}
    \newline
    \newline
    \newline
    \begin{subfigure}[htb]{0.47\textwidth}
        \includegraphics[width=1\textwidth]{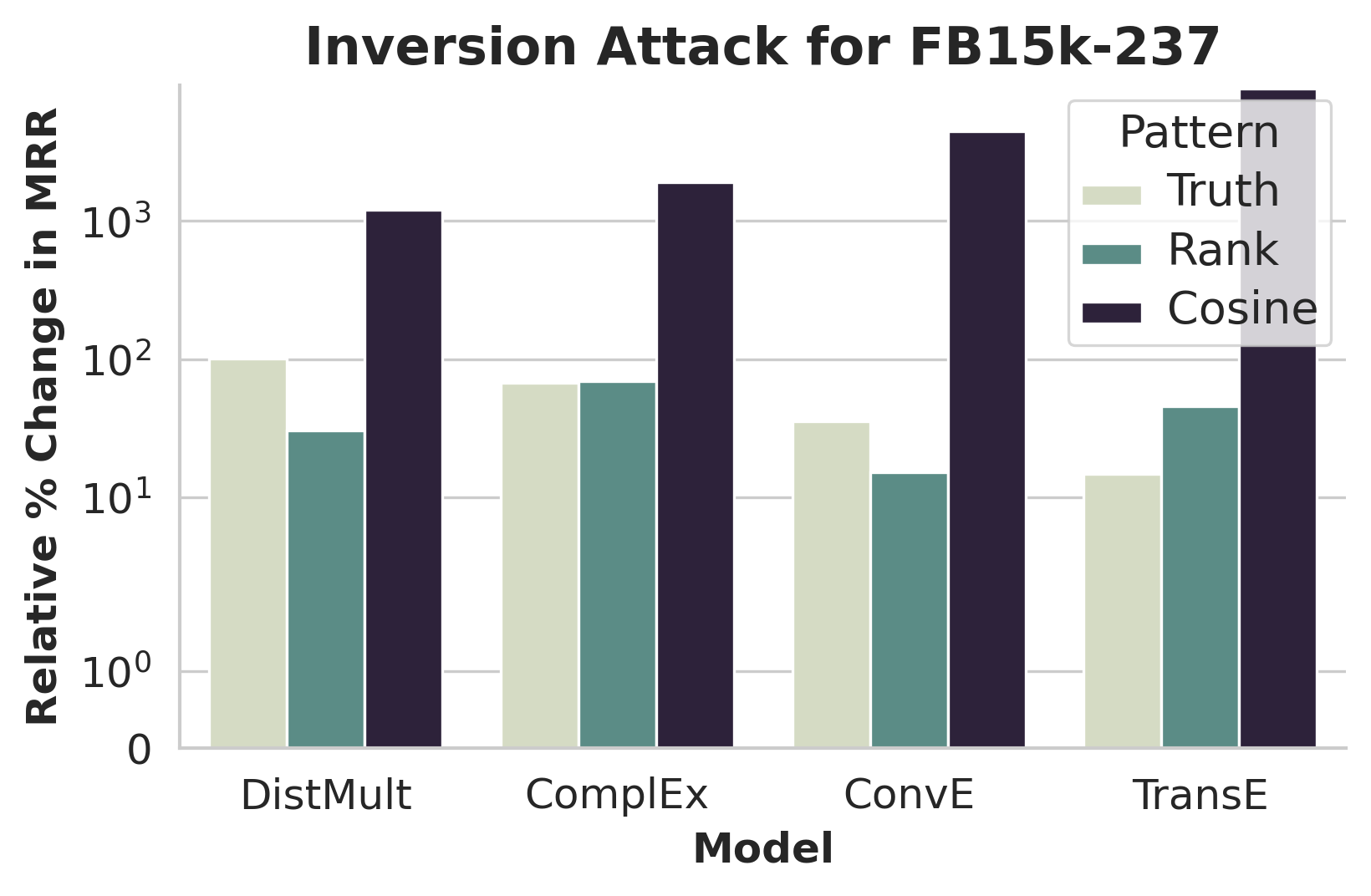}
    \end{subfigure}
    \quad
    \begin{subfigure}[htb]{0.47\textwidth}
        \includegraphics[width=1\textwidth]{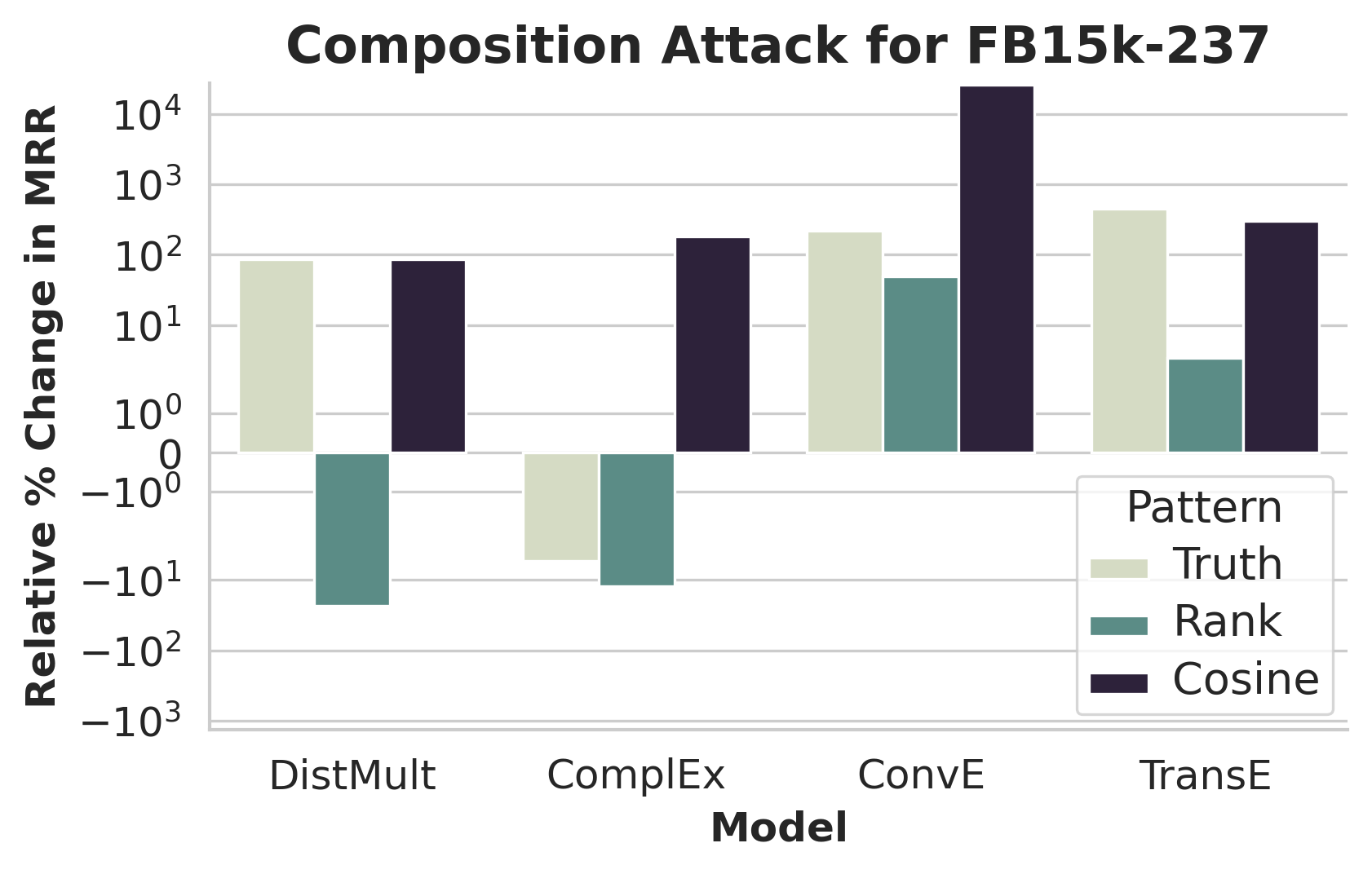}
    \end{subfigure}
    \caption{Mean of the relative increase in MRR of object and subject side \textbf{decoy triples} due to proposed attacks on WN18RR and FB15k-237. The increase is computed relative to the original MRR of decoy triples as $(poisoned - original)/original$. The scale on the y-axis is a symmetric log scale. Higher values are better; as they show the effectiveness of attack in improving decoy triples' ranks relative to their original ranks.}
    \label{fig:rip_decoy_graphs}
\end{figure}

The proposed attacks are designed to generate adversarial triples that improve the KGE model performance on decoy triples $(s,\mathtt{r},o')$ and $(s',\mathtt{r},o)$. 
Thus, an analysis was done to determine whether the performance of KGE models improves or degrades over decoy triples after poisoning.
For the decoy triples on object side $(s,\mathtt{r},o')$, the change in object side MRR was computed relative to the original object side MRR of these triples. Similarly, for the decoy triples on subject side $(s',\mathtt{r},o)$, the change in subject side MRR was computed relative to the original subject side MRR of these decoy triples. Figure \ref{fig:rip_decoy_graphs} shows the plots for the mean change in MRR of object and subject side decoy triples.

It can be noticed that the composition attacks against TransE are effective in improving the ranks of decoy triples on both WN18RR and FB15k-237. Thus, it is likely that the composition attack does not work against TransE for WN18RR because the original dataset does not contain any composition relations; thus adding this pattern improves model's performance on \emph{all} triples instead of just the target triples because of the sensitivity of TransE to composition pattern.

It also explains why the increase is more significant for WN18RR than FB15k-237. WN18RR does not have any composition relations but FB15k-237 does; thus, adding these to WN18RR shows significant improvement in performance. This behaviour for the composition pattern also indicates that the selection of adversarial entities in Step 3 of the composition attacks can be improved. 

Further investigation of these and additional hypotheses about the proposed adversarial attacks are interesting directions for future work. 

\section{Additional Evaluation}
\subsection{Analysis of Runtime Efficiency}
 \label{sec:rip_runtime_analysis}
 In this section, the runtime efficiency of the baseline and proposed attacks is compared. Given the original KGE model and the set of target triples, the absolute time taken to generate the complete set of adversarial additions for all target triples is recorded. Table \ref{tab:rip_runtime} shows the time taken (in seconds) to select the adversarial triples using different attack strategies for all models on WN18 dataset. 
 Similar patterns were observed for attack execution on other datasets. 
 
 \begin{table}[]
\centering
\fontsize{10}{10} \selectfont
\setlength{\tabcolsep}{3.5pt}
\begin{tabular}{c  l  rrrr  }
\toprule
     & & \textbf{DistMult} & \textbf{ComplEx} & \textbf{ConvE} & \textbf{TransE} \\
   
\midrule
\multirow{5}{*}{\shortstack[l]{\textbf{Baseline} \\ \textbf{Attacks}}}
    & Random\_n              &  10.08    & 10.69      & 8.76     & 7.83         \\
    & Random\_g1              &  \textbf{8.28}     &  \textbf{8.16}     & \textbf{7.64}    & \textbf{6.49}      \\
    & Random\_g2              &  16.01     &  15.82     &  18.72     & 13.33      \\
\cline{2-6} \\[-12pt]
    & Direct-Add            &  94.48     &  \textbf{255.53}   & 666.85   & \textbf{81.96}          \\
    & CRIAGE                 & \textbf{21.77}     &  -     & \textbf{21.96}    & -       \\
\midrule
\multirow{9}{*}{\shortstack[l]{\textbf{Proposed} \\ \textbf{Attacks}}}
    & Sym\_truth            & \textbf{19.63}            & 35.40     & \textbf{22.76}    & 31.59        \\
    & Sym\_rank             & 23.47            & \textbf{27.25}    & 25.82    &  25.03               \\
    & Sym\_cos               & 22.52            & 28.62       &  25.69   & \textbf{23.13}        \\
\cline{2-6} \\[-12pt] 
    & Inv\_truth           & \textbf{11.43}           & \textbf{15.69}      & 24.13    & 31.89    \\
    & Inv\_rank           & 15.27           & 18.14   & 30.99    & 21.82        \\
    & Inv\_cos           &  14.96           &  20.47      & \textbf{23.02}    & \textbf{20.63}    \\
\cline{2-6} \\[-12pt]
    
    & Com\_truth       &  2749.60           &  1574.44    & 6069.79      & 470.34        \\
    & Com\_rank       &  \textbf{22.04}           &  \textbf{31.53}     & 37.81        & 20.88                 \\
    & Com\_cos       &  34.78           &  68.06      & \textbf{32.37}    &  \textbf{19.86}                 \\
    
\bottomrule    

\end{tabular}
\caption{Runtime efficiency of the baseline and proposed adversarial adversarial against different KGE models for WN18 dataset. The absolute time taken in seconds to generate the complete set of adversarial additions for all the target triples is reported.}
\label{tab:rip_runtime}
\end{table}

For CRIAGE, the reported time does not include the time taken to train the auto-encoder model for that attack. Similarly, for soft-truth based composition attacks, the reported time does not include the time taken to pre-compute the clusters of entities.
It is observed that the proposed attacks are more efficient than the baseline Direct-Add attack which requires a combinatorial search over the candidate adversarial triples. In addition, the proposed attacks have comparable efficiency to CRIAGE. 
Among the different proposed attacks, composition attacks based on soft-truth score take more time than others because they select the decoy entity by computing the soft-truth score for multiple clusters.

\subsection{Qualitative Analysis}
Since adversarial attacks cause model failure for specific predictions, these failure points can be used to understand the predictive performance of KGE models. This section provides examples of the adversarial additions generated using the Inference Attacks. Tables \ref{tab:rip_examples_wn18rr_complex_max} and \ref{tab:rip_examples_fb15k_complex_max} show the adversarial additions for target triples which undergo the maximum change in ranks due to poisoning. While symmetry and inversion patterns have two adversarial additions, the composition pattern has four. Table \ref{tab:rip_examples_wn18rr_complex_max} provides examples for WN18RR ComplEx and Table \ref{tab:rip_examples_fb15k_complex_max} provides examples for FB15k-237 ComplEx. The entities in the benchmark data splits are represented as numeric IDs not string expressions. The string representations of entities for Wordnet (WN18RR) are obtained using the \emph{definitions.txt} file from the original split at \url{https://everest.hds.utc.fr/doku.php?id=en:smemlj12}. For Freebase (FB15k-237), the entity representations are obtained using a community build browser at \url{https://freebase.toolforge.org/}. The values for maximum change in ranks for WN18RR are -- $(13385.5 , 34866 , 40554)$ for Symmetry attacks, $(22212.5 , 40292.5 , 40549.5)$ for Inversion attacks, and $(17150.5 , 40460 , 35417)$ for Composition attacks. On the other hand, these values for FB15k-237 are -- $(1292 , 596 , 4400.5)$ for Symmetry, $(832 , 1119 , 1903)$ for Inversion, and $(675 , 614 , 2180.5)$ for Composition. 

As indicated by the results in Tables \ref{tab:rip_benchmark_mrr_WN18RR} and \ref{tab:rip_benchmark_mrr_FB15k-237}, the maximum change in ranks due to poisoning of WN18RR is significantly more than FB15k-237. This observation was explained above in terms of the graph connectivity - since FB15k-237 has more neighbours for the target triple, the KGE model prediction relies on a larger number of triples and is thus, more difficult to perturb than WN18RR. Consider the examples for the Sym\_truth attacks. For WN18RR, the target (\emph{date\_NN\_6} ,\  \emph{hypernym} ,\  \emph{month\_NN\_1}) has $11$ neighbouring triples, whereas for FB15k-237, the triple (\emph{Oprah Winfrey} \ ,\   \emph{award/ ranking/ list}    \ ,\   \emph{Time 100}) has $76$ neighbours. Furthermore, the triple with minimum change in ranks for the Sym\_truth attack on WN18RR is (\emph{family\_bruchidae\_NN\_1}, \emph{hypernym}, \emph{arthropod\_family\_NN\_1}). Though this triple has the same target relation \emph{hypernym}, the number of neighbouring triples are $146$. These values provide further support that the KGE model's predictions for triples with fewer neighbours are more vulnerable to adversarial attacks than those for triples with more neighbours.

For both datasets, it can be observed that different attack methods cause maximum change to the ranks of different target triples, though the same set of target triples is used for all attacks. This is likely because the predictions for different target triples are sensitive to different inference patterns or attack strategies. However, the relation \emph{hypernym} is common among the target triples for WN18RR.

Furthermore, the adversarial relations for symmetry pattern are not necessarily symmetric. For example, for the symmetry attack on WN18RR, the relations \emph{hypernym} and \emph{derivationally\_related\_form} are most sensitive. However, only \emph{derivationally\_related\_form} is an actual symmetric relation because it relates similar words with different syntactic categories or parts-of-speech. The relation \emph{hypernym} denotes supertype or generalization, and is not symmetric. Similarly, for FB15k-237, \emph{/people/ person/ religion} is not a symmetric relation. These adversarial relations are, however, still effective likely due to the training graph structure, which makes the model prediction sensitive to this inference pattern. Consider the target triple (\emph{date\_NN\_6} ,\  \emph{hypernym} ,\  \emph{month\_NN\_1}) from WN18RR. The entity \emph{date\_NN\_6} is related to \emph{date\_VB\_3} for $2$ triples, and has the relation \emph{hypernym} for a third triple. Similarly, the entity \emph{month\_NN\_1} is related via \emph{hypernym} to $7$ entities, and via \emph{has\_part} to \emph{week\_NN\_3}. However, there is no direct or $2$-hop link between the target entities. Thus, connecting (\emph{date\_VB\_3} \ , \emph{date\_NN\_6}) and (\emph{month\_NN\_1} \ , \emph{week\_NN\_3}) via the \emph{hypernym} relation improves the ranks of decoy triples and reduces the likelihood of assigning a high rank to the target triple.

Among the target triples for inversion pattern, for FB15k-237, the model identifies \emph{/location/ hud\_county\_place/ place} as the inverse of \emph{/people/ ethnicity/ people}. While the two relations are not exact inverses of each other, their domain and range would loosely be inverse. Among the target triples for composition pattern, the model identifies \emph{film/ performance/ film} and  \emph{location/ mailing\_address/ citytown} as the composition for \emph{/people/ person/ gender}.

\begin{table}[]
    \centering
    \fontsize{10}{10}\selectfont
    \begin{tabularx}{\textwidth}{l >{\raggedright\arraybackslash}X}
    \toprule
       \textbf{Attack}  &   \multicolumn{1}{c}{\textbf{Target and Adversarial Triples}}  \\
    \midrule
    \multirow{2}{*}{\shortstack[l]{\textbf{Sym} \\ \textbf{Truth}}}
       &     \emph{date\_NN\_6} ,\  \emph{hypernym} ,\  \emph{month\_NN\_1}  \\
       &     \emph{date\_VB\_3} ,\  \emph{hypernym} ,\  \emph{date\_NN\_6}  \\
       &     \emph{month\_NN\_1} ,\  \emph{hypernym} ,\  \emph{week\_NN\_3}  \\ [8pt]
       
       
    \multirow{2}{*}{\shortstack[l]{\textbf{Sym} \\ \textbf{Rank}}}
       &     \emph{roman\_deity\_NN\_1} ,\  \emph{hypernym} ,\  \emph{immortal\_NN\_2}  \\
       &     \emph{greco-roman\_deity\_NN\_1} ,\  \emph{hypernym} ,\  \emph{roman\_deity\_NN\_1}  \\
       &     \emph{immortal\_NN\_2} ,\  \emph{hypernym} ,\  \emph{mythology\_NN\_2}  \\ [8pt]
       
       
    \multirow{2}{*}{\shortstack[l]{\textbf{Sym} \\ \textbf{Cos}}}
       &     \emph{scratch\_VB\_3} ,\  \emph{derivationally\_related\_form} ,\  \emph{scabies\_NN\_1}  \\
       &     \emph{itch\_VB\_2} ,\  \emph{derivationally\_related\_form} ,\  \emph{scratch\_VB\_3}  \\
       &     \emph{scabies\_NN\_1} ,\  \emph{derivationally\_related\_form} ,\  \emph{itching\_NN\_1}  \\ 
       
       
    \midrule
    
    \multirow{2}{*}{\shortstack[l]{\textbf{Inv} \\ \textbf{Truth}}}
       &     \emph{pyrrhocoridae\_NN\_1} ,\  \emph{hypernym} ,\  \emph{arthropod\_family\_NN\_1}  \\
       &     \emph{suborder\_heteroptera\_NN\_1} ,\  \emph{derivationally\_related\_form} ,\  \emph{pyrrhocoridae\_NN\_1}  \\
       &     \emph{arthropod\_family\_NN\_1} ,\  \emph{derivationally\_related\_form} ,\  \emph{kingdom\_animalia\_NN\_1}  \\ [8pt]
       
       
    \multirow{2}{*}{\shortstack[l]{\textbf{Inv} \\ \textbf{Rank}}}
       &     \emph{roman\_deity\_NN\_1} ,\  \emph{hypernym} ,\  \emph{immortal\_NN\_2}  \\
       &     \emph{greco-roman\_deity\_NN\_1} ,\  \emph{derivationally\_related\_form} ,\  \emph{roman\_deity\_NN\_1}  \\
       &     \emph{immortal\_NN\_2} ,\  \emph{derivationally\_related\_form} ,\  \emph{mythology\_NN\_2}  \\ [8pt]
       
       
    \multirow{2}{*}{\shortstack[l]{\textbf{Inv} \\ \textbf{Cos}}}
       &     \emph{flying\_lizard\_NN\_1} ,\  \emph{hypernym} ,\  \emph{agamid\_lizard\_NN\_1}  \\
       &     \emph{family\_agamidae\_NN\_1} ,\  \emph{derivationally\_related\_form} ,\  \emph{flying\_lizard\_NN\_1}  \\
       &     \emph{agamid\_lizard\_NN\_1} ,\  \emph{derivationally\_related\_form} ,\  \emph{genus\_draco\_NN\_1}  \\
       
    \midrule
    
    \multirow{2}{*}{\shortstack[l]{\textbf{Com} \\ \textbf{Truth}}}
       &     \emph{republic\_of\_guinea\_NN\_1} ,\  \emph{instance\_hypernym} ,\  \emph{african\_nation\_NN\_1}  \\ [5pt]
       &     \emph{republic\_of\_guinea\_NN\_1} ,\  \emph{hypernym} ,\  \emph{konakri\_NN\_1}  \\
       &     \emph{konakri\_NN\_1} ,\  \emph{member\_of\_domain\_usage} ,\  \emph{wetback\_NN\_1}  \\ [5pt]
       &     \emph{terminate\_VB\_1} ,\  \emph{hypernym} ,\  \emph{republic\_of\_cameroon\_NN\_1}  \\
       &     \emph{republic\_of\_cameroon\_NN\_1} ,\  \emph{member\_of\_domain\_usage} ,\  \emph{african\_nation\_NN\_1}  \\ [8pt]
       
       
    \multirow{2}{*}{\shortstack[l]{\textbf{Com} \\ \textbf{Rank}}}
       &     \emph{zaglossus\_NN\_1} ,\  \emph{member\_meronym} ,\  \emph{spiny\_anteater\_NN\_1}  \\ [5pt]
       &     \emph{zaglossus\_NN\_1} ,\  \emph{derivationally\_related\_form}  ,\  \emph{tachyglossidae\_NN\_1}  \\
       &     \emph{tachyglossidae\_NN\_1} ,\  \emph{hypernym} ,\  \emph{pipistrellus\_pipistrellus\_NN\_1}  \\ [5pt]
       &     \emph{zaglossus\_NN\_1} ,\  \emph{derivationally\_related\_form}  ,\  \emph{tachyglossidae\_NN\_1}   \\
       &     \emph{tachyglossidae\_NN\_1}  ,\  \emph{hypernym} ,\  \emph{spiny\_anteater\_NN\_1}  \\ [8pt]
       
       
    \multirow{2}{*}{\shortstack[l]{\textbf{Com} \\ \textbf{Cos}}}
       &     \emph{garnish\_NN\_2} ,\  \emph{hypernym} ,\  \emph{ornamentation\_NN\_2}  \\ [5pt]
       &     \emph{garnish\_NN\_2} ,\  \emph{derivationally\_related\_form}  ,\  \emph{interior\_decoration\_NN\_1}  \\
       &     \emph{interior\_decoration\_NN\_1} ,\  \emph{member\_meronym} ,\  \emph{dress\_VB\_9}  \\ [5pt]
       &     \emph{trim\_VB\_6} ,\  \emph{derivationally\_related\_form}  ,\  \emph{grace\_VB\_2}   \\
       &     \emph{grace\_VB\_2}  ,\  \emph{member\_meronym} ,\  \emph{ornamentation\_NN\_2}  \\

    \bottomrule
    \end{tabularx}
    \caption{Examples of target triples from WN18RR ComplEx with maximum change in rank due to adversarial attacks. For each target triple, the adversarial additions are shown for degrading ranks on the object-side \emph{(s,r,?)} and the subject-side \emph{(?,r,o)} queries. Symmetry and inversion attacks have two additions per target triple, whereas composition attacks have four additions per target triple.}
    \label{tab:rip_examples_wn18rr_complex_max}
\end{table}

\begin{table}
    \begin{subtable}{1\textwidth}
    \centering
    \fontsize{10}{10}\selectfont
     \setlength{\tabcolsep}{4.0pt}
  \begin{tabular}{  m{0.1\textwidth}  >{\raggedright\arraybackslash}p{0.16\textwidth}  >{\raggedright\arraybackslash}p{0.5\textwidth} >{\raggedright\arraybackslash}p{0.16\textwidth}  } 
  \toprule
    \textbf{Attack} & \multicolumn{3}{c}{\textbf{Target and Adversarial Triples}} \\ 
    \midrule
    \multirow{2}{*}{\shortstack[l]{\textbf{Sym} \\ \textbf{Truth}}}
    &   \emph{Oprah Winfrey}   &   \emph{/award/ ranked\_item/ appears\_in\_ranked\_lists./ award/ ranking/ list}    &  \emph{Time 100}   \\[8pt]
    &     \emph{The Women of Brewster Place}  &  \emph{/award/ ranked\_item/ appears\_in\_ranked\_lists./ award/ ranking/ list}  &  \emph{Oprah Winfrey} \\ [8pt]
    &      \emph{Time 100}  &  \emph{/award/ ranked\_item/ appears\_in\_ranked\_lists./ award/ ranking/ list}  &  \emph{Sarah Palin} \\ 
    
    &   & &   \\
    
    \multirow{2}{*}{\shortstack[l]{\textbf{Sym} \\ \textbf{Rank}}}
    &   \emph{Stanley Kubrick}   &   \emph{/people/ person/ religion}    &  \emph{Judaism}   \\ [8pt]
    &   \emph{Agnosticism}   &   \emph{/people/ person/ religion}    &  \emph{Stanley Kubrick}   \\ [8pt]
    &   \emph{Judaism}    &   \emph{/people/ person/ religion}    &  \emph{Sam Mendes}   \\ 
    &   & &   \\
    
    \multirow{2}{*}{\shortstack[l]{\textbf{Sym} \\ \textbf{Cos}}}
    &   \emph{Lieutenant}   &   \emph{/business/ job\_title/ people\_with\_this\_title./ business/ employment\_tenure/ company}    &  \emph{United States Navy}   \\ [8pt]
    &   \emph{United States of America}   &   \emph{/business/ job\_title/ people\_with\_this\_title./ business/ employment\_tenure/ company}    &  \emph{Lieutenant}   \\ [8pt]
    &   \emph{United States Navy}    &   \emph{/business/ job\_title/ people\_with\_this\_title./ business/ employment\_tenure/ company}    &  \emph{United States Army}   \\ 
    
    \midrule
    
    \multirow{2}{*}{\shortstack[l]{\textbf{Inv} \\ \textbf{Truth}}}
    &   \emph{Italian American}   &   \emph{/people/ ethnicity/ people}    &  \emph{John Travolta}   \\ [8pt]
    &   \emph{American English}   &   \emph{/location/ hud\_county\_place/ place}    &  \emph{Italian American}   \\ [8pt]
    &   \emph{John Travolta}   &   \emph{/location/ hud\_county\_place/ place}    &  \emph{United States of America}   \\ 
    &   & &   \\
    
    \multirow{2}{*}{\shortstack[l]{\textbf{Inv} \\ \textbf{Rank}}}
    &   \emph{Coldplay}   &   \emph{/common/ topic/ webpage./ common/ webpage/ category}    &  \emph{Official Website}   \\ [8pt]
    &   \emph{Lead vocalist}   &   \emph{/tv/ tv\_program/ languages}    &  \emph{Coldplay}   \\ [8pt]
    &   \emph{Official Website}   &   \emph{/tv/ tv\_program/ languages}    &  \emph{Jay-Z}   \\ 
    &   & &   \\
    
    \multirow{2}{*}{\shortstack[l]{\textbf{Inv} \\ \textbf{Cos}}}
    &   \emph{The Royal Conservatoire of Scotland}   &   \emph{/education/ educational\_institution/ students\_graduates./ education/ education/ student}    &  \emph{Alan Cumming}   \\ [8pt]
    &   \emph{Tony Award for Best Actor in a Musical}   &   \emph{/film/ film/ featured\_film\_locations}    &  \emph{The Royal Conservatoire of Scotland}   \\ [8pt]
    &   \emph{Alan Cumming}   &   \emph{/film/ film/ featured\_film\_locations}    &  \emph{United States of America}   \\ 
    
    \bottomrule
  \end{tabular}
  \caption{Symmetry and Inversion Attacks}
  \end{subtable}
\caption{Examples of target triples from FB15k-237 ComplEx with maximum change in rank due to adversarial attacks.For each target triple, the adversarial additions are shown for degrading ranks on the object-side \emph{(s,r,?)} and the subject-side \emph{(?,r,o)} queries. Symmetry and inversion attacks have two additions per target triple, whereas composition attacks have four additions per target triple.}
\label{tab:rip_examples_fb15k_complex_max}
\end{table}

\begin{table}
\ContinuedFloat
    \begin{subtable}{1\textwidth}
    \centering
    \fontsize{10}{10}\selectfont
     \setlength{\tabcolsep}{4.0pt}
  \begin{tabular}{  m{0.1\textwidth}  >{\raggedright\arraybackslash}p{0.16\textwidth}  >{\raggedright\arraybackslash}p{0.5\textwidth} >{\raggedright\arraybackslash}p{0.16\textwidth}  } 
  \toprule
    \textbf{Attack} & \multicolumn{3}{c}{\textbf{Target and Adversarial Triples}} \\ 
    \midrule
    \multirow{2}{*}{\shortstack[l]{\textbf{Com} \\ \textbf{Truth}}}
    &   \emph{Bryan Adams}   &   \emph{/people/ person/ gender}    &  \emph{Male}   \\
    &   & &   \\
    &     \emph{Bryan Adams}  &  \emph{/film/ actor/ film./ film/ performance/ film}  &  \emph{St. John's University} \\ 
    &      \emph{St. John's University}  &  \emph{/organization/ organization/ headquarters./ location/ mailing\_address/ citytown}  &  \emph{New York City} \\ 
    &   & &   \\
    &     \emph{Jane Lynch}  &  \emph{/film/ actor/ film./ film/ performance/ film}  &  \emph{Autism} \\ 
    &      \emph{Autism}  &  \emph{/organization/ organization/ headquarters./ location/ mailing\_address/ citytown}  &  \emph{Male} \\ 
    
    \midrule
    \multirow{2}{*}{\shortstack[l]{\textbf{Com} \\ \textbf{Rank}}}
    &   \emph{Mila Kunis}   &   \emph{/people/ person/ gender}    &  \emph{Female}   \\
    &   & &   \\
    &     \emph{Mila Kunis}  &  \emph{/film/ actor/ film./ film/ performance/ film}  &  \emph{Autism} \\ 
    &      \emph{Autism}  &  \emph{/organization/ organization/ headquarters./ location/ mailing\_address/ citytown}  &  \emph{Male} \\ 
    &   & &   \\
    &     \emph{Lily Tomlin}  &  \emph{/film/ actor/ film./ film/ performance/ film}  &  \emph{Into the Wild} \\ 
    &      \emph{Into the Wild}  &  \emph{/organization/ organization/ headquarters./ location/ mailing\_address/ citytown}  &  \emph{Female} \\ 
    
    \midrule
    \multirow{2}{*}{\shortstack[l]{\textbf{Com} \\ \textbf{Cos}}}
    &   \emph{Pink Floyd}   &   \emph{/common/ topic/ webpage./ common/ webpage/ category}    &  \emph{Official Website}   \\
    &   & &   \\
    &     \emph{Pink Floyd}  &  \emph{/award/ award\_winning\_work/ awards\_won./ award/ award\_honor/ award\_winner}  &  \emph{Zooey Deschanel} \\ 
    &      \emph{Zooey Deschanel}  &  \emph{/people/ person/ spouse\_s./ people/ marriage/ location\_of\_ceremony}  &  \emph{Los Angeles} \\ 
    &   & &   \\
    &     \emph{Grammy Award for Best Rock Vocal Performance by a Duo or Group}  &  \emph{/award/ award\_winning\_work/ awards\_won./ award/ award\_honor/ award\_winner}  &  \emph{Eagles} \\ 
    &      \emph{Eagles}  &  \emph{/people/ person/ spouse\_s./ people/ marriage/ location\_of\_ceremony}  &  \emph{Official Website} \\ 
    
    \bottomrule
  \end{tabular}
  \caption{Composition Attacks}
  \end{subtable}
\caption{Examples of target triples from FB15k-237 ComplEx (continued).}
\end{table}

\subsection{Comparison with Attribution Attacks}
In Chapter \ref{ch:instance_attribution}, the instance attribution methods were used to select adversarial additions by identifying the most influential triple for a target triple, and replacing one of the entities of the target triple with a dissimilar entity in the latent space. While this heuristic method overcomes the search space for adversarial additions, it still requires a search over the neighbourhood of the target triple. This section compares the effectiveness and efficiency of the attribution attacks from Chapter \ref{ch:instance_attribution} with the inference attacks proposed in this Chapter. The experimental setup is the same as the remaining experiments in this chapter - that is, two adversarial additions are crafted per target triple. To select two adversarial additions from attribution attacks, instead of the most influential triple, the top-2 most influential triples are selected and their entities are replaced with dissimilar entities.

\begin{table}[]
\centering
\small
\begin{tabular}{c  l  ll ll }

\toprule
     & & \multicolumn{2}{c}{\textbf{WN18RR}} & \multicolumn{2}{c}{\textbf{FB15k-237}} \\
   \cmidrule(lr){3-4}  \cmidrule(lr){5-6} 
     & & \textbf{MRR}   & \textbf{H@1}  & \textbf{MRR}   & \textbf{H@1} \\
\midrule
    \textbf{Original}  &       &     0.89         &   0.84      &    0.61          &   0.45    \\

\midrule
    \multirow{6}{*}{\shortstack[l]{\textbf{Attribution} \\ \textbf{Attacks}}}
    & Dot Metric               &     0.85(-5\%)	         &    0.80     &     0.55(-10\%)         &   0.42    \\
    & $\bm{\ell_2}$ Metric     &     0.80(-10\%)         &    0.75     &     0.54(-12\%)         &   0.40     \\
    & Cos Metric               &     0.80(-10\%)         &    0.74     &     0.54(-12\%)         &  0.41     \\
\cline{2-6} \\[-12pt] 
    & GD (dot)                 &     0.76(-15\%)        &    0.65      &      0.54(-12\%)        &  0.40    \\
    & GL ($\bm{\ell_2}$)       &     0.75(-16\%)        &    0.65       &     0.54(-13\%)         &  0.40           \\
    & GC (cos)                &      0.75(-16\%)        &    0.65        &    0.54(-11\%)          & 0.41           \\
    
\midrule
    \multirow{3}{*}{\shortstack[l]{\textbf{Inference} \\ \textbf{Attacks}}}
    & Sym          &     \textbf{0.56(-37\%)}         &    \textbf{0.24}      &    \textbf{0.51(-17\%)}          & \textbf{0.38}    \\
    & Inv          &     0.80(-10\%)        &    0.79      &    0.53(-14\%)          &   0.39    \\
    
    & Com        &    0.82(-8\%)     &   0.70     &     0.55(-11\%)         &  0.40     \\
     
\bottomrule    

\end{tabular}
\caption{Comparison of the effectiveness of adversarial additions based on Attribution Attacks and Inference Attacks for the ComplEx model on WN18RR and FB15k-237. For Inference Attacks, the best results for each relation inference pattern from Tables \ref{tab:rip_benchmark_mrr_WN18RR} and \ref{tab:rip_benchmark_mrr_FB15k-237} is reported here. Percentage values were computed before rounding off to two decimals.}
\label{tab:rip_performance_ia}
\end{table}

Table \ref{tab:rip_performance_ia} shows the predictive performance of ComplEx for WN18RR and FB15k-237 due to the attribution attacks and inference attacks. It is observed that the symmetry pattern based attacks outperform the other methods for both datasets. Furthermore, the instance attribution attacks perform similar to the inversion and composition patterns. As discussed in previous sections, this indicates the sensitivity of ComplEx to the symmetry pattern. However, another explanation for the better performance of inference attacks is that they exploit the KGE ranking protocol for evaluation. According to the KGE evaluation protocol (Section \ref{sec:kge_eval_protocol}), the existence of a missing target fact is decided by its rank against the possible entity corruptions. While attribution attacks target the score predicted for the target facts, inference attacks target their predicted rank. Thus, the inference attack strategy is likely more effective than attribution attacks. This also indicates the possibility of designing stronger adversarial attacks against KGE models by jointly targeting the predicted score as well as the rank of the target fact.

Table \ref{tab:rip_runtime_ia} provides a comparison of the runtime efficiency of different attack methods against DistMult model for WN18RR and FB15k-237. It is observed that the symmetry attacks are most efficient for both datasets - performing similar to the random baselines for WN18RR and better than the random baselines for FB15k-237. Additionally, as observed for adversarial deletions in Section \ref{sec:ia_runtime_analysis}, the attribution attacks (especially gradient similarity based attribution) are more efficient for WN18RR than for FB15k-237. This is because of their reliance on the number of triples in the neighbourhood of the target triple. On the other hand, the difference in the efficiency of inference attacks for WN18RR and FB15k-237 is similar to the difference in the efficiency of random baselines. This is because the efficiency of inference attacks depends on the number of relations, entities and target triples instead of neighbourhood triples. This observation provides empirical evidence to support the computational complexity analysis of inference patterns based attacks from Section \ref{sec:rip_complexity_analysis} of this Chapter.

\begin{table}[]
\centering
\small
\begin{tabular}{c  l  ll  }

\toprule
     & & \textbf{WN18RR} & \textbf{FB15k-237} \\
   
\midrule
    \multirow{5}{*}{\shortstack[l]{\textbf{Baseline} \\ \textbf{Attacks}}}
    & Random\_n              &  2.57    &     \textbf{25.35}     \\
    & Random\_g1              &  \textbf{2.13}    &     25.38   \\
    & Random\_g2                & 2.46     &   31.25     \\
\cline{2-4} \\[-12pt]
    & Direct-Add            &   19.30     &     257.68        \\
    & CRIAGE                 &   \textbf{4.91}    &      \textbf{6.03}    \\

\midrule
    \multirow{6}{*}{\shortstack[l]{\textbf{Attribution} \\ \textbf{Attacks}}}
    & Dot Metric               &    7.57          &      13.34          \\
    & $\bm{\ell_2}$ Metric     &    \textbf{7.43}          &      \textbf{9.94}           \\
    & Cos Metric               &    7.56          &      11.47    \\
\cline{2-4} \\[-12pt] 
    & GD (dot)                 &     \textbf{39.53}        &      \textbf{1810.68}           \\
    & GL ($\bm{\ell_2}$)       &     44.09        &      1885.22                  \\
    & GC (cos)                &      52.47        &      2130.87                 \\
    
\midrule
    \multirow{9}{*}{\shortstack[l]{\textbf{Inference} \\ \textbf{Attacks}}}
    & Sym\_truth          &     \textbf{3.02}         &       10.53         \\
    & Sym\_rank           &     3.08         &       10.33          \\
    & Sym\_cos            &     3.06         &       \textbf{9.72}   \\
\cline{2-4} \\[-12pt] 
    & Inv\_truth          &    5.84         &       10.66          \\
    & Inv\_rank           &    5.77         &       10.48                 \\
    & Inv\_cos            &    \textbf{5.65}          &      \textbf{10.05}                 \\
\cline{2-4} \\[-12pt]
    
    & Com\_truth        &   988.02      &     915.64          \\
    & Com\_rank        &   8.73      &       23.37        \\
    & Com\_cos        &    \textbf{8.02}     &        \textbf{22.05}       \\
     
\bottomrule    

\end{tabular}
\caption{Runtime efficiency of the baseline and proposed adversarial additions for DistMult on WN18RR and FB15k-237. The absolute time taken in seconds to generate the complete set of adversarial triples for all target triples is reported. For adversarial additions based on Attribution attacks, neighbourhood triples for each target triple were pre-computed. }
\label{tab:rip_runtime_ia}
\end{table}

\newpage
\section{Summary}
This chapter proposed data poisoning attacks against KGE models based on relation inference patterns like symmetry, inversion and composition. The evaluation of these attacks showed that the proposed attacks outperform state-of-the-art adversarial attacks. 
Since the proposed attacks rely on relation inference patterns, they can also be used to understand the predictive behaviour of KGE models. This is because if a KGE model is sensitive to a relation inference pattern, then that pattern should be an effective adversarial attack. 
It was observed that the attacks based on symmetry pattern generalize across all KGE models. This indicates the sensitivity of different KGE models to this inference pattern. 

Further investigation of the hypotheses about the effect of input graph connectivity and the existence of specific inference patterns in the datasets are interesting directions for future work. 
It is noteworthy that such investigation of attacks will likely be influenced by the choice of datasets. 
In this thesis, the benchmark datasets for link prediction (Section \ref{sec:kge_datasets}) have been used for evaluation. While there are intuitive assumptions about the inference patterns on these datasets, there is no study that formally measures and characterizes the existence of these inference patterns. This makes it challenging to verify the claims made about the inductive abilities of the KGE models, not only by the attacks proposed in this chapter but also by new KGE models proposed in the literature.

Thus, a promising step in understanding knowledge graph embeddings is to propose datasets and evaluation tasks that test varying degrees of specific inductive abilities (Section \ref{sec:related_miscellaneous}). 
These will help evaluate new models and serve as a testbed for poisoning attacks. Furthermore, specifications of model performance on datasets with different inference patterns will improve the usability of KGE models in high-stakes domains like healthcare and finance.

 \chapter{Related Work}
 \label{ch:related_work}
 The research in this thesis focuses on the adversarial vulnerabilities of representation learning for knowledge graphs. The aim of this chapter is to relate this focus to the wider literature on graph representation learning and on understanding the predictive behaviour of latent knowledge graph representations. The discussion in this chapter is not necessary to understand the main contributions of the thesis. However, these studies might serve as useful pointers for an interested reader.

\section{Graph Representation Learning}
Traditional methods for Machine Learning on graphs relied on hand engineered features which are extracted from the structure of the graph \citep{hamilton2017graphrepresentationlearning, hamilton2020graphrepresentationlearning}. 
More recent approaches instead learn these features through representation learning approaches. A graph representation learning algorithm encodes the structural information in a graph by learning a mapping from the nodes (or sub-graphs) into low dimensional feature vectors. The objective is to optimize the mapping such that embedding interactions in lower dimensional vector space reflect the structure of the original graph. A large number of approaches have been proposed for representation learning on graphs. %
\citet{hamilton2017graphrepresentationlearning, hamilton2020graphrepresentationlearning} proposed a unifying framework for the different representation learning approaches on graphs. 
Broadly, there are three families of algorithms based on the graph structural information they aim to preserve -- edges between entities, neighbourhood nodes of an entity, relational paths between entities. 

\subsection{Based on Edge Reconstruction}

Representation learning algorithms which have an objective function based on edge reconstruction, as proposed in \citet{cai2018comprehensive}, are designed particularly for relational graphs or knowledge graphs where the edges in the graph represent relations between entities. The entire relational graph can be considered a set of facts about the entities, represented as subject-relation-object triples. The most common ML task on such directed, labelled graphs is that of knowledge base completion by predicting the missing facts (also called link prediction) \citep{nickel2011rescal, bordes2013transe, yang2015distmult}. The learned embeddings, popularly called Knowledge Graph Embeddings (KGE), and the task of knowledge base completion are the focus of the research in this thesis, and are discussed in Chapter \ref{ch:background}.

The embedding algorithms rely on local and global patterns of relationships to generalize them. Node embeddings are learned by low-rank tensor factorization of the tensor representing the graph. Tensor factorization is an extension of matrix factorization where two matrices are used to represent the set of subjects and objects. However, since the same entity can occur as subject and object, \citet{nickel2011rescal} proposed RESCAL to learn joint embeddings for subjects and objects. A matrix of scores for the occurrence of a fact (or triple) can be obtained by a multi-linear product between the embeddings of subject, relation and object. %
This notion of dot product has been generalized to a ranking based scoring function which combines embedding vectors in such a way that the actual triples in the graph are ranked higher than those which are not in the graph \citep{cai2018comprehensive, trouillon2016complex}. Representation learning techniques based on ranking score based edge reconstruction differ mostly in the way they define this embedding combination or scoring function. 
As proposed in \citet{chandrahas2018towards}, TransE \citep{bordes2013transe} and TransR \citep{lin2015transr} use additive scoring functions while DistMult \citep{yang2015distmult}, HolE \citep{nickel2016hole} and ComplEx \citep{trouillon2016complex} use multiplicative scoring functions. In TransE \citep{bordes2013transe}, relations are interpreted as translations on low-dimensional entity embeddings and the object entity embedding is approximated as embedding of subject entity plus relation embedding. On the other hand, DistMult \citep{yang2015distmult} computes the score matrix as a dot product of the subject embedding, object embedding and relation matrix. This is extended in ComplEx \citep{trouillon2016complex} by replacing the dot product with a Hermitian product for complex values. 
Further discussion on the design of KGE models used for the research in this thesis, as well as the training and evaluation protocols for this family of models is available in Chapter \ref{ch:background}.

The KGE methods discussed above focus on simple knowledge graphs where the entities belong to a fixed, discrete set, and the graph is represented as RDF (Resource Description Framework) triples. However, KG representation can also include multimodal entities, relation attributes or a graph schema definition in Web Ontology Language (OWL). \citet{pezeshkpour2018MultimodalEmbedding, duranniepert2018KBLRN, kristiadi2019incorporatingliterals, pai2021NumericEmbeddings} propose different approaches to encode the multimodal information like images, text, numerical values and logical rules, and combine this encoding with latent vectors from simple KGE models. \citet{lacroix2020temporaltensordecomposition, messner2022temporalboxembedding} extend the ComplEx and BoxE models to temporal KGs with timestamped relations. \citet{chen2021owl2vec} proposes embeddings for OWL ontologies that represent the domain semantics through logic constructors and lexical information. The research in this thesis focuses on the KGE models for simple graph structures expressed as RDF triples.

\subsection{Based on Neighbourhood Aggregation}

Neighbourhood aggregation based representation learning relies on a node's local neighbourhood to learn its embedding. This is unlike edge reconstruction based techniques that utilize the local and global relation patterns in the graph. In neighbourhood aggregation, node attributes are used to initialize node embeddings. This initial embedding is iteratively updated using an aggregation function on the vectors representing neighbourhood nodes. Over the iterations, information from further nodes is aggregated into this node embedding \citep{hamilton2017graphrepresentationlearning}. 

A neural network based approach for such neighbourhood aggregation, called Graph Neural Network (GNN) was first proposed in \citet{scarselli2009gnn}. Inspired by traditional Recurrent Neural Network (RNN), this algorithm relied on repeated application of contraction maps as propagation (or aggregation) function until stable node embeddings were achieved. The GNN architecture was extended further in \citet{li2016ggnn} by using Gated Recurrent Unit (GRU). Instead of this RNN based aggregation, neighbourhood aggregation based on convolution operations was proposed in \citet{bruna2014spectral} and extended in \citet{defferrard2016convolutional}. 
The traditional Convolutional Neural Network (CNN) has been successful on Euclidean data which can be represented by a grid-like structure or vectors. These studies extend the convolution operation to graphs (non-Euclidean data) by using Graph Laplacian in the spectral domain. The spectral graph convolution operation is further simplified in \citet{kipf2017gcn} via a localized first order approximation which is equivalent to an element wise weighted mean aggregation function (spatial convolution). The use of convolution operations for aggregation is generalized to trainable aggregation functions in \citet{hamilton2017inductive}. \citet{bronstein2017geometric} provides an overview of additional similar methods.

Most graph representation learning algorithms based on neighbourhood aggregation are designed for undirected, unlabelled graphs. But, some of these have been generalized to relational (directed, labelled) graphs for link prediction. For example, R-GCN \citep{schlichtkrull2018rgcn} is a generalization of Graph Convolutional Network (GCN) \citep{kipf2017gcn} for relational graphs. However, the architecture does not scale to practical-scale knowledge graphs. This is because GNNs aggregate the neighbourhood information for each node, implying that the same neighbours need to be sampled multiple times. This quickly becomes a bottleneck for GNN implementation. For smaller graphs, the bottleneck is resolved by performing the convolution operation on the complete graph adjacency matrix at once \citep{hamilton2020graphrepresentationlearning}. However, this solution cannot be applied for large adjacency tensors for practical knowledge graphs. Therefore, neighbourhood aggregation based representation learning algorithms are not included in this thesis.

\subsection{Based on Random Walks}

Random walks based representation learning on graphs attempts to preserve the first order proximity or the second order proximity of a node in low dimensional vectors. First proximity refers to local pairwise proximity between nodes, while second order proximity refers to the proximity between the neighbourhood structures of nodes \citep{tang2015line}.  Node embeddings are generated by sampling a set of paths from the graph and maximizing the probability of observing a node's neighbourhood \citep{cai2018comprehensive}. %
Hence, nodes with similar neighbourhood have similar embeddings. The techniques differ in their sampling strategy and the node proximity to be preserved. 

DeepWalk \citep{perozzi2014deepwalk} uses truncated random walks to sample the paths in graph. These paths are treated like sequences of words where the nodes correspond to words. A SkipGram model is used to learn the embeddings by optimizing a hierarchical softmax function.
LINE \citep{tang2015line} uses a Breadth-first sampling (BFS) strategy to expand a node's context while sampling. The BFS approach captures homophily, where nodes belonging to same clusters are embedded closely. On the other hand, a Depth-first Sampling (DFS) strategy captures structural equivalence, where nodes with similar structural roles in the graph are embedded closely. Both BFS and DFS are used for expansion of a node's context in node2vec \citep{grover2016node2vec} via hyperparameters to bias the random walks. Both LINE and node2vec use negative sampling for optimizing the objective function. %
While DeepWalk preserves a node's second order proximity, LINE preserves both the first and second order proximity. node2vec preserves the first order proximity for homophilic and structurally equivalent nodes.

Since these methods encode the neighbourhood structural information to latent representations through random walks, they do not scale to knowledge graphs \citep{hamilton2020graphrepresentationlearning}. Hence, these methods are not included for research in this thesis.

\section{Understanding Knowledge Graph Representations}
\label{sec:related_understandingkge}
The latent distributed representations of knowledge graphs have achieved impressive predictive performance on a variety of knowledge-driven downstream tasks like question answering, fact checking, relation extraction, entity disambiguation and so on \citep{nickel2015review, ji2022knowledgegraphssurveyNLP, chen2020reasoningonknowledgegraphs}. This has made representation learning algorithms the most promising methods for reasoning with knowledge graphs in real-world scenarios. The graph representation learning algorithms are indeed increasingly used in high-stakes domains like healthcare and finance \citep{rotmensch2017healthkgelectronicmedicalrecords, mohamed2020knowledgegraphdrugdiscovery, bonner2021knowledgegraphdrugdiscovery}. 

However, deploying ML systems in real-world scenarios brings new challenges about the reliability and trustworthiness of these systems. The learned models are difficult to interpret and thus require novel methods to explain their predictions. Being black-box, the failure modes of the models are also unknown, which motivates research for designing adversarial attacks and investigating the security vulnerabilities of the models. Additionally, the reasons for impressive predictive performance of the black-box models are unknown. This performance is often achieved by extensive tuning of the model hyperparameters, which motivates research for dissecting and understanding the impact of these hyperparameters on the model predictions.

While the research in this thesis focuses on the adversarial vulnerabilities of Knowledge Graph Embeddings, complementary research on understanding the embeddings includes methods for post-hoc explanations, improved evaluation protocols, theoretical and empirical analyses, etc. This section highlights some literature on these related topics. A brief recap of state-of-the-art methods for adversarial attacks is included for completeness, but a deeper analysis relevant for the thesis contributions is available in Section \ref{sec:sota_kgeattack}.

\subsection{Via Robustness of Model Predictions}
\citet{zhang2019kgeattack} and \citet{pezeshkpour2019criage} propose data poisoning attacks against KGE models and are the most closely related to this research. These attacks are used as baselines to evaluate the thesis contributions. Additionally, \citet{lawrence2021gradientrollback} estimates the influence of training triples on the KGE model predictions to provide post-hoc explanations. The proposed method can also be used for adversarial deletions and is thus, included as a baseline in this research.

In parallel work, \citet{banerjee2021stealthypoisoningKGE} studies risk aware adversarial attacks against KGE models with the aim of reducing the exposure risk of an adversarial attack instead of improving the attack effectiveness. This study is not included in the evaluation protocol for the thesis as the research here focuses on improving the attack effectiveness. \citet{zhang2021adversarialattackcrosslingualalignment} investigates the adversarial vulnerabilities of cross-lingual knowledge alignment through perturbations of knowledge graphs. Another recent study \citet{raman2021deceivingknowledgeaugmentedmodels} designs deceptive perturbations to the knowledge graph that maximally modify the graph structure while preserving the performance of downstream tasks on the graph. The study aims to investigate the role of knowledge graphs in knowledge-driven applications. Most recently, \citet{betz2022AdversarialExplanationsKnowledgeGraphEmbeddings} proposes adversarial attacks against KGE models for gray-box settings. Similar to the adversarial attacks in Chapter \ref{ch:relation_inference} of the thesis, relation inference patterns are exploited to select adversarial edits. However, instead of extracting the inferred patterns from the learned embedding, the study mines logical rules directly from the knowledge graph. This enables the selection of adversarial deletions and additions without knowledge of the victim KGE model. 

\citet{pujara2017kgsparsity} also investigates the failure modes of the KGE models. However, instead of designing adversarial perturbations to the input knowledge graph, the study generates systematic variations of the datasets. Using this suite of datasets, the sensitivity of KGE models to sparse and unreliable data is examined. 

Besides data poisoning attacks, \citet{minervini2017adversarialsets} uses adversarial regularization in the latent space to inject the background knowledge expressed as First Order Logic (FOL) clauses into KGE models. Embedding perturbations that do not comply with the logical constraints are penalized through an inconsistency loss. Given useful prior constraints, the regularization approach improves the predictive performance of KGE models for link prediction. However, these adversarial samples are not in the input domain and aim to improve instead of degrade the model performance. Similarly, \citet{cai2018kbgan} uses adversarial learning to improve the negative sampling process during the KGE model training. 
In addition to the risk of adversarial manipulation of training data, trustworthy deployment of KGE models also needs to address unintentional changes in the data distribution. In particular, \citet{fisher2020debiasingkge} shows that KGE models can encode social biases represented in the training data, and result in unfair predictions for certain populations. The study further proposes a debiasing method to train KGE models that are neutral towards sensitive attributes. This research is extended in \citet{keidar2021automaticbiasdetection} by identifying the sensitive relations to be debiased automatically. On the other hand, \citet{du2022understandinggenderbias} proposes novel measures to quantify the gender bias in KG embeddings and attributes the origin of this bias to training instances using Influence Function \citep{koh2017understanding}.

\subsection{Via Explainability of Model Predictions}
Given a trained KGE model, a post-hoc explanation method explains the model predictions on specific target facts. 
The post-hoc explanations can constitute triples, relation paths, or sub-graphs from the input knowledge graph.
\citet{lawrence2021gradientrollback} proposes Gradient Rollback to estimate the influence of training triples on model predictions and provides a set of top-k most influential triples as post-hoc explanation.
\citet{zhang2019interaction} provides explanations for KGE model predictions as meaningful paths between the subject and object of the triple to be explained. The study uses the embeddings of entities and relations to search for reliable explanation paths. In addition to closed paths between the subject and object, similar explanation structures from the remaining graph are also generated as support for the main explanation.
\citet{nandwani2020oxkbc} also provides relation paths as explanations but focuses on tensor factorization based KGE models. Similarity between the entity embeddings is used to augment the knowledge graph with weighted edges and similar explanation paths are aggregated into second-order templates. A separate neural module is trained to select the best template as the explanation.

Evaluating the methods for post-hoc KGE model explanations can be challenging because of the lack of ground-truth explanations. Thus, \citet{halliwell2021explanationbenchmark} created a benchmark dataset of explanations for missing link prediction on knowledge graphs. This dataset is built using a semantic reasoner on a fixed set of rules.

Unlike the non-interpretable nature of latent representations, rule-based symbolic reasoning on knowledge graphs is inherently interpretable \citep{chen2020reasoningonknowledgegraphs}. This has led to research efforts for combining rule-based reasoning with black-box latent representations to improve the latter's explainability. One research direction here constraints the knowledge graph embedding space by using background logical facts. \citet{guo2016kale} and \citet{guo2018ruge} enforce soft logical rules by modelling the triples and rules in a unified framework and jointly learning embeddings from them. Additional studies that inject logical rules into knowledge graph embeddings include \citet{rocktaschel2015injectingbackgroundknowledge}, \citet{demeester2016ruleinjection} and \citet{minervini2017adversarialsets}. Further, \citet{duranniepert2018KBLRN} combines the logical rules with latent and numerical features in an end-to-end framework using the product of experts approach.
The soft-logical model of relation inference patterns, proposed for adversarial attacks in Chapter \ref{ch:relation_inference} of this thesis is inspired by the literature on joint modelling of facts and rules \citep{guo2016kale, guo2018ruge}.

While rule-based reasoning is interpretable, logical rules are often not readily available. The methods described above rely on state-of-the-art rule mining systems like AMIE+ \citep{galrraga2015AMIE} and AnyBURL \citep{meilicke2019AnyBURL} to obtain the background logical facts. However, mining arbitrary rules from knowledge bases can be computationally expensive and constraining the type of rules to be extracted requires a manual specification prior to rule extraction \citep{chen2020reasoningonknowledgegraphs}. Further, injecting the logical rules to neural networks requires groundings of the rules, the size of which can be very large \citep{cohen2020TensorLogAP}. Thus, another research direction in combining symbolic rules with black-box neural embeddings focuses on the joint learning of logical rules and latent features from knowledge graphs. Neural Theorem Provers (NTPs) \citep{rocktaschel2017ntp} replace symbols in the logical rules with differentiable vector representations, which allows the rules to be learned from the knowledge graphs through backward chaining.
However, the search space for NTPs is exponentially large. This issue is addressed in \citet{minervini2020gntp, minervini2020ctp} by selecting the rules greedily or conditionally. In \citet{arakelyan2021complexqueryanswering}, the NTPs are further extended to answer complex path queries on knowledge graphs in an efficient and explainable manner. On the other hand, \citet{qu2021rnnlogic} uses a probabilistic approach for rule learning that treats logic rules as latent variables. The use of a rule generator and reasoning predictor effectively reduces the search space. 
Another system for learning rules from knowledge graphs is MINERVA \citep{das2018MINERVA}, which uses a reinforcement learning agent to learn a policy for graph traversal as an optimal sequence of decisions. %
Further, \citet{das2020CaseBasedReasoning} proposes a non-parametric approach based on case-based reasoning that does not rely on latent embeddings for reasoning. 
Similarly, end-to-end differentiable systems like NeuraLP \citep{yang2017neuralLP} and DRUM \citep{sadeghian2019DRUM} learn logical rules without latent features.

Despite these advances in interpretable methods for knowledge base reasoning, the rule injection methods remain difficult to scale and the rule learning methods only perform well in simple settings \citep{Jong2019NeuralTP}. Thus, black-box KGE models remain the most popular method for reasoning on knowledge graphs.

\subsection{Via Miscellaneous Methods}
\label{sec:related_miscellaneous}
Over the years, a large number of KGE models have been proposed, and have achieved impressive predictive performance. However, due to multiple components in the KGE training pipeline, there is wide variation in the implementation, training and evaluation of the proposed models \citep{ali2021bringinglightintodark}. Thus, dissecting and investigating the impact of different training components has become important for understanding the KGE model behaviour. One of the earlier studies in this direction was \citet{kadlec2017kgebaselines} which showed that simple KGE architectures can outperform newer and more sophisticated ones by proper hyperparameter tuning. Following this, \citet{mohamed2019kgelossfunctions} discovered that the KGE model's predictive performance is not only impacted by its scoring function, but also by the loss function used to train the model. %
Furthermore, \citet{lacroix2018canonical} showed that the model performance can be improved for the same scoring function by adding reciprocal triples to the training dataset. 

In response to these findings, studies on re-evaluating and benchmarking the predictive performance of KGE models have emerged. \citet{ruffinelli2020olddognewtricks} evaluated five KGE models with different loss functions and training strategies. The study found that early scoring functions like RESCAL \citep{nickel2011rescal} show comparable performance to more recent and advanced architectures when trained with state-of-the-art techniques. %
More recently, this study has been extended in \citet{ali2021bringinglightintodark} to perform a comprehensive benchmarking of the KGE model performance. %
Another recent survey \citet{rossi2021kgesurvey} investigates the effect of different graph structural properties on the predictive performance of KGE models. 

Besides understanding the training phase of KGE models, some studies have focused on identifying inconsistencies in the evaluation protocol. \citet{sun2020kgeevaluation} highlighted that the strong predictive performance of KGE models can be attributed to an incorrect evaluation implementation. The study further proposed an evaluation protocol for assigning the ranks when multiple triples have the same score. In the same direction, \citet{pezeshkpour2020revisitingkgeevaluation} highlighted the unreliability of the standard evaluation metrics and \citet{rim2021kgechecklist} proposed to evaluate KGE models using checklists \citep{ribeiro2020checklist}.

In the literature besides benchmarking and reproducibility, \citet{trouillon2019inductive} studies the inductive abilities of KGE models as binary relation properties for controlled inference tasks with synthetic datasets. On the theoretical side, \citet{wang2018multi} studies the expressiveness of various bilinear KGE models and \citet{basulto2018ontologyembedding} studies the ability of KGE models to learn hard rules expressed as ontological knowledge. More recently, \citet{allen2021interpreting} provides a theoretical connection between the word embeddings and knowledge graph representations. The study proposes a model for encoding the semantic graph relationships as geometric interactions between the latent representations.

\section{Summary}
This chapter provided a discussion of the related work for this thesis. The first part discussed families of graph representation learning algorithms including those that do not scale to knowledge graphs. The second part highlighted complementary literature on understanding the predictive behaviour of KGE models.

\chapter{Conclusion}
\label{ch:conclusion}

\section{Contributions}
This thesis presents the research undertaken towards the adversarial robustness of representation learning on knowledge graphs through the design, implementation and evaluation of data poisoning attacks against them. Representation learning on knowledge graphs is increasingly used in high-stakes domains where the decisions based on the learned representations impact human lives. However, by virtue of being data driven, the representation learning algorithms are black-box and thus, have unknown failure modes. 
Responsible deployment of representation learning algorithms for knowledge graphs in user-facing applications requires an investigation of their security vulnerabilities and robustness against adversarial attacks. 
Towards this investigation, the thesis contributes two data poisoning attacks that \emph{effectively} degrade the predictive performance of the Knowledge Graph Embeddings at inference time by removing or adding triples to the knowledge graph at training time.
\begin{enumerate}
    \item The first contribution of the thesis is poisoning attacks based on instance attribution methods (Chapter \ref{ch:instance_attribution}). The instance attribution methods from Interpretable Machine Learning identify the input data points that are most influential to a model's prediction. These methods are used to select the influential triples as adversarial deletions, that is triples in the training graph that on removal degrade the learned model's predictive performance. Furthermore, a heuristic approach is proposed to select adversarial additions based on the influential triples.
    \item The second contribution of the thesis is poisoning attacks based on the relation inference patterns (Chapter \ref{ch:relation_inference}). These attacks exploit the inductive abilities of KGE models which are expressed as connectivity patterns among the relationships in the knowledge graph, for example, symmetry, inversion and composition. To degrade the KGE model's  predicted ranks on target triples, different relation inference patterns are exploited to improve the performance on decoy triples. This problem formulation breaks down the original combinatorial search space for adversarial additions into three smaller steps for determining the adversarial relations, the decoy entities that most likely violate an inference pattern and the remaining adversarial entities in the inference pattern that are most likely to improve the rank of decoy triples. Unlike the adversarial additions based on instance attribution methods, executing these attacks does not require an enumeration over the neighbourhood triples of the target triple. Instead, the computational complexity is an additive function of the number of entities and relations.
\end{enumerate}

The proposed attacks are evaluated by comparing the predictive performance of the original and the poisoned KGE models on the target triples that are ranked best by the original model. The experiments for four KGE models DistMult, ComplEx, ConvE and TransE on two benchmark datasets WN18RR and FB15k-237 indicate that the proposed attacks are more effective than the baselines based on random edits and state-of-the-art adversarial attacks against KGE models. It was found that the attacks based on simpler instance attribution methods and simpler relation patterns are more or as effective as the more complex attacks. Thus, in a hypothetical white-box attack scenario, an attacker might efficiently execute highly effective adversarial attacks against Knowledge Graph Embeddings.
Additionally, models from all the different representative families are observed to be vulnerable to degradation in their predictive performance. The proposed attacks were also found to be more effective for the sparser WN18RR dataset than the FB15k-237. This indicates that the embeddings learned from fewer training triples are more vulnerable to adversarial manipulation of the training knowledge graph.

The KGE models evaluated in this research have shown competitive predictive performance in recent literature \citep{kadlec2017kgebaselines, ruffinelli2020olddognewtricks} through proper hyperparameter tuning. This makes them promising candidates for use in production pipelines. However, the research in this thesis highlights the brittleness of these performance gains and thus, calls for improved KGE model evaluation that accounts for adversarial robustness in addition to the predictive performance.

\section{Limitations}

While this thesis clearly highlights the adversarial vulnerabilities of KGE models, there are some constraints and limitations of the research presented in the thesis.

\subsection{Evaluation Design}
The conclusions drawn from this research are based on evaluation of the proposed attacks against four KGE models on two benchmark datasets. The choices made towards KGE model selection, dataset selection and implementation design have likely influenced the outcomes of the evaluation. For example, a large number of KGE models have been proposed in the literature, but this research focuses on four representative models from the three families of KGE models (tensor decomposition, geometric and deep learning models as discussed in \citet{rossi2021kgesurvey}). While the KGE models are differentiated through their scoring function design, each KGE model can be trained through multiple loss functions like pointwise, pairwise or setwise loss functions \citep{ali2021bringinglightintodark}. Additionally, there are multiple training strategies like negative sampling, 1vsAll and KvsAll \citep{ruffinelli2020olddognewtricks}. It is very likely that the effectiveness of the proposed adversarial attacks would vary for different scoring functions, loss functions and training design.

In the recent literature, multiple studies \citep{ali2021bringinglightintodark, rossi2021kgesurvey, kadlec2017kgebaselines, ruffinelli2020olddognewtricks} on benchmarking the predictive performance of KGE models have been conducted. While these studies focus on the reproducibility of the predictive performance, their design and implementation suggests that a similar study for evaluating the adversarial robustness of KGE models is beyond the scope of this thesis. Nevertheless, a detailed comparative study of the adversarial vulnerabilities of KGE models is a useful direction for future work.

In addition to above design choices, the conclusions rely on characteristics of the datasets chosen in this research. The datasets used here are the benchmark datasets for link prediction using KGE models. It was observed that the instance attribution attacks are more effective on the sparser dataset WN18RR. Similarly, the addition of inversion inference pattern to WN18RR affected the effectiveness of relation inference patterns attacks against ComplEx. 
However, the characteristics of these benchmark datasets need not necessarily be representative of the real-world knowledge graphs in safety-critical domains. For example, the qualitative analysis in Section \ref{sec:ia_qualitative} showed that identifying the correct influential triples for FB15k-237 is challenging because of the semantically incoherent triples in the test set. Additionally, there are limited, though only recent, studies \citep{cao2021inferwiki, ali2021bringinglightintodark} that formally measure and characterize the connectivity patterns or relation inference patterns in benchmark datasets for link prediction. This makes it challenging to generalize the claims made about the effectiveness of adversarial attacks in this research. Analyzing the impact of dataset characteristics on the effectiveness of adversarial attacks is an interesting direction for future work. 

\subsection{Design of Adversarial Attacks}
The relation inference patterns based attacks in this thesis use three steps, of which, the first step is to determine the adversarial relations (Table \ref{tab:rip_attack_summary}). This step utilizes the relation embeddings from KGE models to select the inverse or the composition relations for the given target triple. The algebraic model of inference patterns used for this selection assumes that the scoring function of KGE model is either additive or multiplicative in nature \citep{chandrahas2018towards}. However, some recent KGE models, like RotatE \citep{sun2018rotate} and MuRE \citep{balazevic2019mure}, have both or neither additive and multiplicative components in their scoring functions. Hence, the proposed attacks based on relation inference patterns are only applicable for KGE models with additive or multiplicative scoring functions.

\section{Broader Impact}

This thesis studies the problem of generating data poisoning attacks against KGE models. Data poisoning attacks identify the vulnerabilities in learning algorithms that could be exploited by an adversary to manipulate the model's behaviour \citep{joseph_nelson_rubinstein_tygar_2019, biggio2018wild}. Such manipulation can lead to unintended model behaviour and failure. Identifying these vulnerabilities for KGE models is critical because of their increasing use in domains that need high stakes decision making like heathcare \citep{rotmensch2017healthkgelectronicmedicalrecords, mohamed2020knowledgegraphdrugdiscovery} and finance \citep{hogan2021knowledgegraphs, noy2019knowledgegraphs}. 
In this way, the research presented here aims to safeguard the KGE models against potential harm from adversaries and thus, minimize the negative consequences of deploying these models in user facing applications that require trustworthy predictions.

Arguably, because the vulnerabilities are studied by attacking the KGE models, the proposed attacks can be used by an actual adversary to manipulate the predictive behaviour of models in deployed systems. This paradox of an arms race is universal across security research \citep{biggio2018wild}. For this thesis, the principle of proactive security design, as recommended by \citet{joseph_nelson_rubinstein_tygar_2019} and \citet{biggio2018wild} has been followed. As opposed to reactive security measures where the learning system designers develop countermeasures after the system is attacked, a proactive approach \emph{anticipates} such attacks, simulates them and designs countermeasures before the systems are deployed. Therefore, by revealing the vulnerabilities of KGE, this thesis provides an opportunity to fix them.

Besides the use case of security evaluation, the research can be helpful in understanding the predictive behaviour of KGE models. To propose adversarial attacks based on adversarial deletions, the research uses Instance Attribution methods from Interpretable Machine Learning. These methods identify the most influential triples in the training knowledge graph which can also be used as post-hoc explanations for the learned KGE models (Section \ref{sec:ia_qualitative}). In addition to explaining model predictions, instance based attribution methods can help guide the design decisions during KGE model training. As discussed earlier, there are a vast number of KGE model architectures, loss functions and training strategies. Empirically quantifying the impact of different design choices on the predictive performance of KGE models is often challenging and requires extensive experimental resources \citep{ruffinelli2020olddognewtricks, ali2021bringinglightintodark, rossi2021kgesurvey}. Thus, it would be very promising to explore the use of instance attribution methods to understand the impact of these choices on the KGE model predictions. By tracing back the model predictions to the input knowledge graph, one can gain a better understanding of the success or failure of different design choices.

The proposed attacks based on relation inference patterns can also be used to understand the inductive abilities of KGE models. These attacks rely on the inductive assumptions of a model, which are expressed as the different relation inference patterns, to be able to deceive that model. Thus, theoretically, the effectiveness of attacks based on one inference pattern over another indicates the model's reliance on one inference pattern over another. However, as discussed in the previous section, it is challenging to generalize the conclusions drawn from this research about the inductive abilities of KGE models. This is because the inference patterns for benchmark datasets are not well defined, and these datasets may not be representative of the high-stakes domains.

Thus, there is scope for further research to evaluate the adversarial attacks proposed in this thesis. A promising direction here is to design benchmark tasks and datasets that measure the specific inductive abilities of models. This will not only be useful for evaluating the proposed attacks, but also for understanding the inductive abilities of existing KGE models. This in turn, can guide the community to design better algorithms for representation learning on knowledge graphs. In this direction, research on novel architectures for knowledge graph embeddings should evaluate not only the predictive performance on benchmark datasets, but also the claims made on inductive abilities of these models and their robustness to violations of these implicit assumptions. 

In summary, through methods that lead to KGE models' failure (that is, adversarial attacks), the research presented here aims to provide an understanding of the predictive behaviour of these models, as well as improve the utility of KGE models in user applications that require trustworthy predictions.

\section{Future Directions}

As is the nature of research, the investigation in this thesis brings to light several open research questions for future research. 

\subsection{Additional Evaluation}
As discussed in the limitations of the current research, the conclusions drawn from the research are difficult to generalize because of the evaluation constraints. Thus, an immediate direction for follow-up research is to investigate hypotheses about the effect of dataset characteristics like graph connectivity and existence of specific relation inference patterns. Such investigation would likely require novel datasets and evaluation tasks that test varying degrees of dataset characteristics. These tasks will not only serve as a testbed for the proposed poisoning attacks, but also help evaluate new KGE models. Furthermore, fine-grained comprehensive evaluation under the adversarial perturbations can improve our understanding of the KGE model behaviour. This way, the specifications of model performance and failure on datasets with different characteristics will improve the usability of KGE models in high-stakes domains like healthcare and finance.

\subsection{Additional Threat Models}
The second direction for future work is to evaluate the security vulnerabilities of KGE models in more realistic settings, through investigating the adversarial robustness of KGE models for additional threat models. 
For example, in this research, only triples in the neighbourhood of the target triple are selected as candidates for adversarial perturbations. However, in a realistic attack setting, the attacker might have limited or no access to the neighbourhood of the target triple. Thus, for KGE models deployed in actual ML pipelines, can the predictive performance be degraded through perturbations beyond the neighbourhood of the target triple? Similarly, the perturbations that are adversarial for one target triple may be non-adversarial for another target triple. Thus, another realistic attack setting is to craft the adversarial perturbations for multiple target triples jointly. 

Additionally, the adversarial attacks designed in this research are integrity attacks. Here, the attacker aims to make discreet edits to the training samples; and target the model's predictive performance on specific target samples without affecting the performance on the remaining samples. The notion of discreet or unnoticeable perturbations in this research is defined by the attacker's budget for the number of adversarial perturbations. However, there is scope for further research on encoding the attack unnoticeability in the design of adversarial attacks itself.

Lastly, the attacks in this thesis assume a white-box threat model. This means that the attacker has access to the learned embeddings from the KGE model as well as the complete training dataset. However, in more realistic scenarios where the KGE models are deployed, the attacker may not have access to either or both of these. Designing adversarial attacks that are constrained to black-box settings will provide a more realistic evaluation of the security vulnerabilities of KGE models.

\subsection{Adversarially Robust Knowledge Graph Embeddings}
The contributions of this thesis indicate that the sensitivity of state-of-the-art KGE models to adversarial perturbations can introduce security vulnerabilities in pipelines that use knowledge graph embeddings. Thus, a promising direction for future work is towards mitigating the security vulnerabilities of KGE models. 
Some preliminary ideas to improve the adversarial robustness of KGE models are to use adversarial training techniques; or train an ensemble of different KGE scoring functions; or train an ensemble from different subsets of the training dataset. Since the evaluation of relation inference attacks shows that state-of-the-art KGE models are sensitive to symmetry pattern, it is important to investigate neural architectures that generalize beyond symmetry even though their predictive performance on the benchmark datasets might not be the best.

However, it is noteworthy that building defences against specific adversarial attacks can lead to an arms race where the defences can be sabotaged by stronger adversarial attacks. This is because adversarial attacks are empirical methods to evaluate the adversarial robustness of Machine Learning systems and do not provide any formal guarantees for the learned model's robustness. Thus, to improve their usability in safety-critical domains, it is crucial to build novel KGE models that are certifiably robust and provide provable guarantees for their predictions.

\section{Concluding Statement}
To conclude, given the increasing effectiveness of graph ML systems, this thesis contributes towards their responsible integration in daily lives. The proposed methods for adversarial attacks improve our understanding of the adversarial vulnerabilities and the predictive performance of representation learning algorithms for knowledge graphs, providing a foundation for further research towards adversarially robust and trustworthy graph Machine Learning.

\appendix
\cleardoublepage

\chapter{Additional Details for Instance Attribution Attacks}
\section{Training Details}
\label{apx:instance_attribution_training_details}
\subsection{Training KGE models} 
\begin{table}[]
    \centering
    \small
    \begin{tabular}{c  c c   c c}
    \toprule
                  &  \multicolumn{2}{c}{\textbf{WN18RR}}       &   \multicolumn{2}{c}{\textbf{FB15k-237}}   \\
                  \cmidrule(lr){2-3}  \cmidrule(lr){4-5}  
                  &  \textbf{MRR}   &  \textbf{Hits@1}    &   \textbf{MRR}   &  \textbf{Hits@1}  \\
    \midrule
         DistMult &     0.48   &   0.44    &  0.34  &  0.24  \\
         ComplEx  &     0.51   &   0.47    &  0.34  &  0.25   \\
         ConvE    &     0.44   &   0.41    &  0.32  &  0.23  \\
         TransE   &     0.21   &   0.02    &  0.33  &  0.24  \\
    \bottomrule
    \end{tabular}
    \caption{MRR and Hits@1 results for original KGE models on WN18RR and FB15k-237}
    \label{tab:ia_original_mrr}
\end{table}

In this research, four KGE models are implemented - DistMult, ComplEx, ConvE and TransE. 
The 1-N training strategy proposed in \citet{lacroix2018canonical} is used but the reciprocal relations are not added to the training dataset.
Thus, for each triple, scores are generated for $(s,r) \rightarrow o$ and $(o,r) \rightarrow s$.

For TransE scoring function, the L2 norm is used. The loss function used for all models is Pytorch's $\mathtt{ CrossEntropyLoss}$. For regularization, N3 regularization and input dropout are used on DistMult and ComplEx; input dropout, hidden dropout and feature dropout on ConvE; and L2 regularization \citep{bordes2013transe} and input dropout for TransE. 

To ensure same hyperparameters for original and poisoned KGE models, early stopping is not used. An embedding size of 200 is used for all models on both datasets. An exception to this is the TransE model for WN18RR, where embedding dim = 100 was used due to the expensive time and space complexity of 1-N training for TransE.
The hyperparameters for KGE model training were tuned manually based on suggestions from state-of-art implementations \citep{ruffinelli2020olddognewtricks, dettmers2018conve, lacroix2018canonical, ampligraph}.

Table \ref{tab:ia_original_mrr} shows the MRR and Hits@1 for the original KGE models on WN18RR and FB15k-237. To re-train the KGE model on poisoned dataset, same hyperparameters as the original model were used.
All experiments for model training, adversarial attacks and evaluation were run on a shared HPC cluster with Nvidia RTX 2080ti, Tesla K40 and V100 GPUs. 

To ensure reproducibility, the source code of the experiments in this research is publicly available on GitHub at \url{https://github.com/PeruBhardwaj/AttributionAttack}. 
Results of experiments for evaluating the contributions in Chapter \ref{ch:instance_attribution} were obtained from the commit $0d5ca33$ (\url{https://github.com/PeruBhardwaj/AttributionAttack/tree/0d5ca33}).

These results can be reproduced by passing the argument $\mathtt{reproduce-results}$ to the attack scripts. Example commands for this are available in the bash scripts in the codebase. The hyperparameter used to generate the results can be inspected in the $\mathtt{set\_hyperparams()}$ function in the file $\mathtt{utils.py}$ or in the log files.

For the LissA algorithm used to estimate the Hessian inverse in Influence Functions, the hyperparameter values are selected using suggestions from the original study by \citet{koh2017understanding}. The values are selected to ensure that the Taylor expansion in the estimator converges. These hyperparameter values for the reported experiments are available in the function $\mathtt{set\_if\_params()}$ in the file $\mathtt{utils.py}$ of the accompanying codebase.

\subsection{Baseline Implementation Details}
One of the baselines in Section \ref{sec:ia_evaluation} is the Direct-Del and Direct-Add attack from \cite{zhang2019kgeattack}. The original study evaluated the method for the neighbourhood of subject of the target triple. This setting is extended to both subject and object to ensure fair comparison with other attacks. Since no public implementation of the attacks is available, the author implement the attacks herself.

\begin{table}
    \centering
    \small
    \begin{tabular}{c  ccc   }
    \toprule
                  &  \multicolumn{3}{c}{\textbf{WN18RR}}         \\
                  \cmidrule(lr){2-4} 
                  &  \textbf{Original}   &  \textbf{High}    &   \textbf{Low}     \\
    \midrule
         DistMult &     1.00      &   0.98        &   0.98        \\
         ComplEx  &     1.00      &   0.96        &   0.95       \\
         ConvE    &     1.00      &   0.99       &   0.99       \\
         TransE   &     1.00      &   0.81        &   0.86        \\
         
    \midrule
                 &   \multicolumn{3}{c}{\textbf{FB15k-237}}   \\
                  \cmidrule(lr){2-4}
                  &  \textbf{Original}   &  \textbf{High}    &   \textbf{Low} \\
    \midrule
        DistMult    &  1.00     &  0.64    &   0.64     \\
        ComplEx     &  1.00     &  0.67    &   0.66     \\
        ConvE       &  1.00     &  0.62     &   0.60    \\
        TransE      &  1.00     &  0.72    &   0.73     \\
                  
    \bottomrule
    \end{tabular}
    \caption{MRR of KGE models trained on original datasets and poisoned datasets from the Direct-Add baseline attack in \citet{zhang2019kgeattack}. High, Low indicate the high (20\%) and low percentage (5\%) of candidates from random down-sampling. }
    \label{tab:ia_ijcai_results}
\end{table}

The Direct-Add attack is based on computing a perturbation score for all possible candidate additions. Since the search space for candidate additions is of the order $\mathcal{E} \times \mathcal{R}$ (where $\mathcal{E}$ and $\mathcal{R}$ are the set of entities and relations), it uses random down sampling to filter out the candidates. The percent of triples down sampled are not reported in the original paper and a public implementation is not available. So, in this research, a high value and a low value are picked for the percentage of triples to be down-sampled and adversarial additions are generated for both fractions. 20\% of all candidate additions are arbitrarily chosen for high; and 5\% of all candidate additions as low.

Thus, two poisoned datasets are generated from the attack - one that used a high number of candidates and another that used a low number of candidates. Two separate KGE models are trained on these datasets to assess the baseline performance. Table \ref{tab:ia_ijcai_results} shows the MRR of the original model; and poisoned KGE models from attack with high and low down-sampling percents. The results reported for Direct-Add in Section \ref{sec:ia_evaluation} of the thesis are the better of the two results (which show more degradation in performance) for each combination.

\newpage

\section{Additional Comparison with CRIAGE}
\label{apx:criage_bce}
The baseline attack method CRIAGE estimates the influence of a training triple using the BCE loss and is thus likely to be effective only for KGE models that are trained with BCE loss. In Section \ref{sec:ia_eval_baseline}, it was found that the proposed attacks are more effective than the baseline attack.
However, since the original models in this research are trained with cross-entropy loss, an additional analysis of the Instance Similarity attacks against CRIAGE for the DistMult model trained with BCE loss is performed. Table \ref{tab:ia_bce_analysis} shows the reduction in MRR and Hits@1 due to adversarial deletions in this training setting. It is observed that the Instance Similarity attacks outperform the baseline for this setting as well.

\begin{table}[h]
    \centering
    \small
    \begin{tabular}{c  c c   c c}
    \toprule
                  &  \multicolumn{2}{c}{\textbf{WN18RR}}       &   \multicolumn{2}{c}{\textbf{FB15k-237}}   \\
                  \cmidrule(lr){2-3}  \cmidrule(lr){4-5}  
                  &  \textbf{MRR}   &  \textbf{Hits@1}    &   \textbf{MRR}   &  \textbf{Hits@1}  \\
    \midrule
         Original &     1.00   &   1.00    &  1.00  &  1.00  \\
         CRIAGE  &     0.67   &   0.63    &  0.63  &  0.46   \\
         Dot Metric    &     0.86   &   0.81    &  0.61  &  0.44  \\
         $\bm{\ell_2}$ Metric   &     \textbf{0.12}   &   \textbf{0.06}    &  0.60  &  0.43  \\
         Cos Metric   &     \textbf{0.12}   &   \textbf{0.06}    &  \textbf{0.58}  &  \textbf{0.38}  \\
    \bottomrule
    \end{tabular}
    \caption{Reduction MRR and Hits@1 due to adversarial deletions for DistMult (trained with BCE loss) on WN18RR and FB15k-237}
    \label{tab:ia_bce_analysis}
\end{table}

\chapter{Additional Details for Inference Pattern Attacks}
\section{Implementation Details}
\label{apx:inference_pattern_implementation}
\subsection{Training KGE models} 
The codebase\footnote{\url{https://github.com/PeruBhardwaj/InferenceAttack}} for KGE model training is based on the codebase from \citet{dettmers2018conve}\footnote{https://github.com/TimDettmers/ConvE}. The 1-K training protocol is used but without adding the reciprocal relations to the training dataset. Each training step alternates through batches of (s,r) and (o,r) pairs and their labels. The model implementation uses an if-statement for the forward pass conditioned on the input batch mode. 

For the TransE scoring function, an L2 norm and a margin value of 9.0 are used. The loss function used for all models is Pytorch's BCELosswithLogits. For regularization, label smoothing and L2 regularization are used for TransE; and input dropout with label smoothing for remaining models. For ConvE, hidden dropout and feature dropout are also used.  

To ensure same hyperparameters for original and poisoned KGE models, early stopping is not used. An embedding size of 200 is used for all models on both datasets. For ComplEx, this becomes an embedding size of 400 because of the real and imaginary parts of the embeddings. All hyperparameters are tuned manually based on suggestions from state-of-art implementations of KGE models \citep{ruffinelli2020olddognewtricks, dettmers2018conve}. 
The hyperparameter values for all model dataset combinations are available in the codebase. 
Table \ref{tab:rip_original_mrr} shows the MRR and Hits@1 for the original KGE models on WN18RR and FB15k-237. 

For re-training the model on poisoned dataset, same hyperparameters as the original KGE model training are used.
All experiments for model training, adversarial attacks and evaluation are run on a shared HPC cluster with Nvidia RTX 2080ti, Tesla K40 and V100 GPUs. 

Experimental results reported in Section \ref{sec:rip_results} of the thesis were obtained from the codebase version at commit $afb8202$, which is available at \url{https://github.com/PeruBhardwaj/InferenceAttack/tree/afb8202}.

\begin{table}[]
    \centering
    \small
    \begin{tabular}{c  c c   c c}
    \toprule
                  &  \multicolumn{2}{c}{\textbf{WN18RR}}       &   \multicolumn{2}{c}{\textbf{FB15k-237}}   \\
                  \cmidrule(lr){2-3}  \cmidrule(lr){4-5}  
                  &  \textbf{MRR}   &  \textbf{Hits@1}    &   \textbf{MRR}   &  \textbf{Hits@1}  \\
    \midrule
         DistMult &     0.42   &   0.39    &  0.27  &  0.19  \\
         ComplEx  &     0.43   &   0.40    &  0.24  &  0.20   \\
         ConvE    &     0.43   &   0.40    &  0.32  &  0.23  \\
         TransE   &     0.19   &   0.02    &  0.34  &  0.25  \\
    \bottomrule
    \end{tabular}
    \caption{MRR and Hits@1 results for original KGE models on WN18RR and FB15k-237}
    \label{tab:rip_original_mrr}
\end{table}


\subsection{Baseline Implementation Details}
\label{apx:inference_pattern_ijcai_baseline}
One of the baselines in this evaluation is the Direct Attack from \cite{zhang2019kgeattack}. It proposed edits in the neighbourhood of subject of the target triple. This attack is extended to both subject and object to match the evaluation protocol in the thesis. Since no public implementation of the attack is available, the author implemented the attack herself.

The attack is based on computing a perturbation score for all possible candidate additions. Since the search space for candidate additions is of the order $\mathcal{E} \times \mathcal{R}$, the attack uses random down sampling to filter out the candidates. The percent of triples down sampled are not reported in the original paper and the implementation is not available. So, in this research, a high value and a low value of the percentage of triples down sampled are picked and adversarial edits are generated for both fractions. 
The high and low percent values that were used to select candidate adversarial additions for WN18RR are DistMult: (20.0, 5.0); ComplEx: (20.0, 5.0); ConvE: (2.0, 0.1); TransE: (20.0, 5.0). For FB15k-237, these values are DistMult: (20.0, 5.0); ComplEx: (15.0, 5.0); ConvE: (0.3, 0.1); TransE: (20.0, 5.0)

\begin{table}[]
    \centering
    \small
    \begin{tabular}{c  ccc   }
    \toprule
                  &  \multicolumn{3}{c}{\textbf{WN18RR}}         \\
                  \cmidrule(lr){2-4} 
                  &  \textbf{Original}   &  \textbf{High}    &   \textbf{Low}     \\
    \midrule
         DistMult &     0.90      &   0.82        &   0.83        \\
         ComplEx  &     0.89      &   0.76        &   0.79       \\
         ConvE    &     0.92      &   0.90       &   0.90       \\
         TransE   &     0.36      &   0.25        &   0.24        \\
         
    \midrule
                 &   \multicolumn{3}{c}{\textbf{FB15k-237}}   \\
                  \cmidrule(lr){2-4}
                  &  \textbf{Original}   &  \textbf{High}    &   \textbf{Low} \\
    \midrule
        DistMult    &  0.61     &  0.55    &   0.53     \\
        ComplEx     &  0.61     &  0.51    &   0.52     \\
        ConvE       &  0.61     &  0.54     &   0.54    \\
        TransE      &  0.63     &  0.57    &   0.57     \\
                  
    \bottomrule
    \end{tabular}
    \caption{MRR of KGE models trained on original datasets and poisoned datasets from the attack in \citet{zhang2019kgeattack}. High, Low indicate the high and low percentage of candidates used for attack.}
    \label{tab:rip_ijcai_results}
\end{table}

Thus, two poisoned datasets are generated from the attack - one that used a high number of candidates and another that used a low number of candidates. Two separate KGE models are trained on these datasets to assess the attack performance. Table \ref{tab:rip_ijcai_results} shows the MRR of the original model; and poisoned KGE models from attack with high and low downsampling percents. The results reported for this attack's performance in Section \ref{sec:rip_results} are the better of the two results (which show more degradation in performance) for each combination.


\subsection{Attack Implementation Details}

The relation inference pattern based attacks proposed in this research involve three steps to generate the adversarial additions for all target triples. For step1 of selection of adversarial relations, the inversion and composition relations are pre-computed for all target triples. Step2 and Step3 are computed for each target triple in a \emph{for} loop. These steps involve forward calls to KGE models to score adversarial candidates. For this, a vectorized implementation similar to KGE evaluation protocol is used. The adversarial candidates that already exist in the training set are also filtered out. Any duplicates from the set of adversarial triples generated for all target triples are further filtered out. 

For the composition attacks with soft-truth score, the KMeans clustering implementation from \(\mathtt{scikit-learn}\) is used. The elbow method is used on the grid [5, 20, 50, 100, 150, 200, 250, 300, 350, 400, 450, 500] to select the number of clusters. The number of clusters selected for WN18RR are DistMult: 300, ComplEx: 100, ConvE: 300, TransE: 50. For FB15k-237, the numbers are DistMult: 200, ComplEx: 300, ConvE: 300, TransE: 100.

\begin{small}
\cleardoublepage
\manualmark
\markboth{\spacedlowsmallcaps{\bibname}}{\spacedlowsmallcaps{\bibname}} 
\refstepcounter{dummy}
\addtocontents{toc}{\protect\vspace{\beforebibskip}} 
\addcontentsline{toc}{chapter}{\tocEntry{\bibname}}
\label{app:bibliography}
\bibliography{Bibs/Bibliography.bib}  

\end{small}

\end{document}